\documentclass[10pt,twocolumn,letterpaper]{article}

\usepackage{cvpr}              % To produce the CAMERA-READY version
\definecolor{cvprblue}{rgb}{0.21,0.49,0.74}
\usepackage[pagebackref,breaklinks,colorlinks,allcolors=cvprblue]{hyperref}

\def\paperID{*****} % *** Enter the Paper ID here
\def\confName{CVPR}
\def\confYear{2026}

\title{What Are You Doing? A Closer Look at Controllable Human Video Generation}

\author{
Emanuele Bugliarello\textsuperscript{\href{mailto:bugliarello@google.com}{\faEnvelope}} \hspace{0.25cm} Anurag Arnab \hspace{0.25cm} Roni Paiss \\ Christy Koh \hspace{0.25cm} Pieter-Jan Kindermans \hspace{0.25cm} Cordelia Schmid \\[1mm]
Google DeepMind\\[1mm]
{\tt\small {\faGithub}~\url{https://github.com/google-deepmind/wyd-benchmark}}
}

\usepackage{microtype}
\usepackage{booktabs} % for professional tables
\usepackage{mathtools}
\usepackage{tikz}
\usetikzlibrary{bayesnet}
\usetikzlibrary{arrows}
\usepackage{tabularx}
\usepackage{tabulary}
\usepackage{colortbl}
\usepackage{xspace}
\usepackage{float}
\usepackage{wrapfig}
\usepackage[accsupp]{axessibility}  % Improves PDF readability for those with disabilities.

\usepackage{bold-extra}

\newcommand\doublecheck{{\checked\kern-0.5em\checked}}
\usepackage{fontawesome5}
\newcommand\checked{\footnotesize \faIcon{check}}
\usepackage{mathtools}
\newcounter{tableeqn}[table]

\newcounter{tablesubeqn}[tableeqn]

\usepackage{relsize}
\usepackage{amsmath, amsthm, amssymb}
\usepackage{mathrsfs}

\usepackage{algorithm2e}
\RestyleAlgo{ruled}

\usepackage{caption}
\usepackage{graphicx}
\usepackage{enumitem}

\usepackage{color, colortbl}
\PassOptionsToPackage{table,dvipsnames}{xcolor}
\usepackage{multirow,makecell}
\usepackage{rotating}
\usepackage{booktabs}
\usepackage[table]{xcolor}
\usepackage{tabularx}
\usepackage{lscape}
\makeatletter
\DeclareRobustCommand\onedot{\futurelet\@let@token\@onedot}
\def\@onedot{\ifx\@let@token.\else.\null\fi\xspace}

\def\eg{\emph{e.g}\onedot} 
\def\ie{\emph{i.e}\onedot}

\makeatother

\newcommand{\band}{\rowcolor{gray!10}}

\definecolor{myred}{HTML}{DC267F}
\definecolor{mygreen}{HTML}{2593A2}

\definecolor{myblue}{HTML}{0053d6}

\newcommand{\wydlong}{What Are You Doing?}
\newcommand{\wydbold}{{\bfseries{\scshape wyd}}}
\newcommand{\wyd}{{\scshape wyd}}
\newcommand{\tiktok}{{TikTok}\xspace}
\newcommand{\tedtalks}{{TED-Talks}\xspace}
\newcommand{\humanvid}{{HumanVid}\xspace}

\newcommand{\magicanimate}{{MagicAnimate}\xspace}
\newcommand{\magicpose}{{MagicPose}\xspace}
\newcommand{\mimicmotion}{{MimicMotion}\xspace}
\newcommand{\controlnext}{{ControlNeXt}\xspace}
\newcommand{\cavideo}{{Control-A-Video}\xspace}
\newcommand{\tftv}{{TF-T2V}\xspace}
\newcommand{\ctrladapter}{{Ctrl-Adapter}\xspace}
\newcommand{\vace}{{VACE}\xspace}

\begin{document}

\twocolumn[{%
\renewcommand\twocolumn[1][]{#1}%
\maketitle
\vspace{-\baselineskip}
\includegraphics[width=\linewidth, trim={0 11cm 0 0}, clip]{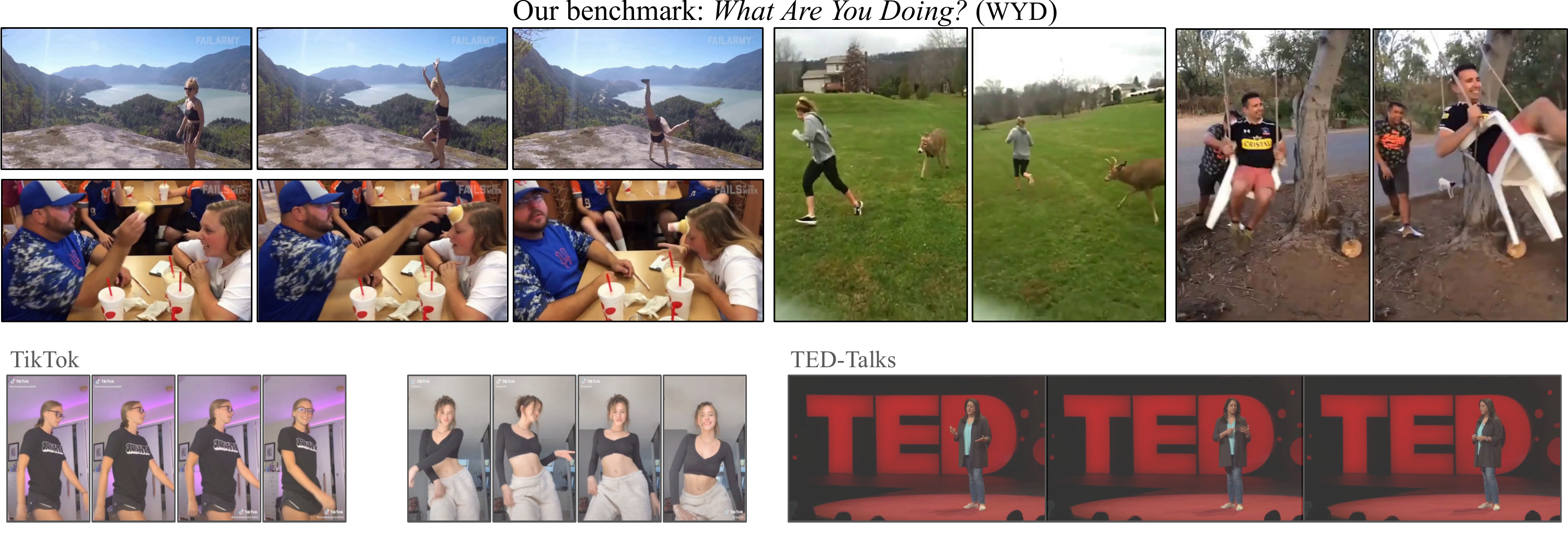}
\vspace{-2.5em}
\captionof{figure}{
    Samples from our \wyd{} dataset (above) and from the commonly used datasets for controllable human video generation (below).
    \wyd{} contains significantly more diverse videos, in terms of number of actors, actions, interactions, scenes as well as camera motion.
    \vspace{1.5em}
}
\label{fig:teaser}
}]

\begin{abstract}
High-quality benchmarks are crucial for driving progress in machine learning research. However, despite the growing interest in video generation, there is no comprehensive dataset to evaluate human synthesis. Humans can perform a wide variety of actions and interactions, but existing datasets, like TikTok, TED-Talks, and HumanVid, lack the diversity and complexity to fully capture the capabilities of video generation models. We close this gap by introducing `What Are You Doing?' (WYD): a new benchmark for fine-grained evaluation of controllable image-to-video generation of humans. WYD consists of 1{,}544 captioned videos that have been meticulously collected and annotated with fine-grained categories. These allow us to systematically measure performance across 9 aspects of human generation, including actions, interactions and motion.
We also propose an evaluation framework, where we adapt existing metrics for better human- and video-level assessment, as shown by human preference.
Equipped with our dataset and metrics, we perform in-depth analyses of state-of-the-art open-source models in controllable image-to-video generation, showing how WYD provides novel insights about their capabilities.
We release data and code to drive progress in human video generation.
\end{abstract}
\vspace{-3.415\baselineskip}
\section{Introduction}

\begin{figure*}[t!]
  \centering
  \includegraphics[width=\linewidth]{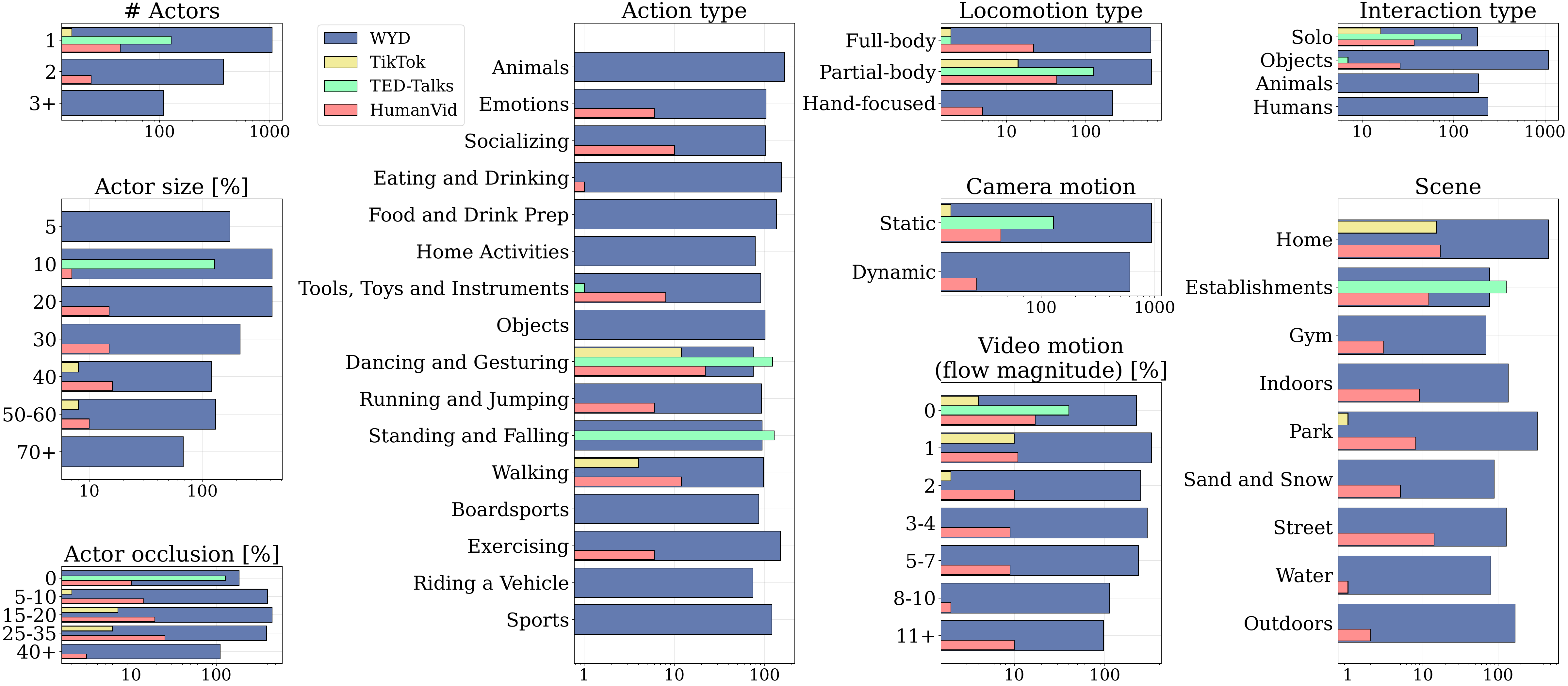}
  \vspace{-1.5em}
  \caption{
  \textbf{Diversity of \wyd}. \wyd{} contains manually-labeled fine-grained annotations for 9 categories and 56 sub-categories relevant to human video generation.
  Manually annotating existing datasets, we verify that \wyd{} is larger and more diverse across all the categories.
  }
  \label{fig:stats_distr}
  \vspace{-.5\baselineskip}
\end{figure*}

Video generation has seen explosive progress in recent years, with state-of-the-art generative models demonstrating unprecedented capabilities in creating photorealistic and coherent video sequences from text prompts~\citep{metamoviegen,wan2025wan,gao2025seedance}.
Nevertheless, generating humans is exceptionally hard as models must contend with the sheer complexity and high dimensionality of realistic human motion and appearance, including non-rigid body deformations and nuanced world interactions.
To drive progress in this critical domain, we introduce \wyd: a benchmark to comprehensively and systematically evaluate model performance against the inherent challenges of human fidelity, temporal stability, and fine-grained human motion.

We specifically focus on pose-controllable generation, given the complexity of precisely describing human movements in words and its practical applications (\eg, simulating a risky stunt in a movie, or making your uncle dance).
Generally, controllability is critical for generative models, as it enables artists to precisely specify \emph{how} a generative model creates content, by conditioning it spatially (\eg, via poses, masks, depth and edge maps~ \cite{peekaboo,followyourpose,humanvid}) and temporally (\eg, via motion vectors and camera positions~\cite{cameractrl,motioni2v}).

As the human brain is an innate expert at recognizing biological motion and appearance, achieving high-quality, controllable human generation stands as a critical feature of generative models.
However, this area has received limited attention so far, and the community typically relies on two datasets: \tiktok~\citep{tiktok} and \tedtalks~\citep{tedtalks}.
They have limited size (\cref{tab:stats}) and a \emph{narrow scope}, such as a single person dancing or talking in a static shot (see \cref{fig:teaser,fig:stats_distr}), making them insufficient to assess the full capabilities of models.

To bridge the gap between model capabilities and what evaluations can capture, we introduce `\emph{\wydlong}'(\wyd): a new dataset of complex, dynamic videos with a wide variety of human appearance, actions and interactions.
\wyd{} includes diverse and high-quality videos filtered semi-automatically, and rigorously finalized through manual annotations (2500+ rater hours).
In particular, we built \wyd{} considering the typical setup of controllable image-to-video generation (\eg, animating a person with specific movements~\cite{Chan_2019_ICCV,Liu_2019_ICCV,9178983,Zhang_2022_CVPR,Bhunia_2023_CVPR,Ni_2023_CVPR,Karras_2023_ICCV,disco,wang2024leo,magicanimate,10.1007/978-3-031-73202-7_19}) by ensuring human actors are clearly visible in the first frame.

In addition, we argue that a detailed evaluation of model capabilities across fine-grained categories is needed to pinpoint developmental areas and to compare different models, but this level of detail is missing from current benchmarks~\citep{tiktok,tedtalks,humanvid}.
We label the 1{,}544 samples in \wyd{} according to nine categories, which enable systematic evaluations of key video-level and human-specific aspects of video generation, and make \wyd{} significantly larger and broader than existing benchmarks (as shown in \cref{tab:stats,fig:stats_distr}).

To reliably measure generated videos of humans, we examine which existing metrics capture key aspects of video generation (\ie, video quality, per-frame correctness, and video motion) as well as human-centric ones (\eg, pose adherence).
In particular, we enable human-specific metrics by customizing existing metrics on manually-curated segmentation masks for the \emph{human actors} (\cref{fig:human_metrics} in \cref{sec:wyd_metrics}).

As a representative use case of our benchmark and evaluation protocol, we investigate eight families of video generation models, including models conditioned on pose~\citep{magicanimate,magicpose,mimicmotion,controlnext,vace} as well as models guided via depth and edge maps~\citep{controlavideo,ctrladapter,tft2v}.
Our results from both human and automatic evaluations show that, using \wyd, we can diagnose several limitations of current SOTA models, facilitated by applying existing metrics to videos belonging to the diverse and fine-grained categories unique to our dataset.
These include challenges that cannot be measured with previous datasets, such as generating cross-shot or atypical movements, interactions with objects, and more.

\begin{table}[t!]
  \setlength{\tabcolsep}{3.0pt}
  \smaller
  \centering
  \resizebox{\linewidth}{!}{
    \begin{tabular}{lrrrr}
\toprule
 & \multicolumn{1}{c}{\textbf{TikTok}} & \multicolumn{1}{c}{\textbf{TED-Talks}} & \multicolumn{1}{c}{\textbf{HumanVid}} & \multicolumn{1}{c}{\wydbold} \\
\midrule
\textcolor{black}{\# Videos (unique clips)} & 16 (14) & 128 (40) & 71 (71) & 1{,}544 (1{,}393) \\
\band\textcolor{black}{Video duration [s]} & 8.3--23.0 & 4.3--23.1 & 2.7--87.2 & 1.5--15.0 \\
\textcolor{black}{Video aspect ratio} & portrait & landscape & portrait, landscape & portrait, landscape \\
\band\textcolor{black}{\# Actors} & 1 & 1 & 1, 2 & 1, 2, 3--8 \\
\multirow{5}{*}{\textcolor{black}{Categories}} & \multirow{5}{*}{N/A} & \multirow{5}{*}{N/A} & \multirow{5}{*}{N/A} & \# actors, actor size\\& & & & \% occlusion, actions\\ & & & & scene, interactions\\ & & & & locomotion, camera\\ & & & &  video motion\\
\band\textcolor{black}{Additional annotations} & dense poses & N/A & N/A & segmentation masks,\\
\band\textcolor{black}{}& & & & 2D pose keypoints\\
\bottomrule
\end{tabular}

  }
  \vspace{-0.5em}
  \caption{
  \textbf{Overview of data statistics.} \wyd{} complements existing datasets for human video generation with more videos, higher diversity and fine-grained categories. See \cref{fig:stats_distr} for more statistics.}
  \label{tab:stats}
  \vspace{-1.\baselineskip}
\end{table}

Our contributions are threefold:
(\emph{i}) We identify the limitations of existing benchmarks for controllable human generation, and meticulously collect \wyd: a large and diverse benchmark with fine-grained annotations.
(\emph{ii}) We propose an evaluation protocol, and show how we can adopt and repurpose existing metrics to capture both video- and human-level aspects, which are validated by human preference.
(\emph{iii}) We conduct system-level evaluations of eight SOTA open-source models on \wyd, and show it is harder than existing benchmarks, revealing systematic limitations and narrow training data distributions, that were previously undetectable.

\section{Related work} \label{sec:rw}

\paragraph{Controllable video generation.}
Recent algorithmic and data engineering innovations led to major progress in video generation~\citep{diffusion,imagenvideo,videodiff,alignyourlatents}.
This increase in quality has encouraged a shift towards controllable generation, giving users finer control and precision than possible from text alone~\citep{controlnet,ma2025controllable}.
For example, providing a reference video is a far simpler input than laboriously specifying a person's precise movements in a scene via text.
Specifically, for human video generation, 2D body keypoints are the most frequently used signal, offering a sparse yet easily interpretable way to condition the output~\citep{animateanyone,tuneavideo,disco,magicanimate,lei2024comprehensive}.
Other common visual controls include depth and edge maps~\citep{controlavideo,controlvideo}, bounding boxes~\citep{motionzero,boximator}, and camera angles~\citep{motionctrl,cameractrl}. 

\paragraph{Benchmarks for video generation.}
As the field gains momentum and video models achieve greater maturity, researchers are starting to establish standardized benchmarks to ensure reliable advancements.
For the common text-to-video task, recent benchmarks consist of a collection of text prompts, designed to evaluate the video quality and video--text alignment in open domains.
For example, VBench~\citep{vbench}, EvalCrafter~\citep{evalcrafter} and Movie Gen Bench~\citep{metamoviegen} propose comprehensive prompt sets that span multiple aspects of generic video generation.
T2V-CompBench~\cite{t2v_compbench} focuses on compositionality, while StoryBench~\citep{storybench} evaluates video generations from a sequence of text prompts.
I2V-Bench~\citep{consisti2v} and AIGCBench~\citep{aigcbench} assess consistency for the image-to-video task.
Notably, most of the existing benchmarks are not suitable for evaluating controllable video generation as they lack reference videos, which are needed for extracting control signals, and for comparison during evaluation.
Therefore, researchers typically rely on existing video understanding datasets~\citep{davis2017,ucf101} or collect small, non-representative, datasets for human evaluations~\citep{controlnext,vace}.

\paragraph{Benchmarks for human video generation.}
The most common benchmarks for controllable human video generation are TikTok~\citep{tiktok} and TED-Talks~\citep{tedtalks,disco}.
The TikTok dataset consists of just 16 videos scraped from TikTok, where a single person, typically at home, covers a large, centered part of the video and mostly dances in place.
The TED-Talks dataset contains 128 videos from 40 talks at TED events, where a single person is on stage and gesticulates while speaking on their spot.
We believe these benchmarks are extremely narrow and fail to capture the richness of human actions and interactions.
Recently, HumanVid~\cite{humanvid} proposed a more diverse, yet small validation set of just 71 videos.
This paper presents a new benchmark for controllable human video generation (\wyd) that addresses their shortcomings.
Compared to existing datasets, \wyd{} is orders of magnitude larger and more diverse (\cref{tab:stats} and \cref{fig:stats_distr}), going beyond the status quo of static, single-person close-up videos.

\paragraph{Metrics for human video generation.}
Previous work typically evaluates models through pixel-level metrics, like FID, FVD, SSIM and LPIPS~\citep{wang2004image,zhang2018perceptual,fid, fvd}.
However, we argue that controllable human video generation requires fine-grained metrics that specifically measure alignment with the humans in reference videos.
Thus, we adapt existing metrics for subject consistency~\citep{consisti2v} and motion adherence~\citep{coco} to general, multi-person videos by collecting and leveraging segmentation tracks for the actors in each \wyd{}~video.

\section{The \emph{\wydlong} (\wyd{}) benchmark} \label{sec:benchmark}

Our goal is to benchmark the capability of video generation models to successfully synthesize human subjects in authentic, real-world settings.
To effectively do so, a benchmark that satisfies different desiderata is required.
Specifically, we need videos where people (\emph{i}) are visible, (\emph{ii}) perform a variety of actions, (\emph{iii}) interact with each other and other objects, within (\emph{iv}) complex and dynamic scenes.

In this section, we describe our approach to curate videos that meet these criteria from previously published datasets, and how we categorize them in order to assess different aspects typical of human video generation.
We are primarily interested in human animation, given its practical applications and popularity in the field~\citep{lei2024comprehensive}.
We refer to our new dataset as \emph{\wydlong} (\wyd).

\begin{figure*}[t!]
  \centering
  \includegraphics[width=\linewidth, trim={0 0 0 0}, clip]{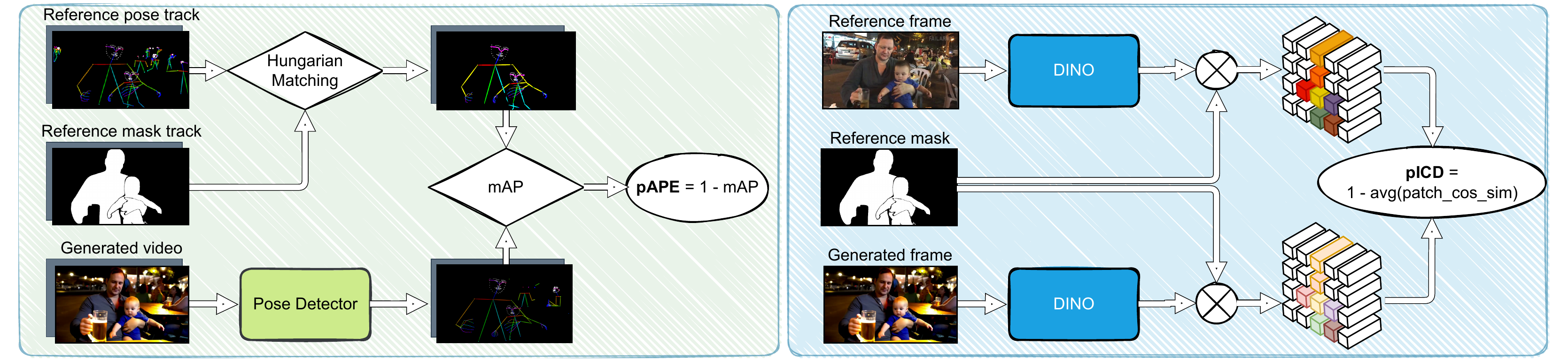}
  \caption{
  \textbf{Overview of our human-centric metrics.} We leverage our segmentation masks to define metrics for person consistency (pICD; left) and pose adherence (pAPE, right) that align well with human preferences (\cref{sec:metrics}).
  }
  \label{fig:human_metrics}
\end{figure*}

\subsection{Data filtering} \label{sec:method_data_pipeline}

To construct a \emph{generic} benchmark for human video generation, we require videos that capture a wide range of human activities and environments.
Therefore, we rely on three publicly licensed datasets collected from Internet platforms, such as YouTube and Flickr: Kinetics~\citep{kinetics400}, DiDeMo~\citep{didemo} and Oops~\citep{oops}.
These datasets include humans with different body poses, clothing, age and background, performing a great variety of actions (including unintentional actions that serve as interesting testbeds for modeling atypical motion).

At a high-level, we (1) start by filtering out videos where the main actors are not humans, and (2) use a shot detector~\citep{pyscenedetect} to remove scene cuts.
We then (3) make sure that human actors are visible in the first and most of the frames using a human pose estimator~\citep{dwpose}. 
We further remove videos that (4) are either too short or too long (1.5--15s), whose (5) video--caption pairs have low similarity according to a fine-grained VLM~\citep{pyramidclip}, and (6) have low resolution.
For details and examples of output videos at each step, see App.~\ref{app:data_prep}.

Finally, we (7) meticulously scrutinize the quality of the resulting videos and remove those with (i) significant blur, (ii) poor lighting, (iii) unstable camera, (iv) low motion, or (v) where the first frame does not capture the main actors.
This process, along with the video categorization described below, was done in multiple rounds and took over 500 hours.
For further quality control, the authors validated the annotations and ensured the videos' quality and high diversity, and that they would serve as a challenging and comprehensive testbed of human video generation for future models.

At the end, \wyd{} contains 1{,}544 high-quality videos (from the original 18{,}351), which enable the fine-grained analyses, such as those described in \cref{sec:results}, within a tractable runtime.

\subsection{Video categorization} \label{sec:method_category}

Our benchmark moves beyond the single aggregated score typical of existing datasets, aiming instead to provide a detailed, multi-faceted understanding of a model's capabilities in human video generation.
Thus, we annotate our data with nine categories that capture crucial aspects of human synthesis.
Each category includes sub-categories (\cref{fig:stats_distr}), all with at least $\approx$100 samples for sufficient statistical~relevance.

For each video in \wyd{}, we manually label: the number of human actors; the average area they cover; the type of action they perform; the kind of body movement; how they interact with their environment; which scene they are in; and whether the camera follows them.
Additionally, we use an optical flow model~\citep{teed2020raft} to estimate the amount of video motion, and a pose detector~\citep{dwpose} to estimate the amount of occlusion.
For details, please refer to App.~\ref{app:data_prep}.

Overall, we find that only a few categories overlap with each other significantly (\cref{fig:category_overlaps} in App.~\ref{app:data_prep}).
Namely, actions and interactions with animals; and video and camera motion, where high-motion videos come from dynamic camera, and videos with low motion correspond to static camera.

\subsection{Dense video annotations}
In addition to labeling our videos with nine categories, we annotate each human actor in a video with tracked segmentation masks, which enable better evaluations of generated humans (\cref{sec:wyd_metrics}).
These are first obtained through SAM~2~\citep{sam2}, and then manually corrected (1{,}000 rater hours).
To verify the reliability of our evaluation framework (\cref{sec:results}), we also manually label (1{,}000 rater hours) the 2D pose keypoints for the actors in 100 representative videos (detailed in App.~\ref{app:data_prep}).

\section{Evaluation protocol}\label{sec:wyd_metrics}

We propose the following evaluation framework (validated in \cref{sec:metrics}) to measure performance on \wyd{} across perceptual and alignment metrics: visual quality, video motion similarity, person consistency and human motion adherence.
Notably, while previous work only evaluates across entire videos, we also investigate human-specific aspects by adapting existing metrics with our manually-collected masks.
\vspace{-.15\baselineskip}

\paragraph{Video-level evaluations.}
We use the following existing metrics to assess video-level performance (see~App.~\ref{app:human_evals}).

\begin{itemize}[itemsep=4pt,topsep=1pt]
    \item \textbf{Video quality (FVD):} Following prior work~\citep{consisti2v,humanvid}, we use the Fr\'echet Video Distance (FVD;~\citep{fvd}) with an I3D backbone~\citep{i3d} to compare the distributions of generated videos and reference videos.
    \item \textbf{Video motion (OFE):} Following prior work~\citep{ctrladapter,evalcrafter}, we compute the optical flow endpoint error (OFE) as a measure of structural dissimilarity between reference and generated videos. We use RAFT~\citep{teed2020raft} to extract optical flows.
\end{itemize}
\vspace{-.15\baselineskip}

\begin{figure*}[t!]
  \centering
  \includegraphics[width=\linewidth]{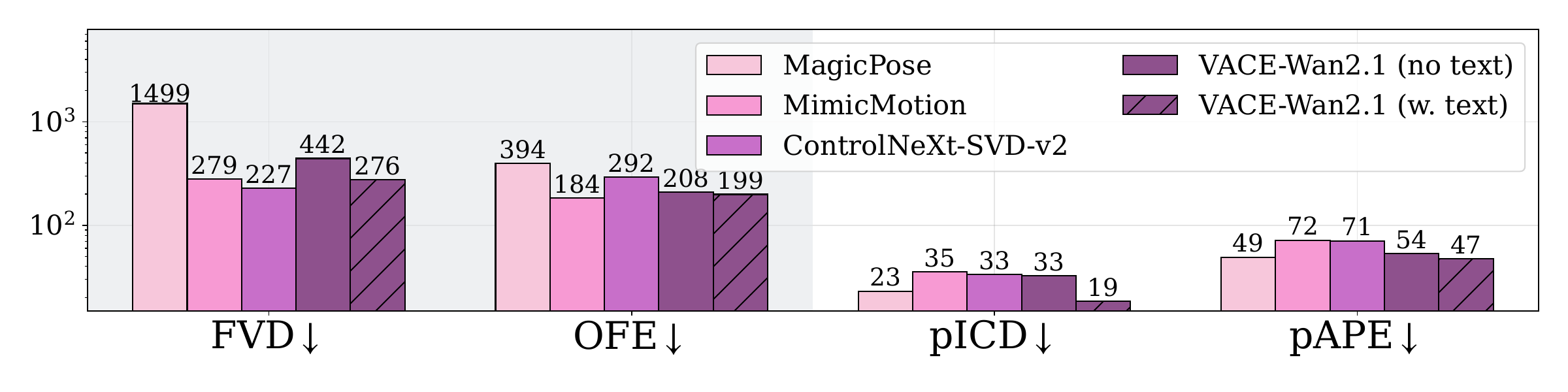}
  \vspace{-2.5\baselineskip}
  \caption{
  \textbf{Overall performance (left: video-level, right: human-level) of SOTA controllable pose-conditioned image-to-video models on \wyd$_{16}$.}
  For \vace, we report results when conditioning on both text and poses (default setup) as well as on poses only (see also ablation in \cref{fig:vace_ablations}), while the other models do not support text inputs.
  Human generation is multifaceted and no model prevails across all metrics.}
  \label{fig:pose_overall}
\end{figure*}

\paragraph{Human-level evaluations.}
As humans are the main focus of \wyd, we adapt existing metrics to measure person-specific performance by leveraging our segmentation masks. 
\begin{itemize}[itemsep=4pt,topsep=1pt]
    \item \textbf{Person consistency (pICD):} The ability to generate coherent human subjects is a central aspect for human video generation.
    \citet{consisti2v} use DINO~\citep{Caron_2021_ICCV} feature similarity across frames to measure subject consistency.
    Inspired by their results, we adapt this metric for human consistency.
    For each frame, we compute the average cosine similarity between the DINOv2~\citep{dinov2} features of the patches corresponding to people (\cref{fig:human_metrics} right).
    For consistency with other error metrics, we report the cosine distance and call this metric the `person image cosine distance' (pICD).
    \item \textbf{Human movement (pAPE):}
    Another crucial requirement for controllable models is for their generated humans to strictly adhere to the underlying driving signal.
    We adapt the standard average precision metric~\citep{coco,followyourpose} between the reference poses and those detected, using DWPose~\citep{dwpose}, in the generated video.
    As a scene may contain extraneous or background human subjects, we use the segmentation masks to map the poses detected in the generated video to the poses corresponding to the human \emph{actors} (\ie, the salient ones) in the reference video (\cref{fig:human_metrics} left).
    We measure AP for each video separately, and report the complement of the average AP (\ie, $1-\text{mAP}$) as an error metric for pose adherence, \ie, `person AP error' (pAPE).
\end{itemize}

\section{Results} \label{sec:results}

We use our new \wyd{} dataset to investigate, for the first time, how difficult different facets of human generations are for state-of-the-art controllable video generation models (as representative models capable of precise human generation).
We show that \wyd{} allows us to pinpoint five novel failures in 8 existing models.
To facilitate this, we open-source the data and the code to fully reproduce our results, which required significant resources (over 6{,}000 A100 GPU hours).
For fair comparisons, our evaluations are at 16fps (supported by all models), and we refer to this version as \wyd$_{16}$.

In this section, we overlay our results by referring to specific failures of our evaluated video generation models.
Due to limited space, we report visual samples in App.~\ref{app:samples}.

\paragraph{Experimental setup.}
Pose key-points are the most common way to condition generative models for human synthesis.
They are relatively sparse, and allow artists to quickly generate humans in specific poses without having to match specific body measures~\citep{ma2017pose,controlnet,wang2025humandreamer}.
We evaluate four state-of-the-art (SOTA), open-source image-to-video models with pose conditioning: \magicpose~\citep{magicpose}, \mimicmotion~\citep{mimicmotion}, ControlNeXt-SVD-v2~\citep{controlnext} and VACE-Wan2.1~\citep{vace}.

Besides pose-guided models, we evaluate three open-source, SOTA models that can be guided with depth or edge maps: \cavideo~\citep{controlavideo}, \ctrladapter~\citep{ctrladapter} and \tftv~\citep{tft2v}.
App.~\ref{app:more_results} also reports performance for text-to-video~\citep{yang2025cogvideox}, image-to-video~\citep{stablevideodiffusion} and dense-pose~\citep{magicanimate} models.

We evaluate open-source models for reproducibility, and point out that no closed-source model supports direct conditioning with poses, edge or depth maps.
We also note that our evaluated models have been trained on internal datasets (or private subsets of public data) which did not include any \wyd{} videos, as confirmed by the authors (\cref{tab:models} in App.~\ref{app:samples}).
In particular, most pose-based models have largely been trained on close-up single-person videos.
While they may incur a distribution shift on \wyd{}, we believe controllable human generation should go beyond those simple cases, and hope \wyd{} will drive progress towards more general videos.

\paragraph{Overall performance on \wydbold$_{16}$.}
\cref{fig:pose_overall} shows the overall performance of pose-guided models on \wyd$_{16}$.
We see that \vace achieves overall better performance, being on-par with \magicpose in generating \emph{humans} (pICD, pAPE) whilst also obtaining competitive video-level performance (FVD, OFE) with \mimicmotion and \controlnext. \magicpose instead obtains a high FVD value due to flickering artifacts.

To show the versatility of our benchmark, we analyze depth- and edge-guided models in \cref{fig:dense_overall} (App.~\ref{app:more_results}).
While \tftv tends to outperform all other models, \cavideo and \ctrladapter have low visual quality (FVD, pICD).
Inspecting \cavideo's generations, for example, we find that its videos follow the underlying driving signal well but at the expenses of distorted colors and artifacts.

\begin{figure}[t]
  \centering
  \includegraphics[width=\linewidth, trim={0 0 0 0}, clip]{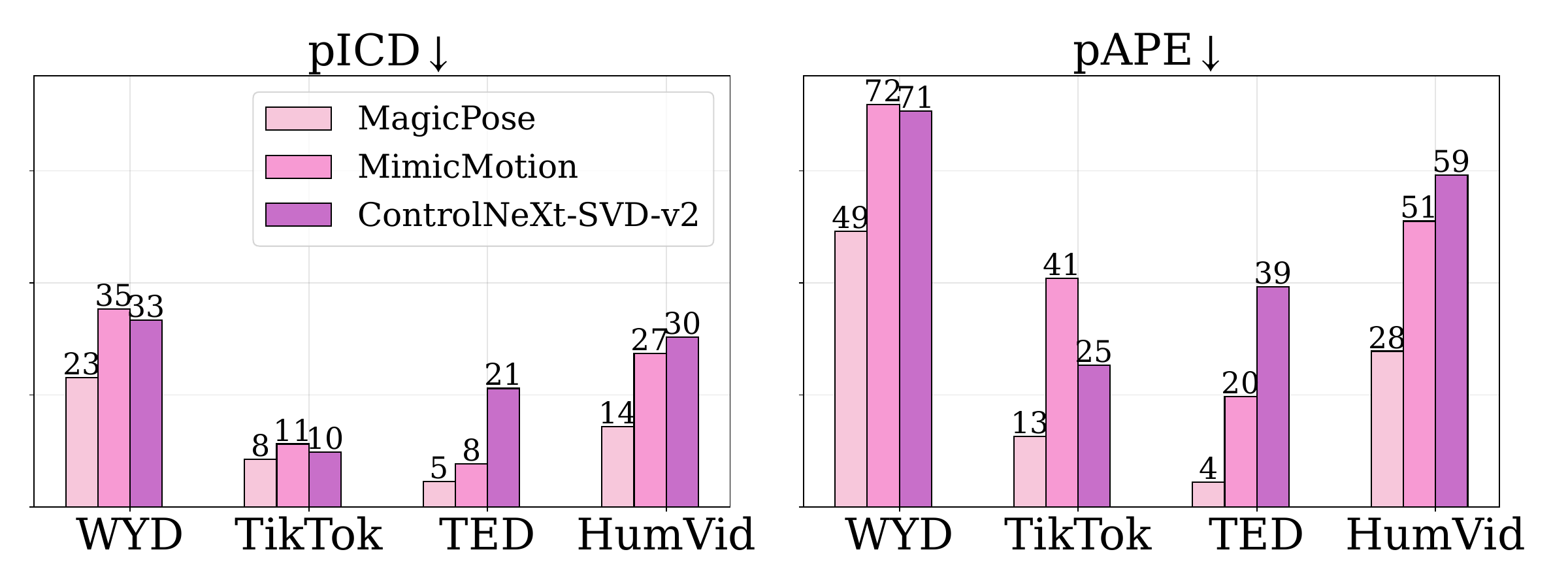}
  \vspace{-1.7\baselineskip}
  \caption{
  \textbf{Performance comparison between \wyd, \tiktok, \tedtalks and \humanvid for models conditioned on image+pose.} \wyd{} yields larger errors, confirming that its greater diversity (wrt categories and size) is more challenging for state-of-the-art models.
  }
  \label{fig:dsets_deltas}%
  \vspace{-.5\baselineskip}
\end{figure}

\paragraph{\wydbold{} is harder than previous benchmarks.}
In \cref{fig:dsets_deltas}, we compare the errors in human metrics between \wyd{} and \tiktok{} or \tedtalks{}.
\wyd{} is consistently harder in both person consistency and movement, with error rates 1.8--4.6$\times$ higher for pICD, and 1.8--12.3$\times$ higher for pAPE.
That is, \wyd{} poses significant challenges to SOTA systems, suggesting it can be used to drive progress in this area.
Moreover, in \cref{fig:pose_reconstruct} (App.~\ref{app:more_results}), we show that the bottlenecks in \wyd{} stem from the inherent difficulty of generation, as the models' auto-encoders can significantly better reconstruct the videos.

\begin{figure}[t!]
  \centering
  \includegraphics[width=0.99\linewidth, trim={0 0 0 0}, clip]{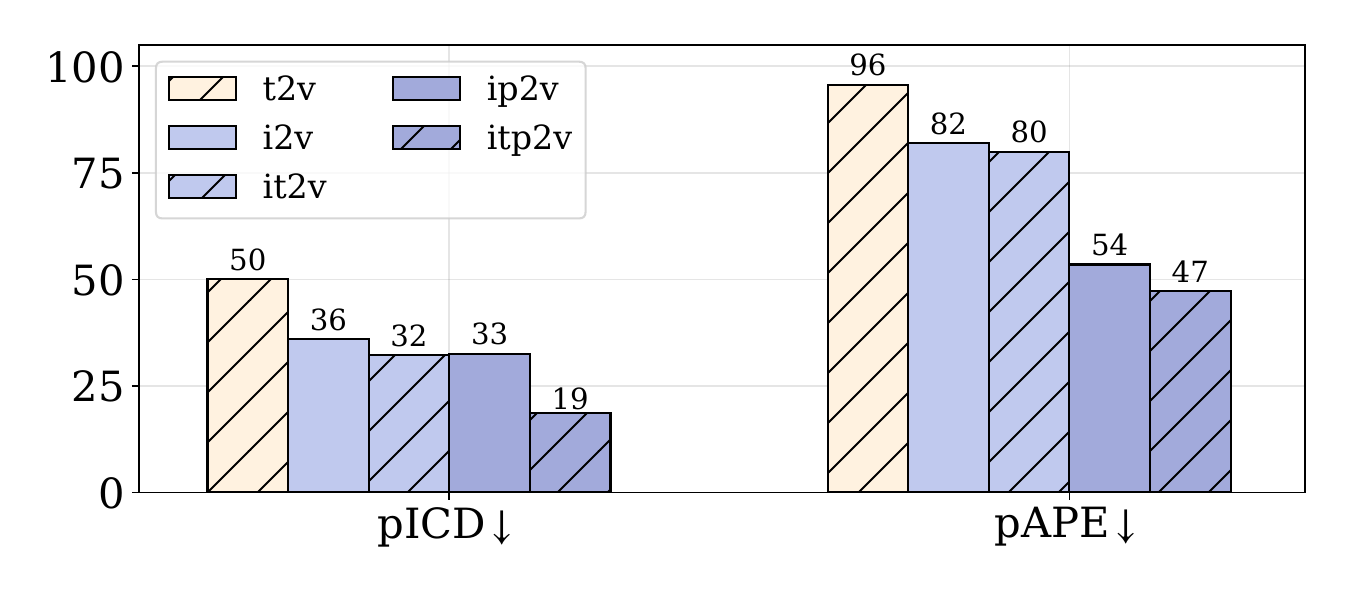}
  \vspace{-1.07\baselineskip}
  \caption{
  \textbf{Performance comparison when combining different conditioning signals}. Combining multiple signals (i: image, p: pose, t: text) improves \vace's performance on \wyd.
  }
  \label{fig:vace_ablations}%
  \vspace{-.7\baselineskip}
\end{figure}

\paragraph{Combining multiple modalities.}
We study the extent to which different modalities contribute to \vace (our only multi-task model) performance on \wyd.
As shown in \cref{fig:vace_ablations},~adding both pose and text guidance improves model performance.
As~expected,~pose is especially useful for better motion adherence, while detailed LLM-generated captions improve person consistency.
For dense-conditioned models, we see that \ctrladapter's generations also improve when adding text guidance~(see \cref{fig:text_deltas} in App.~\ref{app:more_results}).

\paragraph{Ablating pose signals.}
So far, we performed system-level evaluations above, using the pre- and post-processing used by the original authors. However, \magicpose uses the OpenPose~\citep{openpose} human pose estimator, whilst \mimicmotion and \controlnext all use the more performant DWPose~\citep{dwpose}, to generate pose conditioning signals from the reference video.
We find that replacing OpenPose with the more accurate DWPose detector improves \magicpose's performance, but the model still lags behind the remaining models (App.~\ref{app:more_results}).
Similarly, the relative ordering of models is unchanged when they all use OpenPose instead.
Therefore, our findings are consistent irrespective of the employed pose-detector.

Moreover, to investigate the sensitivity of our results to pose estimator inaccuracies, we manually annotated 100 videos (184 actors) with 2D body keypoints.
Our results in \cref{fig:pose_labels} (App.~\ref{app:more_results}) shows that the models' performance is largely unaffected, and relative performance unchanged.
This validates the correctness and robustness of our findings.

\subsection{Diagnosing model performance}

The categories we collected for \wyd{} are not only useful to ensure its richness of human videos, but can also be used to investigate model performance across these axes.
That is, we can apply our evaluation framework to different \emph{slices} of \wyd{} and gain insights into the strengths and weaknesses of models.
This information can be used to guide model development or, as we do next, to show specific limitations of our best models (\mimicmotion, \controlnext and \vace).

\begin{figure*}[t!]
  \centering
  \includegraphics[width=\linewidth]{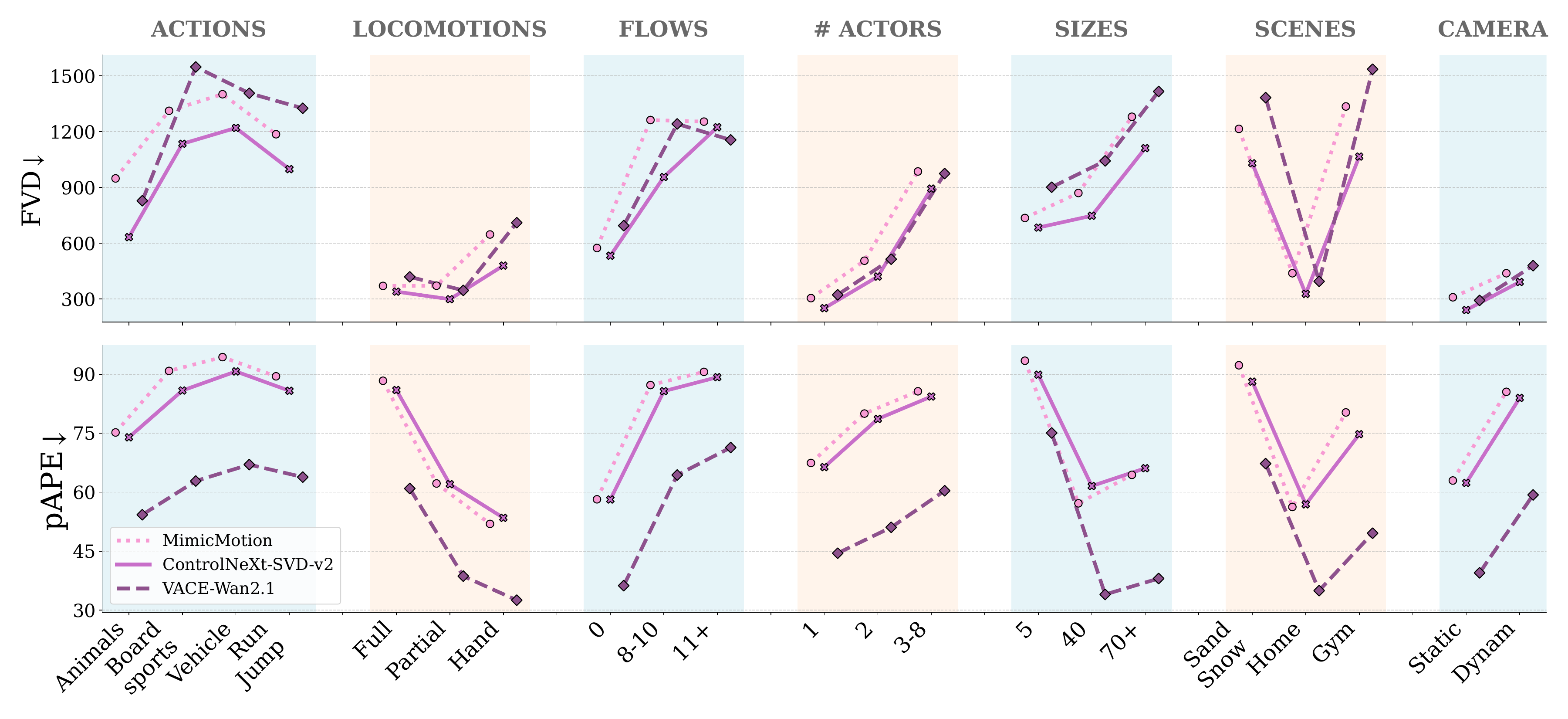}
  \vspace{-2\baselineskip}
  \caption{
  \textbf{Performance across multiple categories.} Existing models struggle with animating \emph{multiple people}, modeling complex \emph{interactions with animals}, in \emph{uncommon environments}, and within dynamic videos (\emph{camera motion}, \emph{full-body movement} and \emph{rapid actions} like running).
  }
  \label{fig:cat_perf}
\end{figure*}

\begin{figure*}[ht]
    \centering
    
    \begin{minipage}[b]{0.33\textwidth}
        \centering
        \includegraphics[width=\textwidth, trim={0 0 8.2cm 0}, clip]{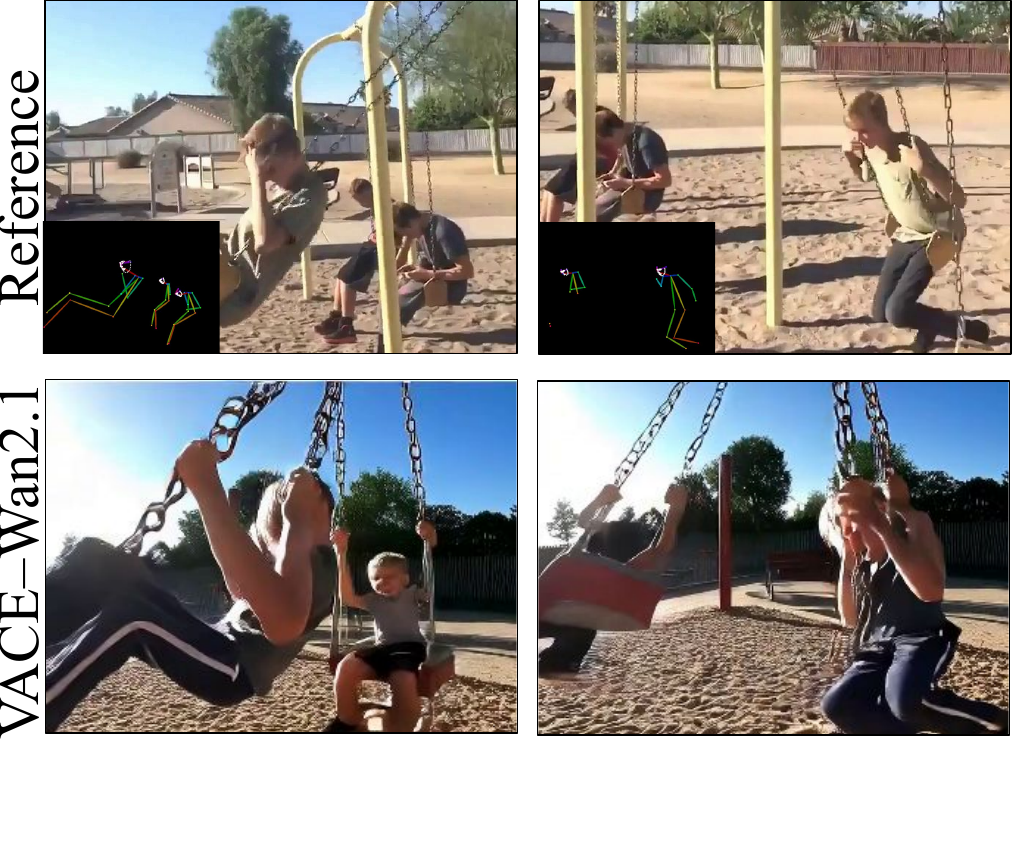}
        \vspace{-2.7\baselineskip}
        \captionof{figure}{Failure on multi-person video.}
        \label{fig:swing}
    \end{minipage}\hfill
    \begin{minipage}[b]{0.33\textwidth}
        \centering
        \includegraphics[width=\textwidth, trim={0 0 16.4cm 0}, clip]{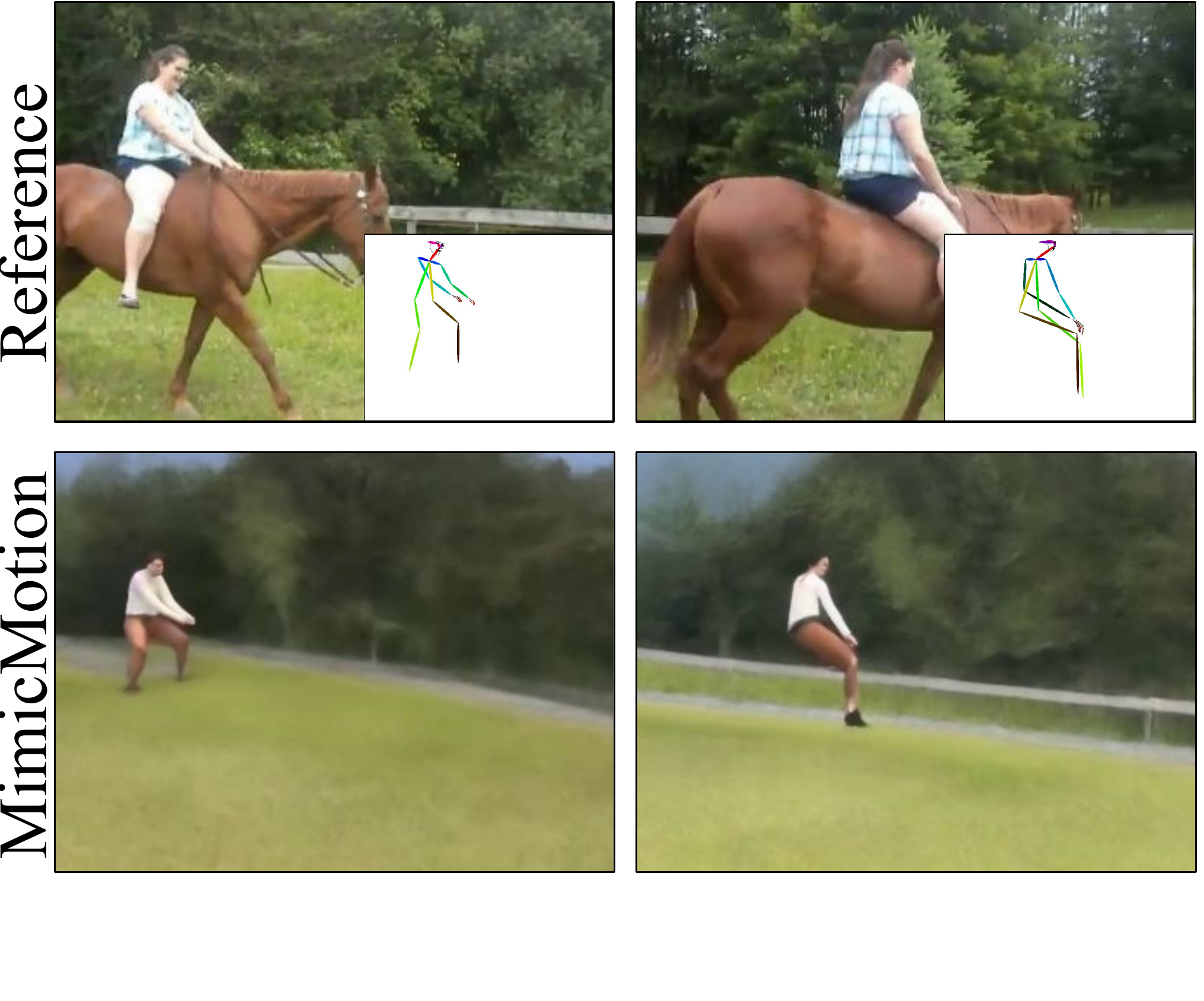}
        \vspace{-2.7\baselineskip}
        \captionof{figure}{Human–animal interaction failure.}
        \label{fig:horse}
    \end{minipage}\hfill
    \begin{minipage}[b]{0.33\textwidth}
        \centering
        \includegraphics[width=\textwidth, trim={0 0 16.4cm 0}, clip]{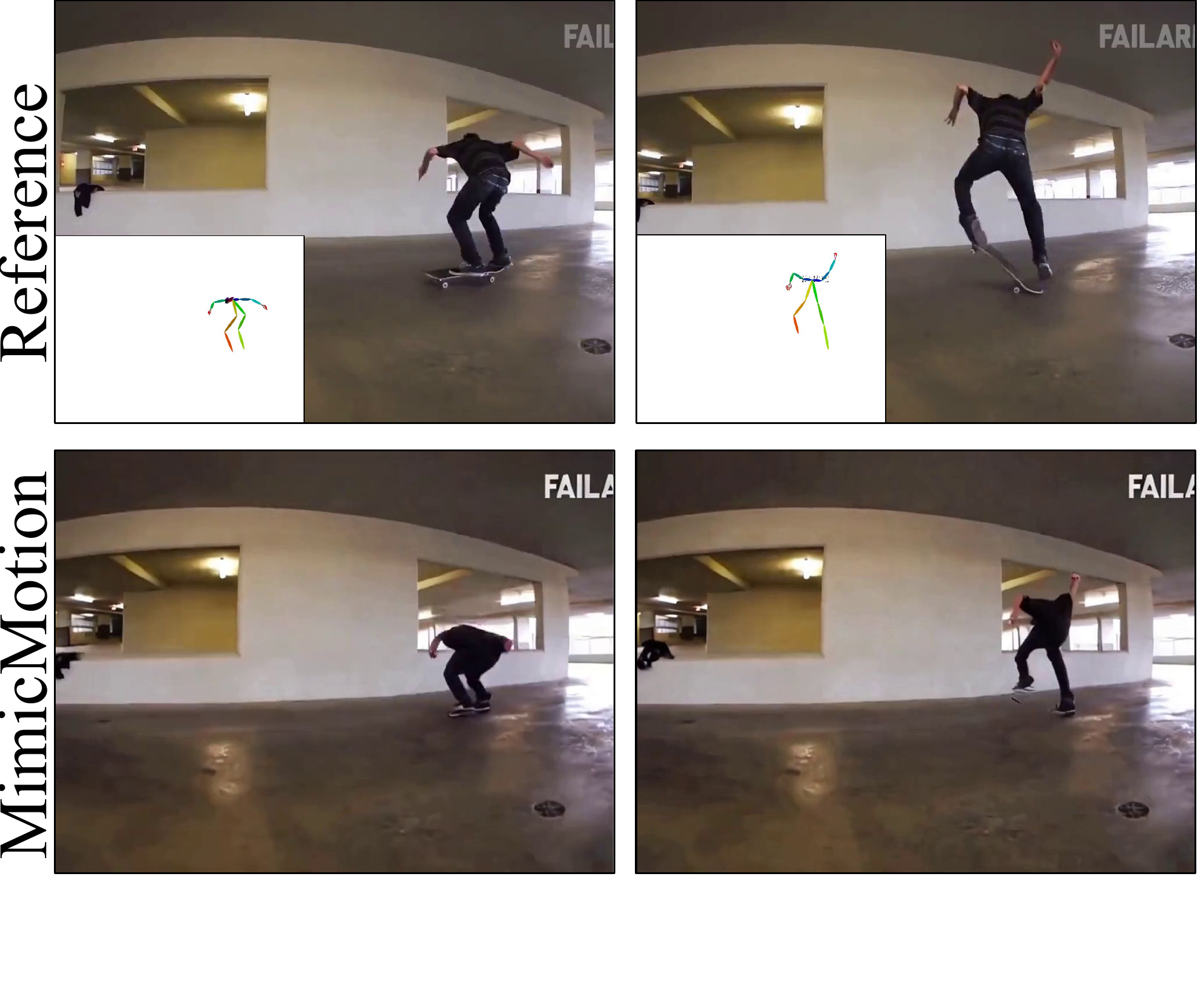}
        \vspace{-2.7\baselineskip}
        \captionof{figure}{Failure on dynamic video.}
        \label{fig:skating}
    \end{minipage}
    
    \vspace{-.5\baselineskip}
\end{figure*}

\paragraph{Towards multi-person videos.}
With \wyd{}, we enable evaluating model performance for controlling multiple humans for the first time.
As shown in \cref{fig:cat_perf}, both visual quality (FVD) and pose adherence (pAPE) degrade for \emph{2} and \emph{3+} actors.
Investigating the generations qualitatively, we find that models blend humans' appearances, especially when \emph{humans interact} with each other (as shown in \cref{fig:swing}).

\paragraph{Towards complex interactions.}
FVD shows that \mimicmotion struggles to generate high-quality videos of actions with \emph{animals}.
Inspecting the model's generations for this category, we see that the model tends to discard animals entirely, resulting in surreal generations (see \cref{fig:horse}).
This is likely because the model was mostly trained on close-up single-person videos (as confirmed by the authors; see~App.~\ref{app:samples}).

\paragraph{Towards diverse environments.}
Our results (\cref{fig:cat_perf}) show that all the models perform much better for videos taking place at \emph{home}.
Common, indoor environments (\eg, \emph{gym}) as well as more exotic, outdoor ones (\eg, \emph{sand and snow}) show large room for improvement for our benchmarked models.

\paragraph{Towards dynamic videos.}
Our metrics confirm that generating videos with high-level of motions are harder to generate for SOTA models.
In fact, performance worsen w.r.t. both FVD and pAPE when (i) the camera is \emph{not static}, (ii) the optical flow magnitude is larger (\eg, \emph{10\%}), (iii) the actor moves their entire body (\emph{full-body} locomotion), or (iv) they perform rapid actions (\eg, \emph{boardsports}, \emph{running} or \emph{riding a vehicle}).
This is exemplified in~\cref{fig:skating}.

\paragraph{Towards humans of varying sizes.} 
While current models are mostly trained on close-up videos, we can measure model performance across humans covering different scales in \wyd.
Human actors that cover larger portions of the video are easier to control (lower pAPE) but the resulting videos typically have lower quality (higher FVD; \cref{fig:cat_perf}), suggesting greater difficulty for human details.

\vspace{-.5\baselineskip}
\paragraph{}
All in all, the findings we obtain from our dataset and evaluation protocol are aligned with the experience that researchers and practitioners in the field have from using these models.
We believe that being able to verify them in \wyd{} is crucial, as it acts as an encouraging signal to track and improve performance across these axes with our benchmark.

\section{Validating automatic metrics}\label{sec:metrics}

Automatic evaluation metrics are instrumental in benchmarking a model's performance.
However, automatically evaluating different aspects of video generation is difficult:
While multiple metrics have been proposed~\citep{t2vscore,he2024videoscore,gecko}, many have been shown to have little correlation with human judgment in previous work~\citep{storybench,kirstain2023pick,lin2024evaluating}.

Therefore, a careful validation of our proposed evaluation framework (\cref{sec:wyd_metrics}), specifically in our context of controllable human video generation, is crucial to establishing its credibility.
To do so, we compare existing metrics for each key aspect of (human) video generation, and show that the metrics we chose reflect human preferences well. \cref{tab:sidexside} summarizes our results (for more details, see App.~\ref{app:human_evals}).

\paragraph{Validation setup.}
We assess the performance of different metrics by aligning them with human judgments in two settings.
First, we perform side-by-side model comparisons.
We use four templates to assess video quality, video motion, human quality, and people movements.
Second, we manually verify the model rankings given by the metrics against generated samples (see App.~\ref{app:human_evals} for details).

\begin{table}[t!]
  \setlength{\tabcolsep}{3.0pt}
  \smaller
  \centering
  \begin{tabular}{llr}
\toprule
    & \textbf{Metric} & \textbf{Performance [\%]} \\
\midrule
                                                                  & FVD             & \textbf{96.36}             \\
                                                                  & FID            & 22.24                      \\
\multirow{-4}{*}{\textbf{Video quality}}                          & JEDi            & \textbf{96.36}                      \\
\midrule
\band 
\cellcolor{gray!10}                                          & OFE             & \textbf{82.10}             \\
\band 
\multirow{-2}{*}{\cellcolor{gray!10}\textbf{Video motion}}   & DPT             & 67.37                      \\
\midrule
                                                                  & pICD             & \textbf{72.67}             \\
                                                                  & ICD             & 67.33             \\
                                                                  % & PSNR            & 59.04                      \\
                                                                  & RMSE            & 38.55                      \\
\multirow{-4}{*}{\textbf{Human consistency}}                          & SSIM            & 62.65                      \\
\midrule
\band 
\cellcolor{gray!10}                                          & pAPE            & \textbf{71.95}             \\
\band 
\multirow{-2}{*}{\cellcolor[HTML]{EFEFEF}\textbf{Human movement}} & pOFE            & 61.45 \\
\bottomrule
\end{tabular}
  \vspace{-0.4em}
  \caption{\small
  \textbf{Side-by-side (SxS) evaluations.} We report Spearman rank correlation for video quality, and pair-wise accuracy otherwise. Our selected metrics agree with human preferences from SxS studies.}
  \label{tab:sidexside}
\end{table}

\begin{figure}[t!]
  \centering
  \includegraphics[width=0.96\linewidth, trim={0cm 35cm 0cm 0}, clip]{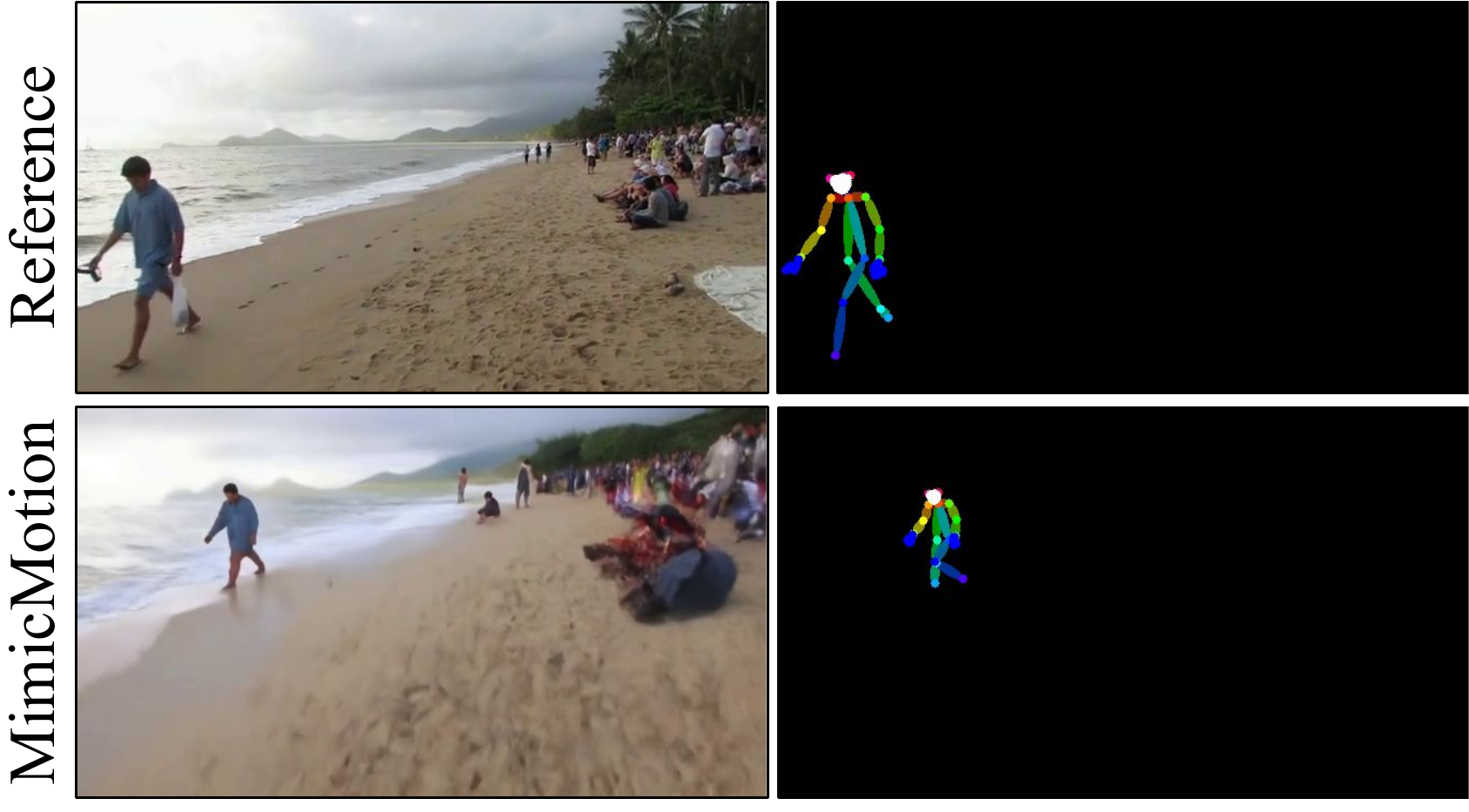}
  \caption{\small
  \textbf{Example of pose re-scaling in \mimicmotion.} The pose detected in the generated video (bottom right) is re-scaled and re-centered compared to the reference video's pose (top right). Humans are not typically sensitive to such changes, but pAPE is.}
  \label{fig:pose_rescaling}%
  \vspace{-.1\baselineskip}%
\end{figure}

\paragraph{Measuring human movement via poses.}
Pose AP is the gold standard metric to evaluate keypoint detection~\citep{coco}, and has previously been used for single-person generation~\citep{followyourpose}.
Our pose AP error (pAPE) adapts pose AP to robustly evaluate multi-person video generation by leveraging our collected segmentation and pose tracks (\cref{sec:wyd_metrics}).
\cref{tab:sidexside} shows that pAPE agrees well (72\%) on human evaluations of people's movements.
Moreover, pAPE allowed us to discover that \mimicmotion and \controlnext always re-scale and re-center the generated humans (\cref{fig:pose_rescaling}).
This is part of the models' pre-processing code~\citep{mimicmotion_code, controlnext_code}, and may stem from the overdependence on simpler datasets with only a single, large, centered actor.
This further emphasizes the need for more diverse benchmarks for controllable human generation.

\paragraph{pICD captures person consistency.}
We evaluate the ability of visually-controlled models to produce humans that closely resemble their ground-truth references.
\citet{consisti2v} use DINO~\citep{Caron_2021_ICCV} feature similarity across frames to measure subject consistency.
Inspired by their results, we adapt this metric (ICD) to measure human consistency by leveraging our collected segmentation masks (pICD;~\cref{sec:wyd_metrics}).
\cref{tab:sidexside} shows that pICD is 5\% more accurate than frame-wide ICD, and better captures person consistency according to humans.

\paragraph{FVD better measures video quality.}
We compare different metrics as proxies for overall video quality: FID~\citep{fid}, FVD, JEDi~\citep{jedi} and VMAF~\citep{vmaf}.
We measure Spearman rank correlation between the ranked human judgments and the scores of each metric.
We find that the widely used FID and SSIM metrics~\citep{animateanyone,magicpose} are inadequate, while FVD and JEDi strongly agree with humans.
We use FVD being faster.

\section{Conclusion}

Video generation has a tremendous potential to impact our society, and it is thus imperative to comprehensively assess model capabilities.
Our major contribution towards this end is the collection of \wyd: a large and comprehensive dataset to evaluate the synthesis of humans in real-world settings.
Our analysis showed that \wyd{} is more diverse and challenging than prior benchmarks for controllable human generation.
It is equipped with fine-grained categories that allowed us to discover several failure modes of SOTA controllable models, through both automatic and human evaluations.
By releasing \wyd, we aim to move beyond the current narrow scope of close-up single-person video generation, driving progress towards the ambitious goal of \emph{generic} human generation.

While our dataset can be used to benchmark a variety of tasks (including T2V and I2V), we focused on visually-conditioned image-to-video models in this work, as these methods are capable of precisely controlling human actors.

To evaluate these models, we leveraged the manually annotated segmentation masks in \wyd{} to adapt existing metrics for human-specific assessment: pICD measures the consistency of human appearance in a video, while pAPE captures fine-grained human movement.
Although we showed that these metrics have higher correlations with human preferences than previous ones, we next point out their current limitations as well as opportunities for future work.

Firstly, although DWPose offers great 2D pose detection accuracy, we found \emph{sometimes} that it makes incorrect predictions, especially for humans generated poor quality.
Therefore, it is key to develop more accurate and robust (video) pose estimators.
Such models could readily be used in our pAPE metric as they become available, and are instrumental to accurately transfer human motion across videos.

Secondly, while our \emph{video quality metric} suggests that it is easy to generate videos of human--object interactions, our qualitative analysis showed that objects may disappear.
As they typically cover a small percentage of the frame, metrics like FVD are largely unaffected by this, and do not reveal this issue.
While we focused on key metrics for human generation to introduce \wyd, we note that \wyd{} enables future research towards new and more nuanced metrics (\eg, for human--object interactions) in complex, real-world settings.

\clearpage
\section*{Acknowledgments}
We would like to thank Thomas Mensink, Jordi Pont-Tuset, Benoit Corda, Monika Wysoczańska, Thomas Kipf, Nikos Kolotouros, Paul Voigtlaender, David Ross, Miki Rubinstein, Tomáš Izo, Rahul Sukthankar, and the Veo team for fruitful discussions and support throughout this project.
{
    \small
    \bibliographystyle{ieeenat_fullname}
    \bibliography{main}
}

\clearpage
\appendix
\section*{Overview}

Our Appendix includes the following content.

\cref{app:samples} shows videos generated by our seven models.
\cref{fig:generation_samples_1_1,fig:generation_samples_1_2,fig:generation_samples_1_3,fig:generation_samples_2_1,fig:generation_samples_2_2,fig:generation_samples_2_3,fig:generation_samples_3_1,fig:generation_samples_3_2,fig:generation_samples_3_3,fig:generation_samples_4_1,fig:generation_samples_4_2,fig:generation_samples_4_3,fig:generation_samples_5_1,fig:generation_samples_5_2,fig:generation_samples_5_3} (pp. 19--33) show samples generated by all the models for five examples.\\

\begin{figure*}[t!]
    \centering
    \includegraphics[width=\linewidth, trim={0 2cm 0 0}, clip]{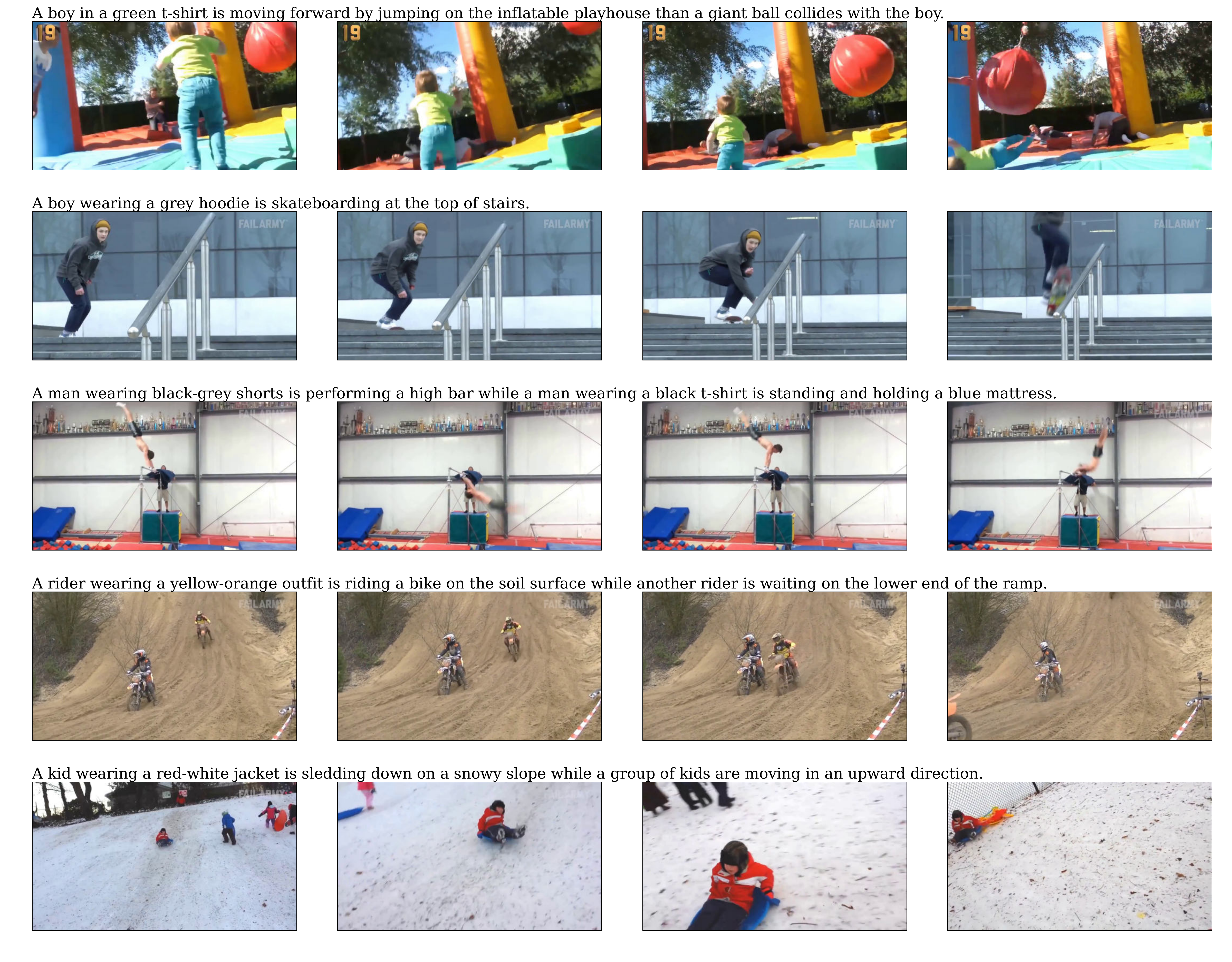}
    \caption{Samples from the final \wyd{} dataset.}
    \label{fig:drop_pipeline_8}
\end{figure*}

\cref{app:data_prep} (pp. 34--42) provides further details from our data preparation pipeline.
\cref{fig:wyd_pipeline} shows a high-level overview of the filtering process.
\cref{tab:wyd_filtering_size,tab:human_actors} list additional details of the dataset construction (p. 34), while \cref{fig:duration_distr,fig:resolution_distr} compare video duration and resolution across the \wyd, \tiktok and \tedtalks datasets (p. 38).
\cref{fig:drop_pipeline_1,fig:drop_pipeline_2,fig:drop_pipeline_3,fig:drop_pipeline_4,fig:drop_pipeline_5,fig:drop_pipeline_6,fig:drop_pipeline_7} (pp. 35--37) show sample videos that were removed as part of the filtering process, and \cref{fig:drop_pipeline_8} (p. 37) shows more examples from the final \wyd{} dataset.
\cref{fig:category_ui,fig:segmentation_ui,fig:pose_ui} (pp. 40--41) show our UIs for video categorization, segmentation and pose annotations.
\cref{fig:category_overlaps} displays the overlap in videos between any two categories.\\

\cref{app:more_results} reports additional results from our experiments.
\cref{fig:dense_overall} (p. 43) shows the overall performance of depth- and edge-conditioned models on \wyd$_{16}$.
\cref{fig:text_deltas} shows the difference in errors of depth- and edge-conditioned models when adding captions as an additional source of guidance.
\cref{fig:pose_reconstruct,fig:pose_detector,fig:pose_labels} (pp. 43--44) instead provide quantitative support for the ablations of pose-guided models discussed in \cref{sec:results}.
Moreover, \cref{fig:category_col1,fig:category_col2,fig:category_col3,fig:category_col4} (pp. 44--46) report and discuss category-level performance of our top-performing models (\mimicmotion, \controlnext and \vace).\\

\cref{app:human_evals} includes further details about our human evaluation protocols.
We report our instructions and setup for side-by-side human evaluations in p. 47, and show our UI in \cref{fig:sidexside_ui} (p. 46).
\cref{fig:video_quality,fig:video_motion,fig:people_motion} (pp. 47--48) present and discuss how metrics that we considered to measure different aspects of video generation score our evaluated models.\\

Finally, we share some ethical considerations related to controllable human video generation in \cref{app:ethics} (p. 48), where we additionally remark that our \wyd{} dataset is meant to be used for academic research purposes only.

\newpage
\section{Samples of generated videos}\label{app:samples}

\begin{table*}[t!]
  \setlength{\tabcolsep}{1.0pt}
  \smaller
  \centering
  % \vspace{-1\baselineskip}
  \resizebox{\linewidth}{!}{
    \begin{tabular}{lccllcccccllcccccl}
\toprule
\textbf{Model} & & & \textbf{Condition} & \textbf{Extractor} & & & & & & \textbf{Training data} & \textbf{Close-up single-person videos?} & & & & & & \textbf{WYD overlap?} \\
\midrule
MagicAnimate~\citep{magicanimate}       & & & Dense pose    & Detectron2     & & & & & & TikTok~\citep{tiktok}            & Yes & & & & & & No  \\
MagicPose~\citep{magicpose}             & & & 2D pose       & OpenPose       & & & & & & TikTok~\citep{tiktok}            & Yes & & & & & & No  \\
MimicMotion~\citep{mimicmotion}         & & & 2D pose       & DWPose         & & & & & & Internal          & Yes & & & & & & No  \\
ControlNeXt-SVD-v2~\citep{controlnext}  & & & 2D pose       & DWPose         & & & & & & Internal               & No & & & & & & N/A \\
VACE-Wan2.1~\citep{vace}  & & & 2D pose       & DWPose         & & & & & & Internal               & N/A & & & & & & N/A \\
Control-A-Video~\citep{controlavideo}   & & & Depth / Canny & MiDaS / OpenCV & & & & & & WebVid~\citep{webvid10m} subset* + internal & No  & & & & & & No  \\
TF--T2V~\citep{tft2v}                   & & & Depth         & MiDaS          & & & & & & WebVid~\citep{webvid10m} subset* + internal & No  & & & & & & No  \\
Ctrl-Adapter~\citep{ctrladapter}        & & & Depth / Canny & MiDaS / OpenCV & & & & & & Panda-70M~\citep{panda70m}         & No  & & & & & & No  \\
\bottomrule
\end{tabular}
  }
  \caption{\textbf{Overview of evaluated models.} We list models' conditions and used extractors, their training data and whether it mostly consists of close-up single-person videos, and whether any of the video datasets used in \wyd{} were used by them during training. We thank the authors for clarifying information about their training data and confirming the absence of overlap with our evaluation videos. * Note that different models rely on different subsets of WebVid.}
  \label{tab:models}
\end{table*}

An overview of the evaluated models is shown in \cref{tab:models}.
\cref{fig:generation_samples_1_1,fig:generation_samples_1_2,fig:generation_samples_1_3,fig:generation_samples_2_1,fig:generation_samples_2_2,fig:generation_samples_2_3,fig:generation_samples_3_1,fig:generation_samples_3_2,fig:generation_samples_3_3,fig:generation_samples_4_1,fig:generation_samples_4_2,fig:generation_samples_4_3,fig:generation_samples_5_1,fig:generation_samples_5_2,fig:generation_samples_5_3} show and discuss the limitations of samples generated by all the models for five \wyd{} examples.

\begin{figure*}[t]
  \centering
  \includegraphics[width=0.95\linewidth, trim={0cm 0cm 0 0}, clip]{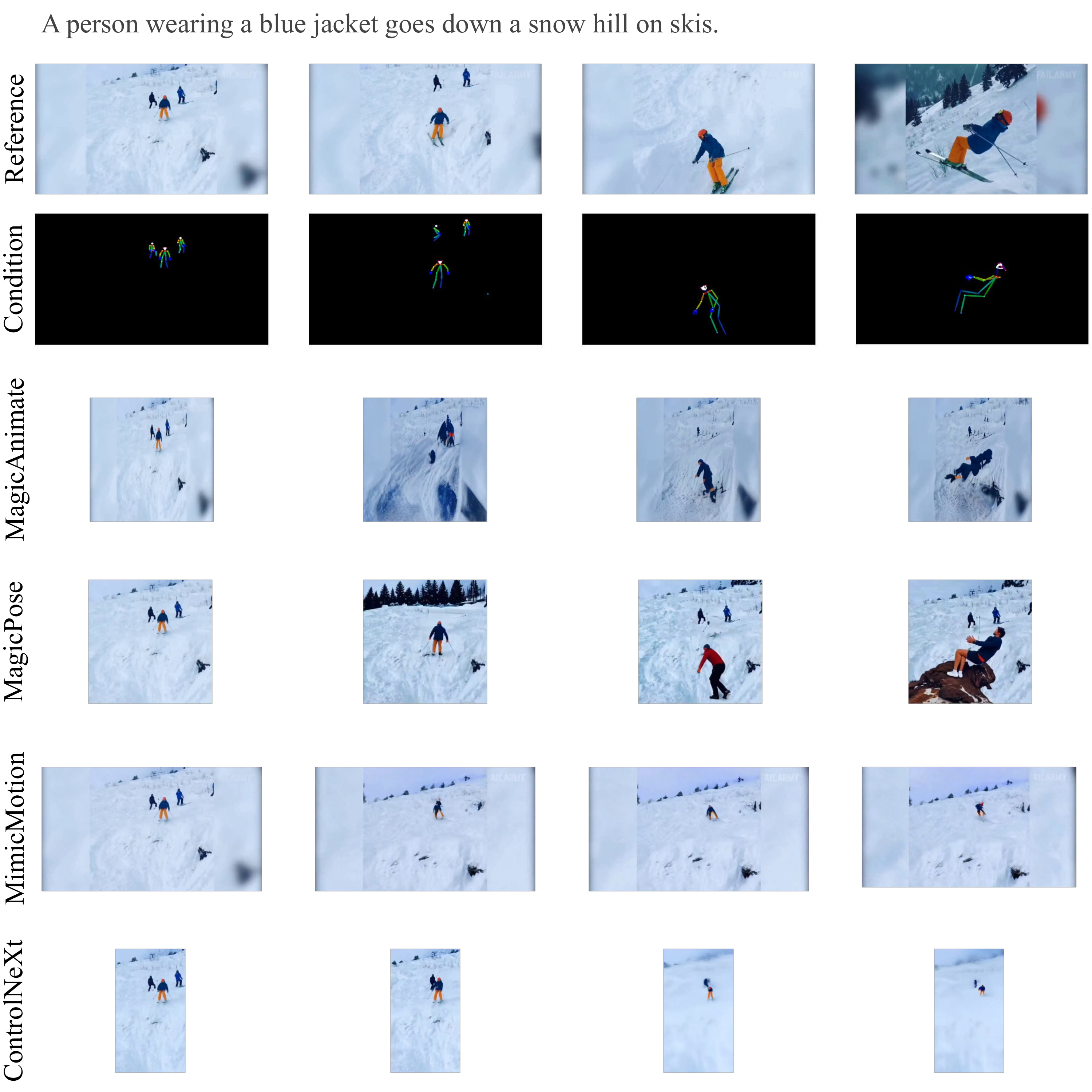}
  \includegraphics[width=0.95\linewidth, trim={0cm 0cm 0 0}, clip]{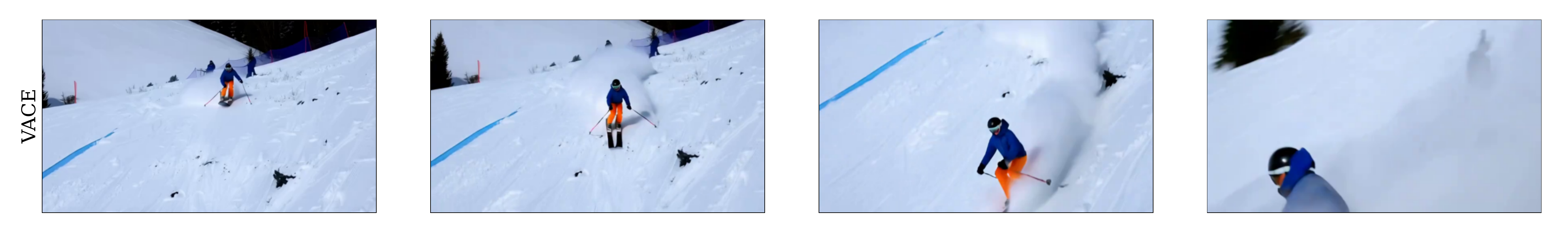}
  \caption{
  Example generations of our evaluated pose-conditioned models (\magicanimate uses dense poses).
  We can see how people's appearance changes in \magicpose, although matching the human movements the best.
  We can also see the size mismatches in \controlnext and \mimicmotion.
  }
  \label{fig:generation_samples_1_1}%
\end{figure*}
\begin{figure*}[t]
  \centering
  \includegraphics[width=0.95\linewidth, trim={0cm 7cm 0 0}, clip]{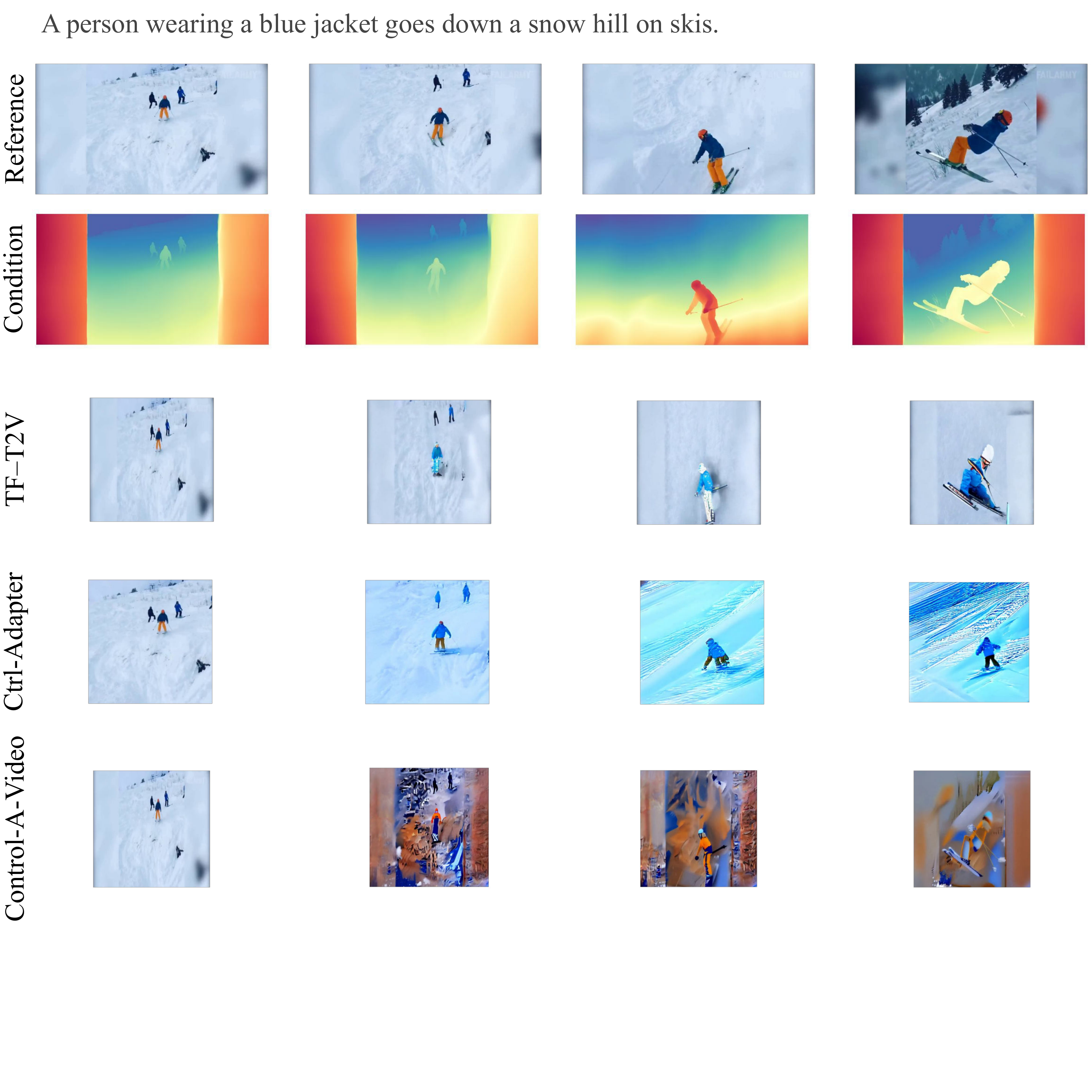}
  \vspace{-0.5\baselineskip}
  \caption{
  Example generations of our evaluated depth-conditioned models.
  We can see how people's appearance changes in \tftv, increasing saturation in \ctrladapter and distortions in \cavideo.
  }
  \label{fig:generation_samples_1_2}%
\end{figure*}
\begin{figure*}[t]
  \centering
  \includegraphics[width=0.95\linewidth, trim={0cm 16cm 0 0}, clip]{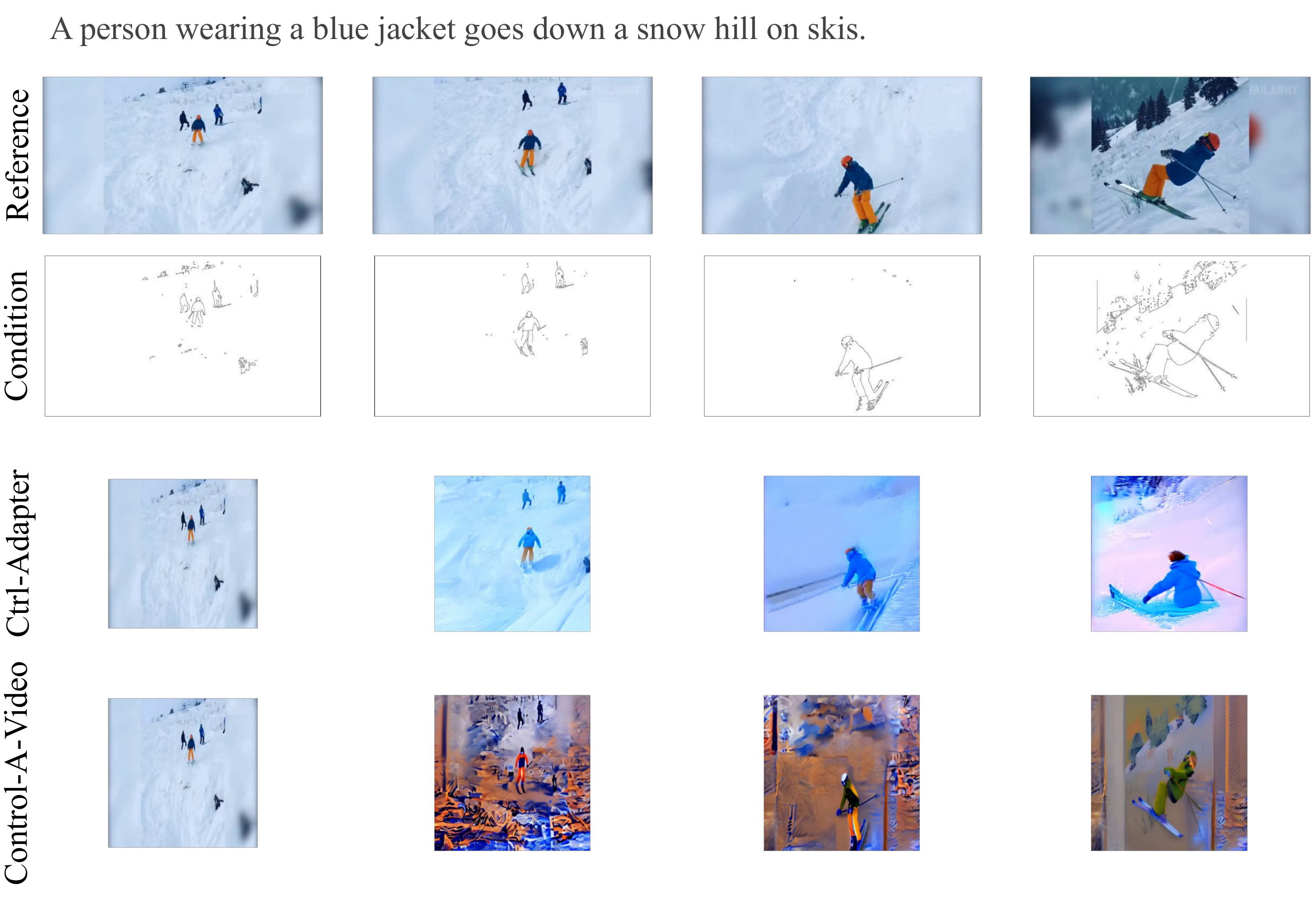}
  \vspace{-0.5\baselineskip}
  \caption{
  Example generations of our evaluated edge-conditioned models.
  We can see increasing saturation in \ctrladapter and distortions in \cavideo.
  }
  \label{fig:generation_samples_1_3}%
\end{figure*}

\begin{figure*}[t]
  \centering
  \includegraphics[width=0.95\linewidth, trim={0cm 0cm 0 0}, clip]{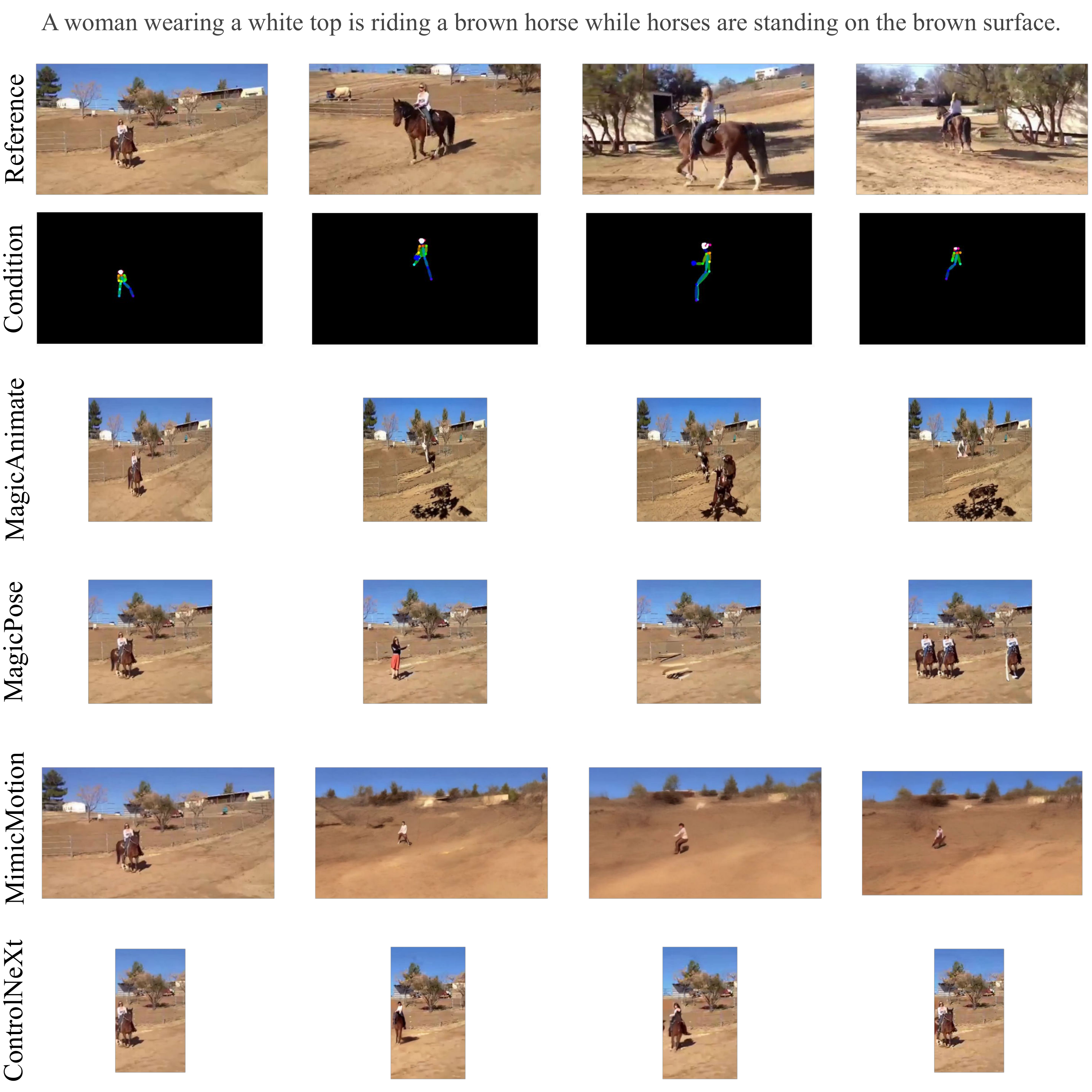}
  \includegraphics[width=0.95\linewidth, trim={0cm 0cm 0 0}, clip]{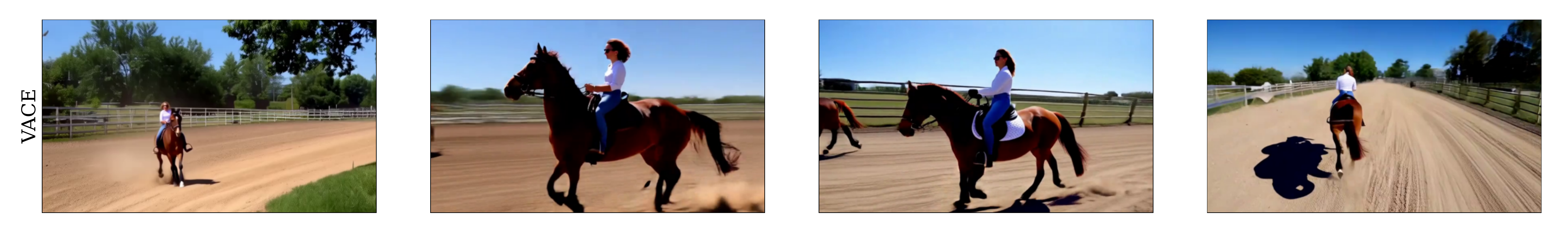}
  \caption{
  Example generations of our evaluated pose-conditioned models (\magicanimate uses dense poses).
  We note the challenges in camera motion for all models, the distortions of characters in \magicanimate, and flickering effects in \magicpose, as well as horse disappearance in \mimicmotion.
  }
  \label{fig:generation_samples_2_1}%
\end{figure*}
\begin{figure*}[t]
  \centering
  \includegraphics[width=0.95\linewidth, trim={0cm 7cm 0 0}, clip]{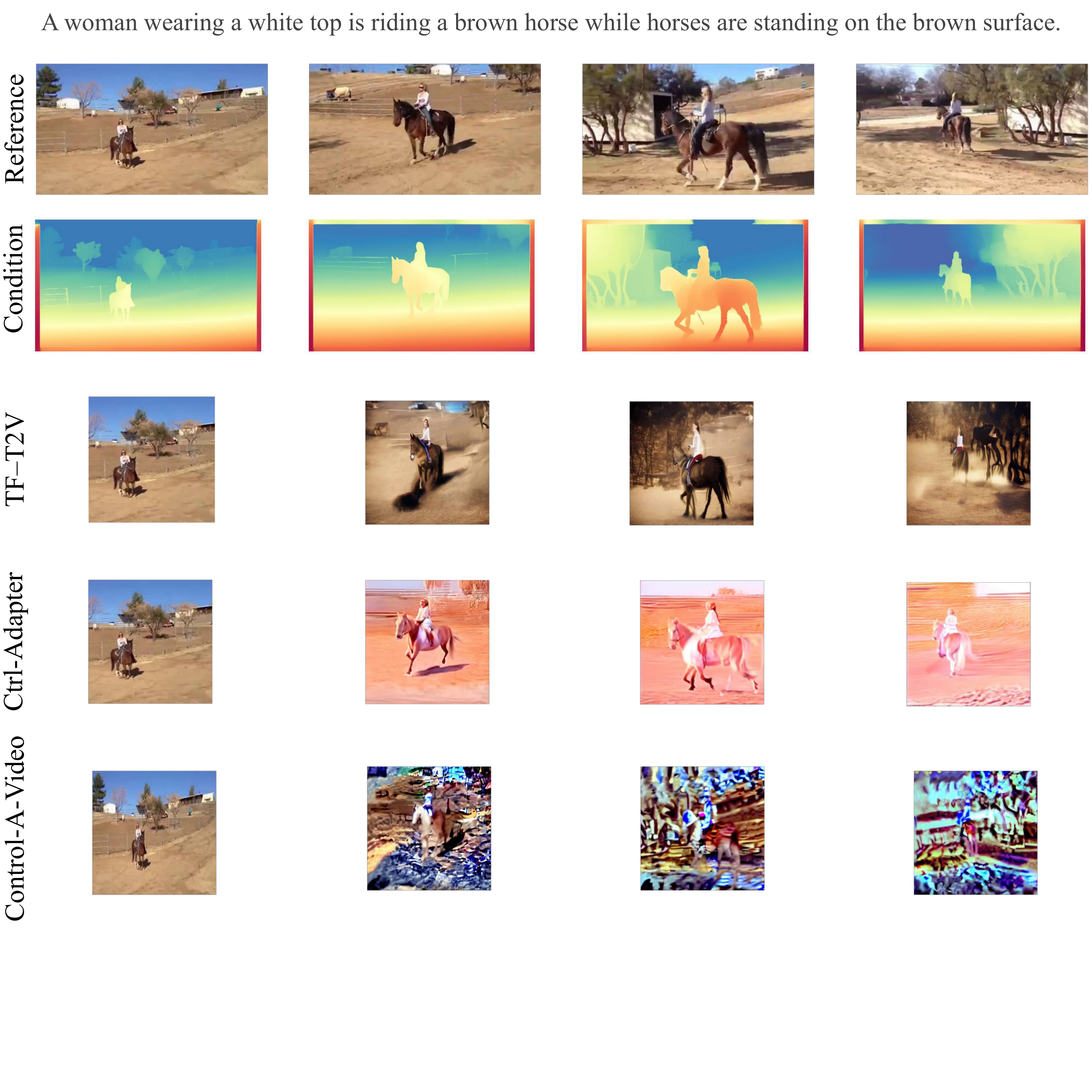}
  \vspace{-0.5\baselineskip}
  \caption{
  Example generations of our evaluated depth-conditioned models.
  We can see increasing saturation in \ctrladapter and distortions in \cavideo, while \tftv best matches the overall scene.
  }
  \label{fig:generation_samples_2_2}%
\end{figure*}
\begin{figure*}[t]
  \centering
  \includegraphics[width=0.95\linewidth, trim={0cm 16cm 0 0}, clip]{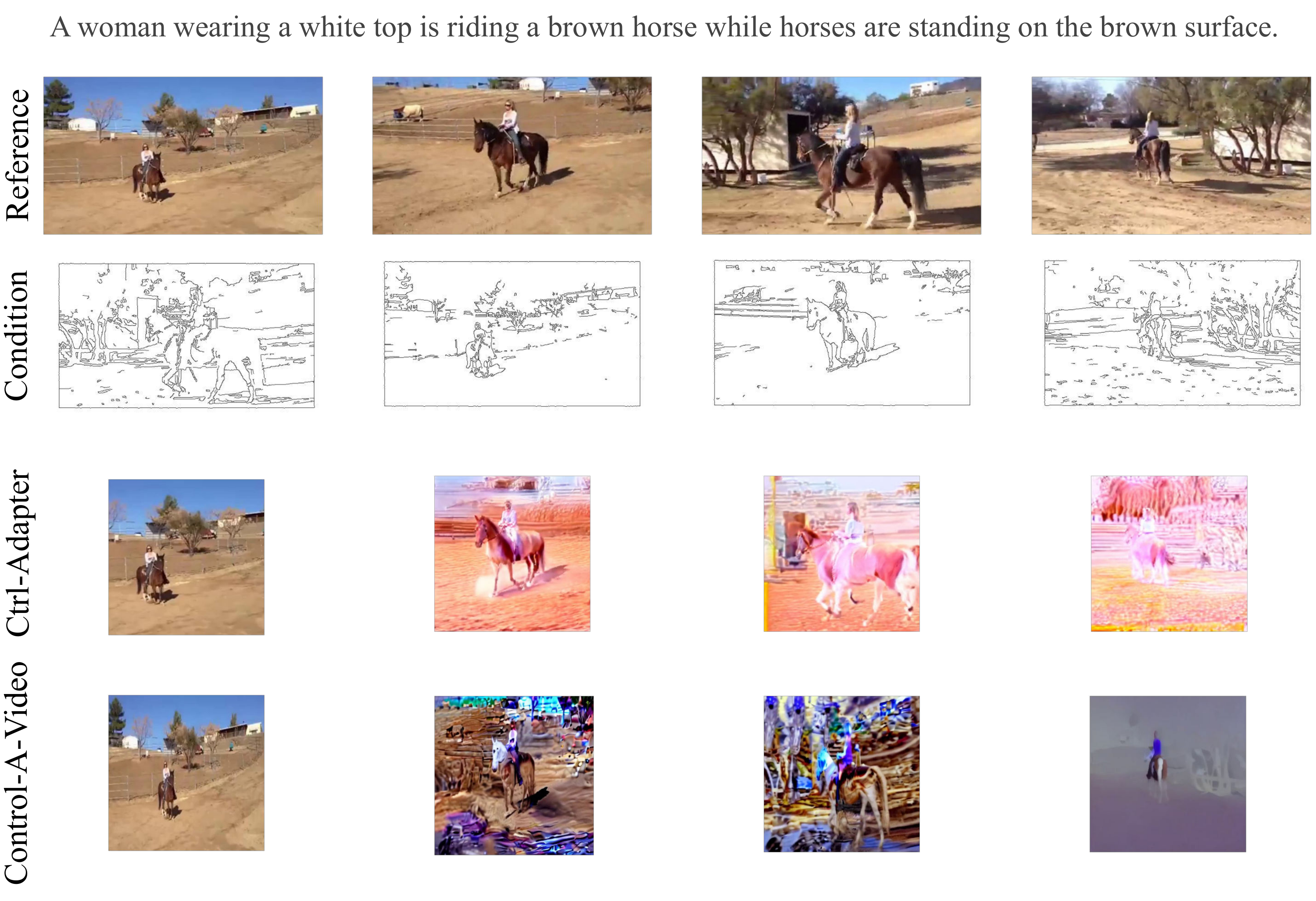}
  \vspace{-0.5\baselineskip}
  \caption{
  Example generations of our evaluated edge-conditioned models.
  We can see increasing saturation in \ctrladapter and distortions in \cavideo.
  }
  \label{fig:generation_samples_2_3}%
\end{figure*}

\begin{figure*}[t]
  \centering
  \includegraphics[width=0.95\linewidth, trim={0cm 0cm 0 0}, clip]{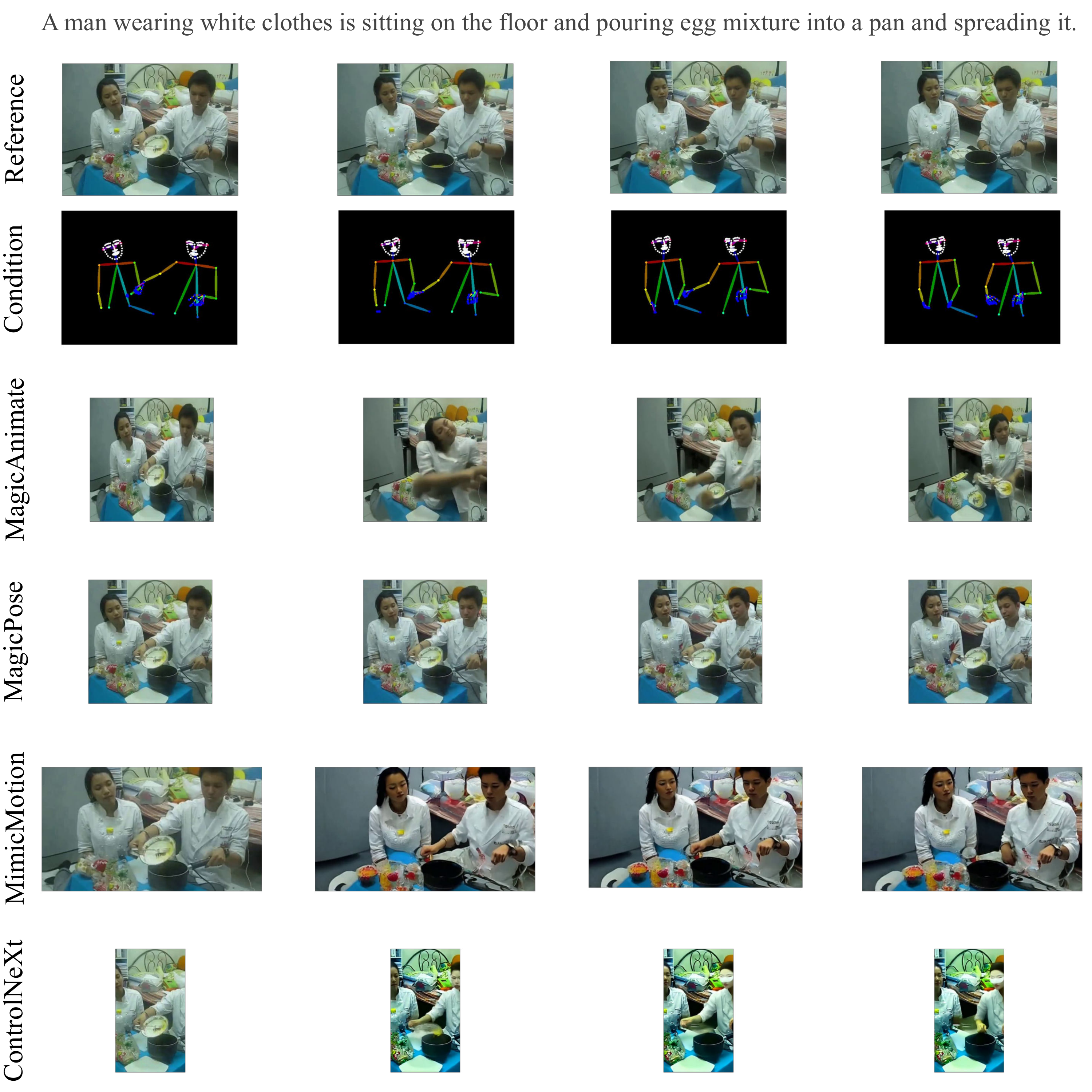}
  \includegraphics[width=0.95\linewidth, trim={0cm 0cm 0 0}, clip]{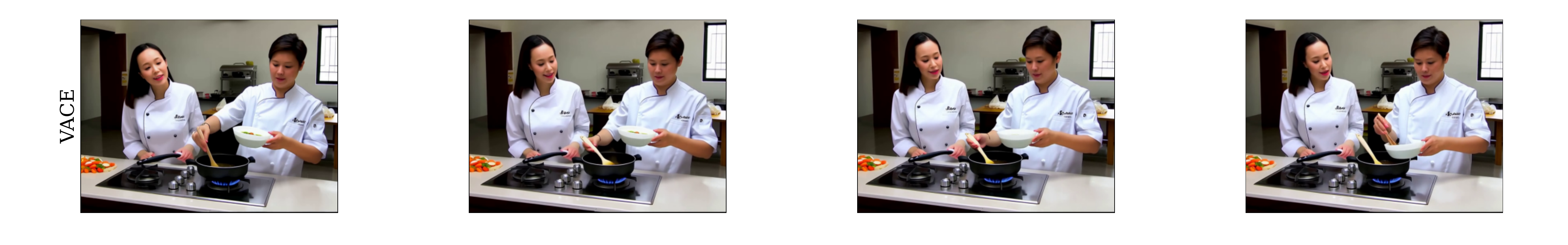}

  \caption{
  Example generations of our evaluated pose-conditioned models (\magicanimate uses dense poses).
  We can see how \mimicmotion changes the facial traits of humans towards specific age and beauty standards, and how it also fails to make the man interact with the pan.
  Due to its pre-processing, \controlnext misses the face of the man in the first frame and later creates a different one.
  }
  \label{fig:generation_samples_3_1}%
\end{figure*}
\begin{figure*}[t]
  \centering
  \includegraphics[width=0.95\linewidth, trim={0cm 7cm 0 0}, clip]{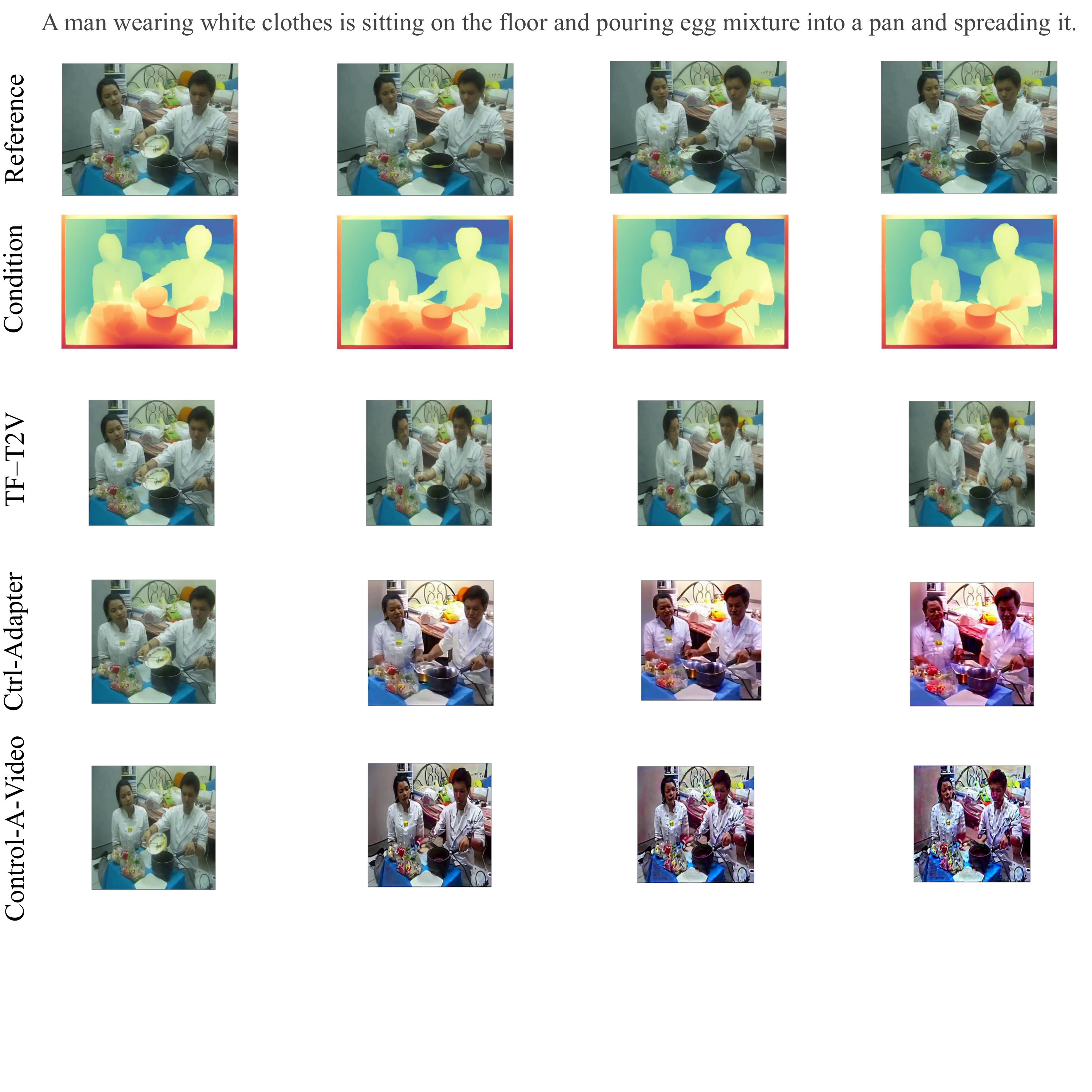}
  \vspace{-0.5\baselineskip}
  \caption{
  Example generations of our evaluated depth-conditioned models.
  We can see how \ctrladapter change the facial traits of humans towards specific age and beauty standards.
  We can still see increasing saturation in \ctrladapter and distortions in \cavideo, although less than in previous, dynamic examples.
  }
  \label{fig:generation_samples_3_2}%
\end{figure*}
\begin{figure*}[t]
  \centering
  \includegraphics[width=0.95\linewidth, trim={0cm 16cm 0 0}, clip]{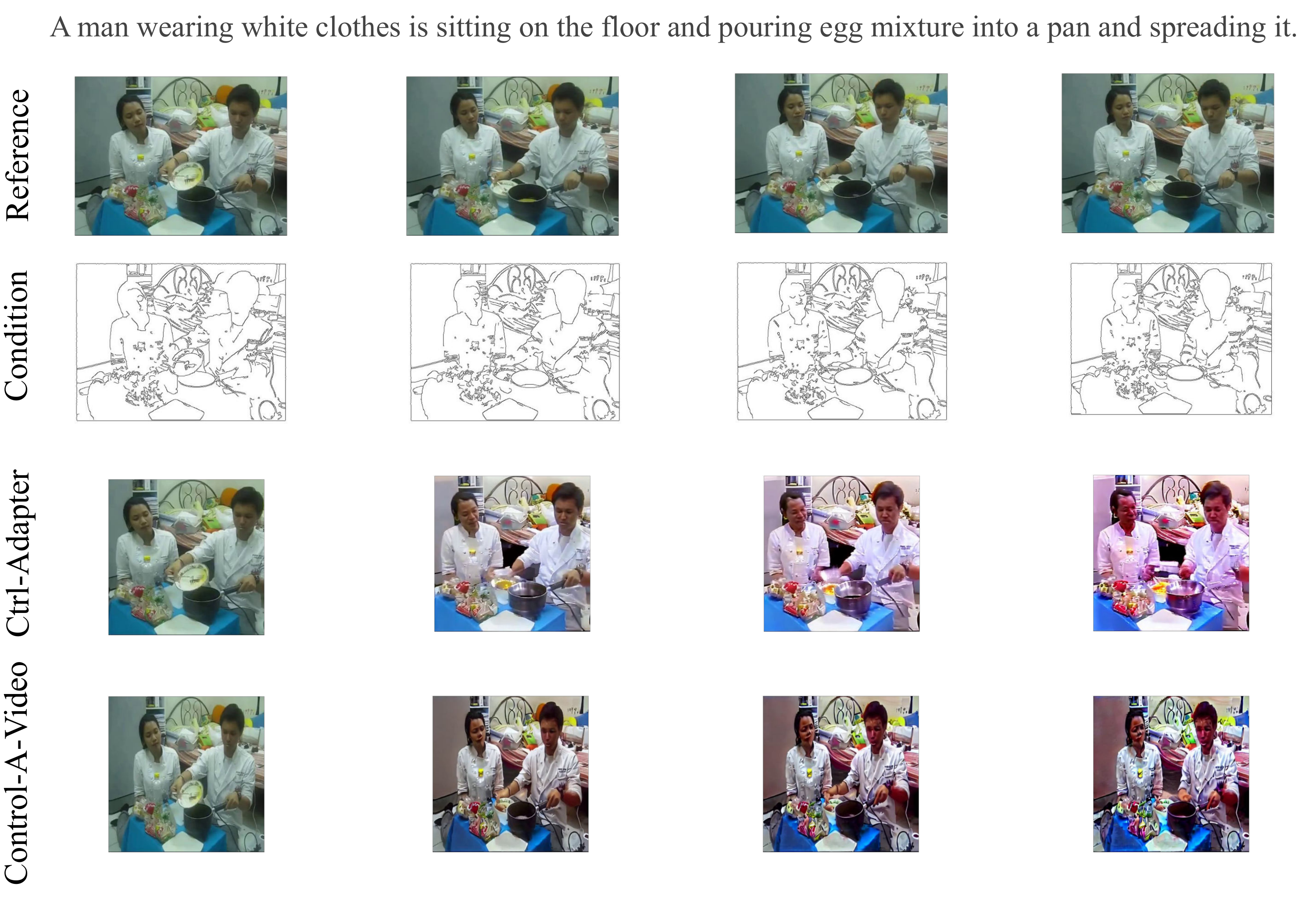}
  \vspace{-0.5\baselineskip}
  \caption{
  Example generations of our evaluated edge-conditioned models.
  Similar to their depth-conditioned counterparts, \ctrladapter shows increasing saturation and \cavideo presents distortions.
  }
  \label{fig:generation_samples_3_3}%
\end{figure*}

\begin{figure*}[t]
  \centering
  \includegraphics[width=0.95\linewidth, trim={0cm 0cm 0 0}, clip]{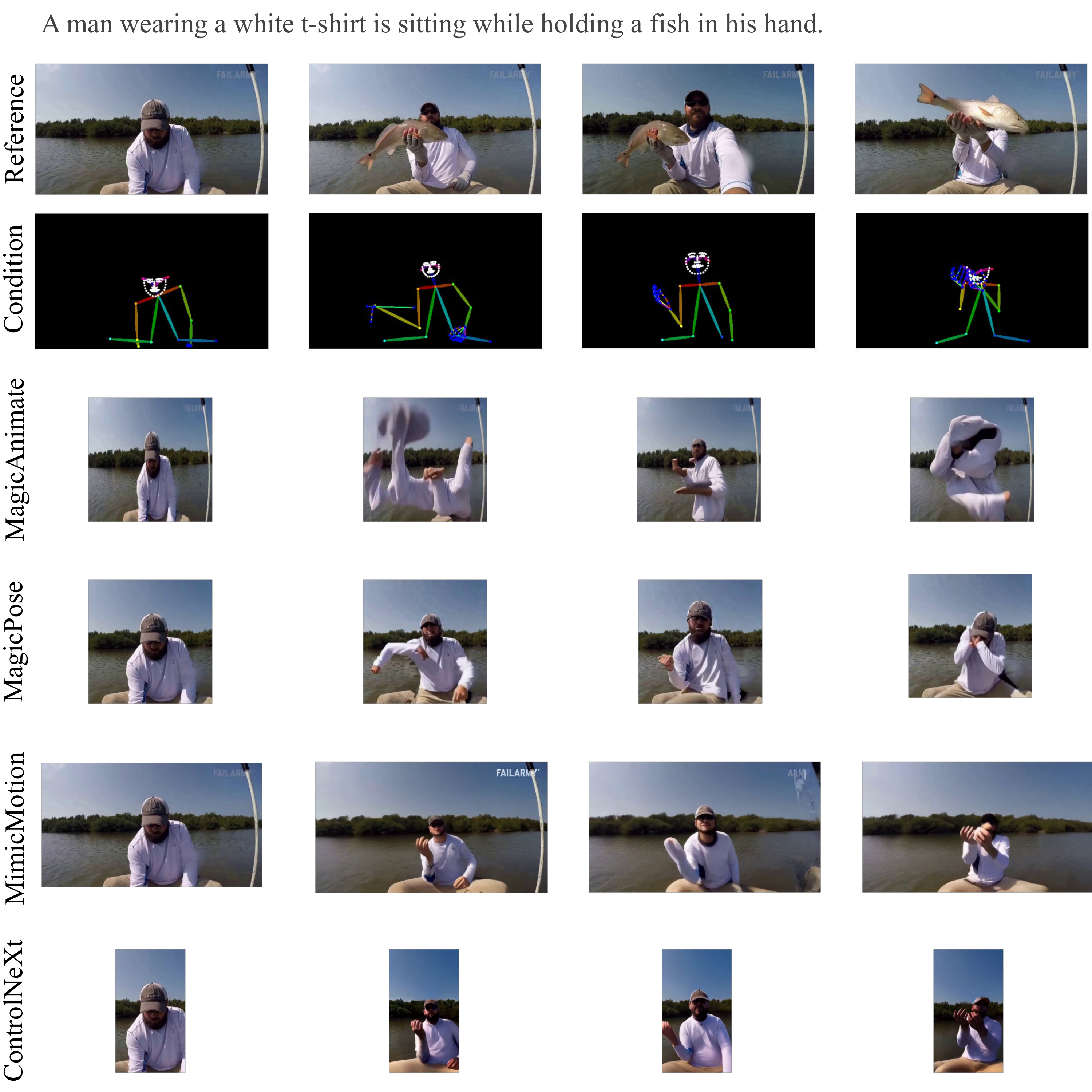}
  \includegraphics[width=0.95\linewidth, trim={0cm 0cm 0 0}, clip]{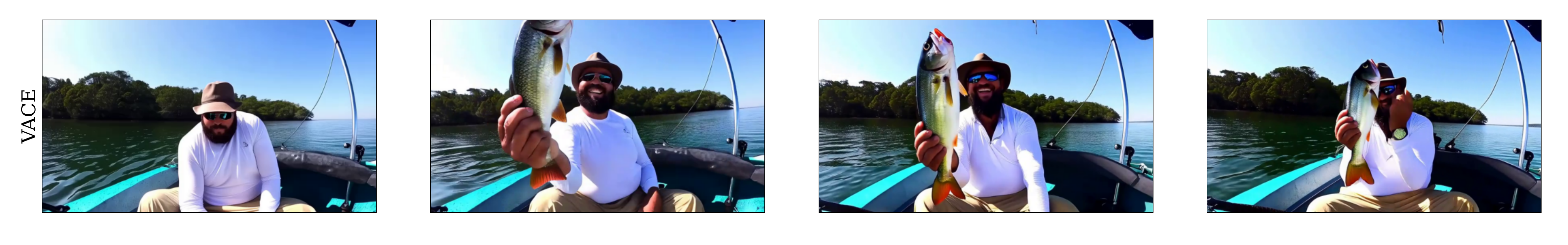}
  \caption{
  Example generations of our evaluated pose-conditioned models (\magicanimate uses dense poses).
  Besides the large distortions and artifacts in \magicanimate, all pose-conditioned models fail to generate the fish.
  \mimicmotion again changes the man's appearance by removing his beard.
  }
  \label{fig:generation_samples_4_1}%
\end{figure*}
\begin{figure*}[t]
  \centering
  \includegraphics[width=0.95\linewidth, trim={0cm 7cm 0 0}, clip]{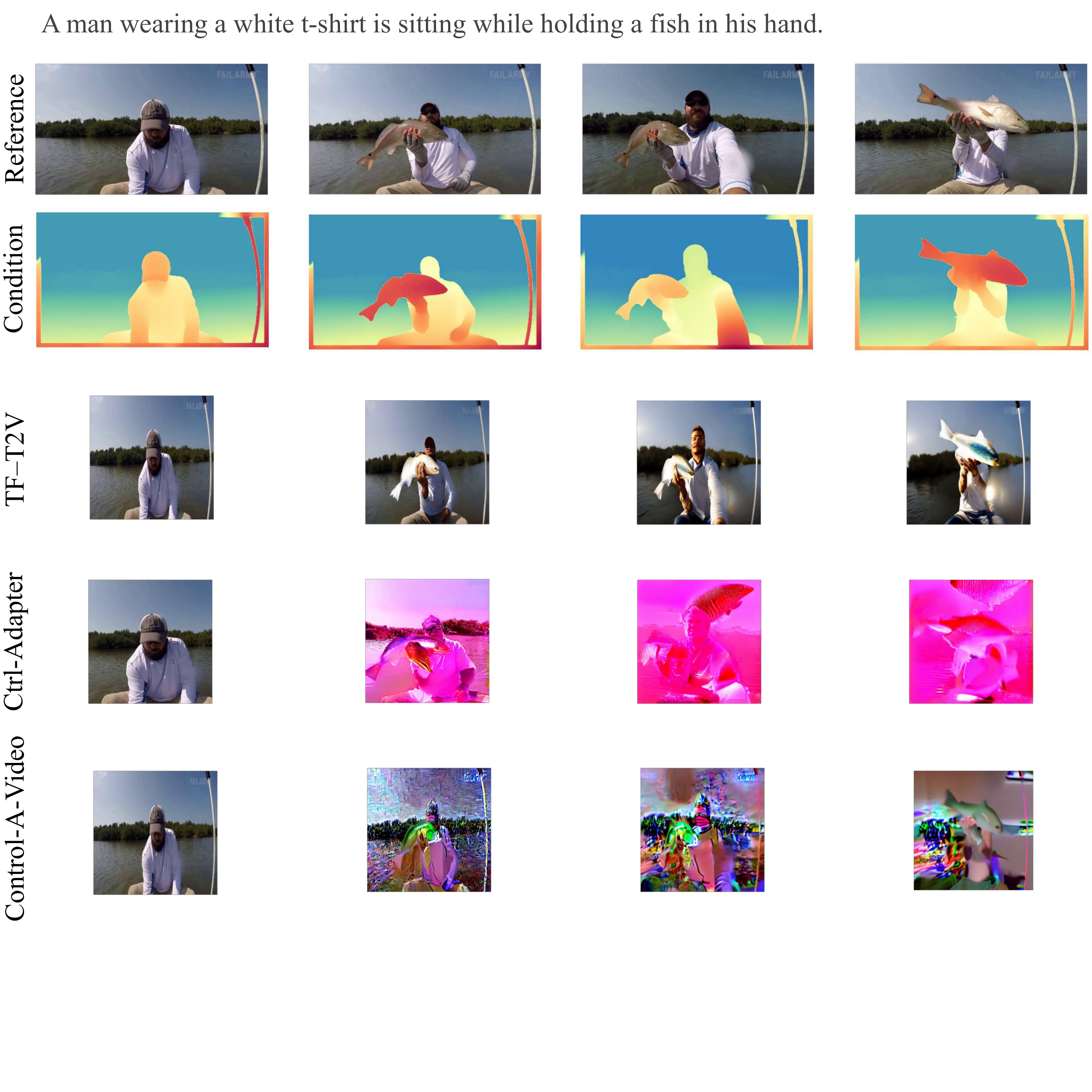}
  \vspace{-0.5\baselineskip}
  \caption{
  Example generations of our evaluated depth-conditioned models.
  We see high levels of saturation and distortions in \ctrladapter and \cavideo, respectively.
  \tftv is the only model that synthesizes the fish well.
  }
  \label{fig:generation_samples_4_2}%
\end{figure*}
\begin{figure*}[t]
  \centering
  \includegraphics[width=0.95\linewidth, trim={0cm 16cm 0 0}, clip]{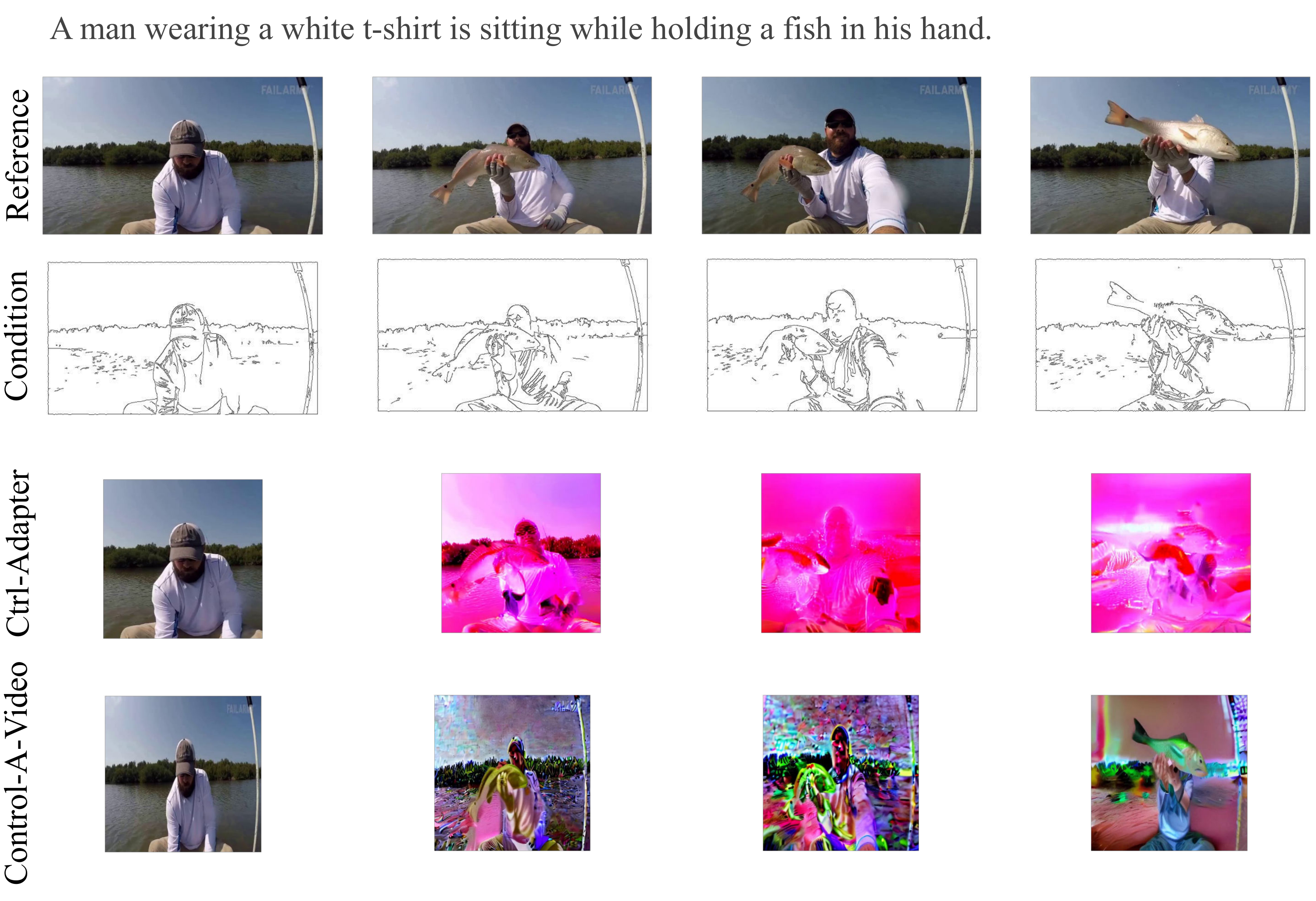}
  \vspace{-0.5\baselineskip}
  \caption{
  Example generations of our evaluated edge-conditioned models.
  \ctrladapter and \cavideo generates videos with very high levels of saturation and distortion.
  }
  \label{fig:generation_samples_4_3}%
\end{figure*}

\begin{figure*}[t]
  \centering
  \includegraphics[width=0.95\linewidth, trim={0cm 0cm 0 0}, clip]{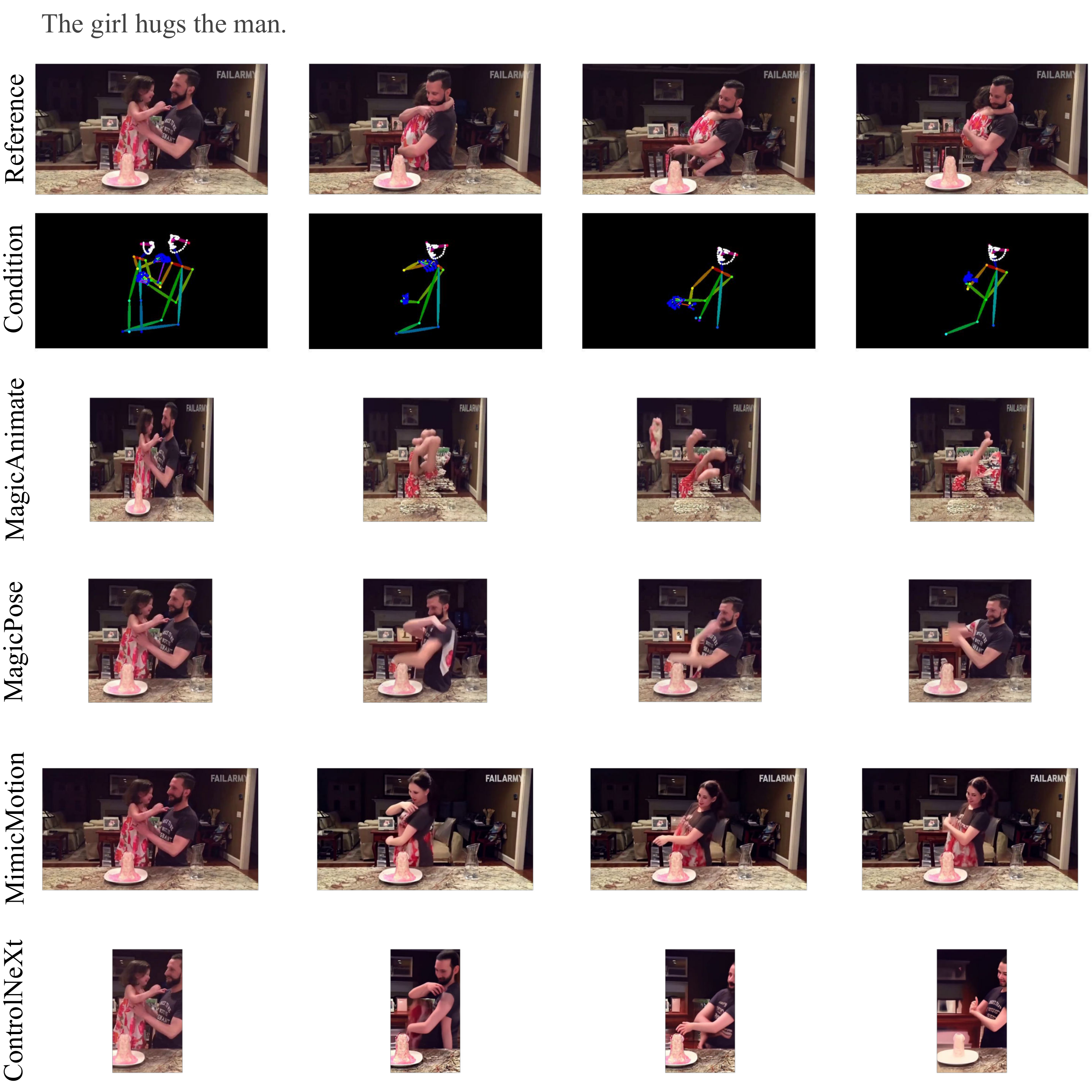}
  \includegraphics[width=0.95\linewidth, trim={0cm 0cm 0 0}, clip]{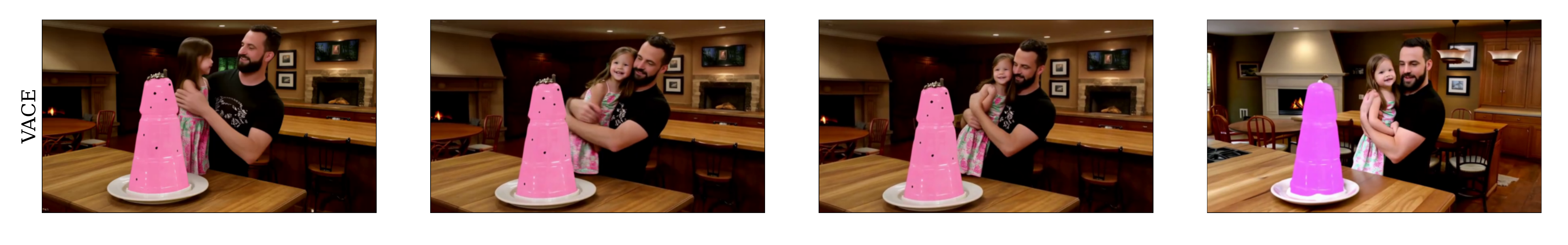}
  \caption{
  Example generations of our evaluated pose-conditioned models (\magicanimate uses dense poses).
  All models struggle with generating multiple humans interacting with each other consistently due to the limitations of the pose extractor.
  For example,  \magicpose and \controlnext make the girl disappear, while \mimicmotion merges the girl and the man into a woman.
  }
  \label{fig:generation_samples_5_1}%
\end{figure*}
\begin{figure*}[t]
  \centering
  \includegraphics[width=0.95\linewidth, trim={0cm 7cm 0 0}, clip]{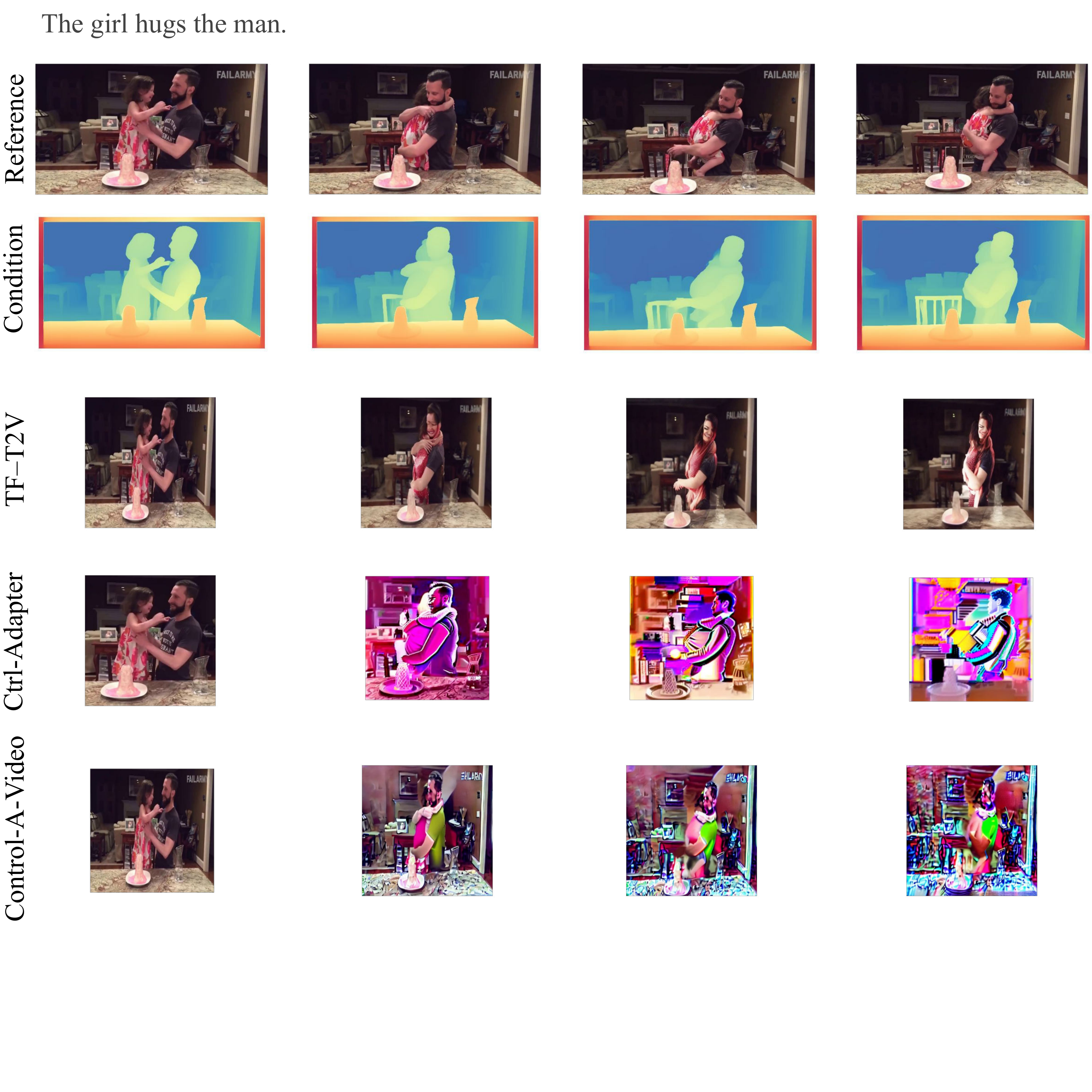}
  \vspace{-0.5\baselineskip}
  \caption{
  Example generations of our evaluated depth-conditioned models.
  We can see that even \tftv struggles to make the two humans interact with each other, resulting in a video where the man and the girl are merged into a single person with low-quality facial traits.
  }
  \label{fig:generation_samples_5_2}%
\end{figure*}
\begin{figure*}[t]
  \centering
  \includegraphics[width=0.95\linewidth, trim={0cm 16cm 0 0}, clip]{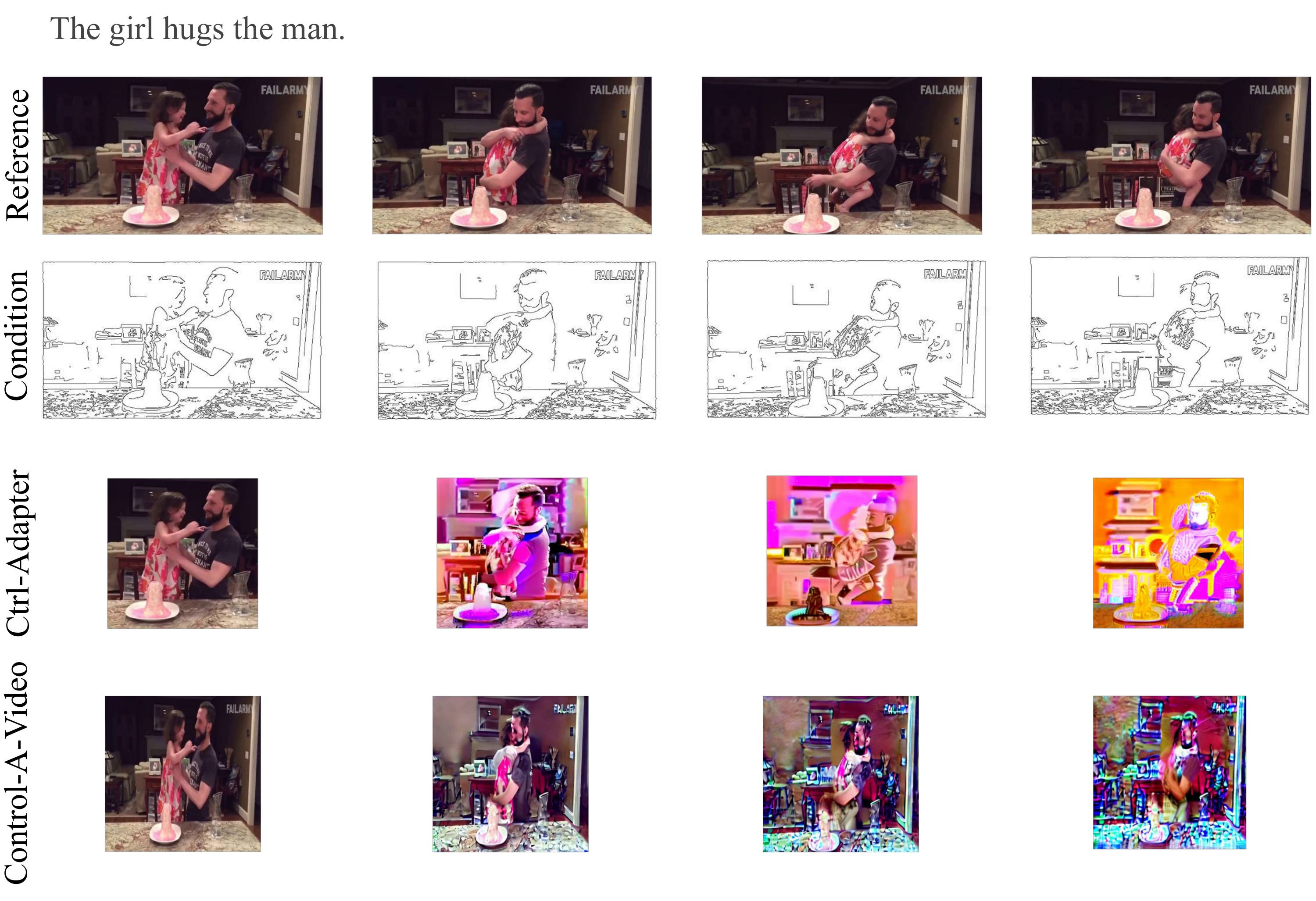}
  \vspace{-0.5\baselineskip}
  \caption{
  Example generations of our evaluated edge-based models.
  Once again, the generations of \ctrladapter and \cavideo are affected by high levels of saturation and distortion, despite being a static scene.
  }
  \label{fig:generation_samples_5_3}%
\end{figure*}

\clearpage
\section{Data annotation details}\label{app:data_prep}

\begin{figure*}[t!]
    \centering
    \includegraphics[width=\textwidth, trim={0 21.5cm 0 0}, clip]{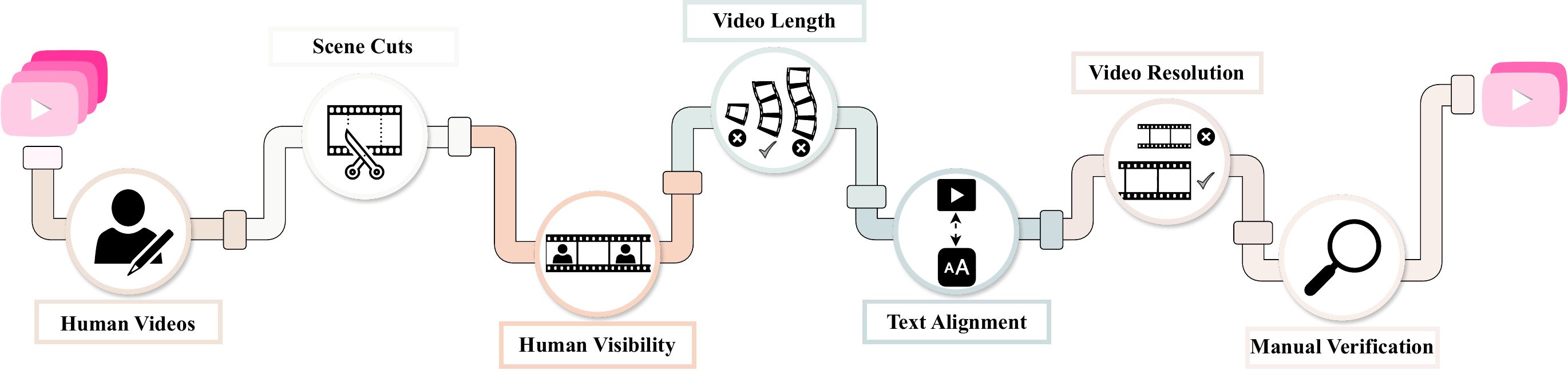}
    \caption{\textbf{\wydbold{} data filtering pipeline.} Our pipeline includes 7 steps: identifying videos with human actors, removing scene cuts, ensuring human visibility, removing short/long videos, keeping videos with high text alignment, removing low-res videos and manual verification.}
    \label{fig:wyd_pipeline}
\end{figure*}

\begin{table}[t!]
    \setlength{\tabcolsep}{3.0pt}
    \small
    \centering
    \resizebox{\linewidth}{!}{
        \begin{tabular}{lcccccc}
\toprule
                 & \multicolumn{3}{c}{\emph{Dev}}      & \multicolumn{3}{c}{\emph{Test}} \\
\textbf{Processing step}                                 & \textbf{UVO} & \textbf{Oops} & \textbf{DiDeMo} & \textbf{UVO} & \textbf{Oops} & \textbf{DiDeMo} \\
\midrule
StoryBench                                      & 2332         & 3472 & 2607   & 3895         & 3726 & 2319  \\
Videos of humans                                & 2085         & 3009 & 1479   & 3497         & 3168 & 1356  \\
Scene cuts                                      & 1960         & 2962 & 1497   & 3264         & 3126 & 1342  \\
Human visibility                                & ~~996          & 1800 & ~~847    & 1638         & 1938 & ~~783  \\
Video length                                    & ~~941          & 1545 & ~~706    & 1550         & 1650 & ~~670  \\
Text alignment                                  & ~~545          & ~~859  & ~~550    & ~~870          & ~~953  & ~~511  \\
Video resolution                                & ~~545          & ~~859  & ~~501    & ~~870          & ~~953  & ~~462  \\

Manual verification (\wyd)                      & ~~235          & ~~323  & ~~~~97     & ~~346          & ~~424  & ~~119 \\
\bottomrule
\end{tabular}

    }
    \caption{Number of StoryBench entries in our datasets after each step in the \wyd{} data preparation pipeline (see \cref{fig:wyd_pipeline} and \cref{sec:method_data_pipeline}).
    }
    \label{tab:wyd_filtering_size}
    \vspace{-0\baselineskip}
\end{table}

\begin{table*}[t!]
    \setlength{\tabcolsep}{3.0pt}
    \small
    \centering
    \resizebox{\linewidth}{!}{
        \begin{tabular}{llllllll}
\toprule
 Baby girl one & Bikers & Daddy & Group of People Two & Lady & Mother & Persons & Tourists \\
A baby & Boy & Dancers & Group of children & Lady one & Musicians & Pianist & Twins \\
A boy & Boy  & Divers & Group of men & Lady three & Officer & Player & User \\
A girl & Boy Four & Drummer & Group of musicians & Lady two & Old lady one & Players & Woman \\
A man & Boy One & Everyone & Group of people & Man & Old lady two & Police officer & Woman  \\
A old man & Boy Three & Family & Group of people  & Man  & Old man & Priest & Woman   \\
A person & Boy Two & Fighter & Group of people one & Man Five & Old woman & Rangers & Woman One \\
A person  & Boy five & Fighters & Group of people one  & Man Four & Others & Referee two & Woman Three \\
A woman & Boy four & Fourth man  & Group of people two & Man One & Passengers & Rest & Woman Two \\
Baby & Boy one & Gentleman & Group two & Man One  & People & Rider & Woman five \\
Baby  & Boy one  & Girl & Guitarist & Man Three & Peoples & Rider one & Woman four \\
Baby Boy & Boy three & Girl  & Guy & Man Two & Performer & Rider one  & Woman one \\
Baby boy & Boy two & Girl  three & He & Man five & Person & Rider two & Woman one  \\
Baby girl & Boy two  & Girl One & Judge & Man four & Person  & Riders & Woman three \\
Baby girl one  & Bride & Girl Two & Kid & Man one & Person Four & Runner & Woman two \\
Baby girl two & Bridegroom & Girl five & Kid  & Man one  & Person One & Second man  & Woman two  \\
Baby girl two  & Child & Girl four & Kid Four & Man one   & Person Three & Singer & Women \\
Baby one & Child  & Girl one & Kid Three & Man six & Person Two & Singers & Women  \\
Baby two & Child one & Girl one  & Kid Two & Man three & Person five & Someone & Women one \\
Band & Child one  & Girl third & Kid four & Man three  & Person four & Speaker & Young man \\
Batter & Child two & Girl three & Kid one & Man two & Person one & Swimmers & kid \\
Bichon frise & Child two  & Girl two & Kid one  & Man two  & Person one  & They & kid one \\
Bicyclist & Couple & Girl two  & Kid three & Marchers & Person three & Third & kid three \\
Bicyclists & Cyclist & Girls & Kid two & Mascot & Person three  & Third boy & kid two \\
Bike rider & Cyclist Two & Group & Kid two  & Members & Person two & Third girl & person \\
Biker & Cyclists & Group of People One & Kids & Men & Person two  & Third man  & woman \\
\bottomrule
\end{tabular}
    }
    \caption{List of unique human actors extracted from StoryBench annotations.
    }
    \label{tab:human_actors}
\end{table*}

\begin{figure*}[t!]
    \centering
    \includegraphics[width=\linewidth, trim={0 2cm 0 0}, clip]{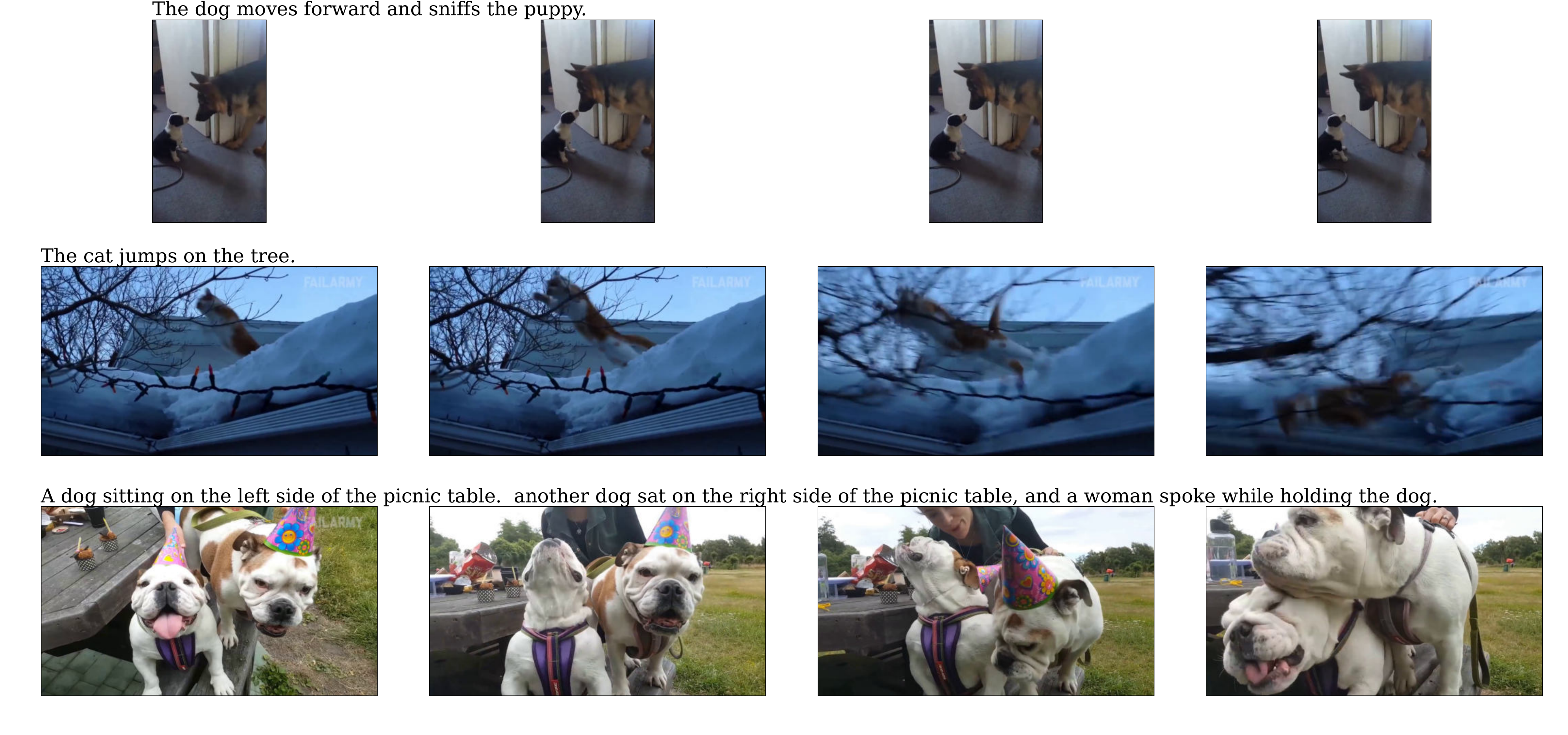}
    \caption{Discarded samples due to no human actor.}
    \label{fig:drop_pipeline_1}
\end{figure*}

\begin{figure*}[t!]
    \centering
    \includegraphics[width=\linewidth, trim={0 2cm 0 0}, clip]{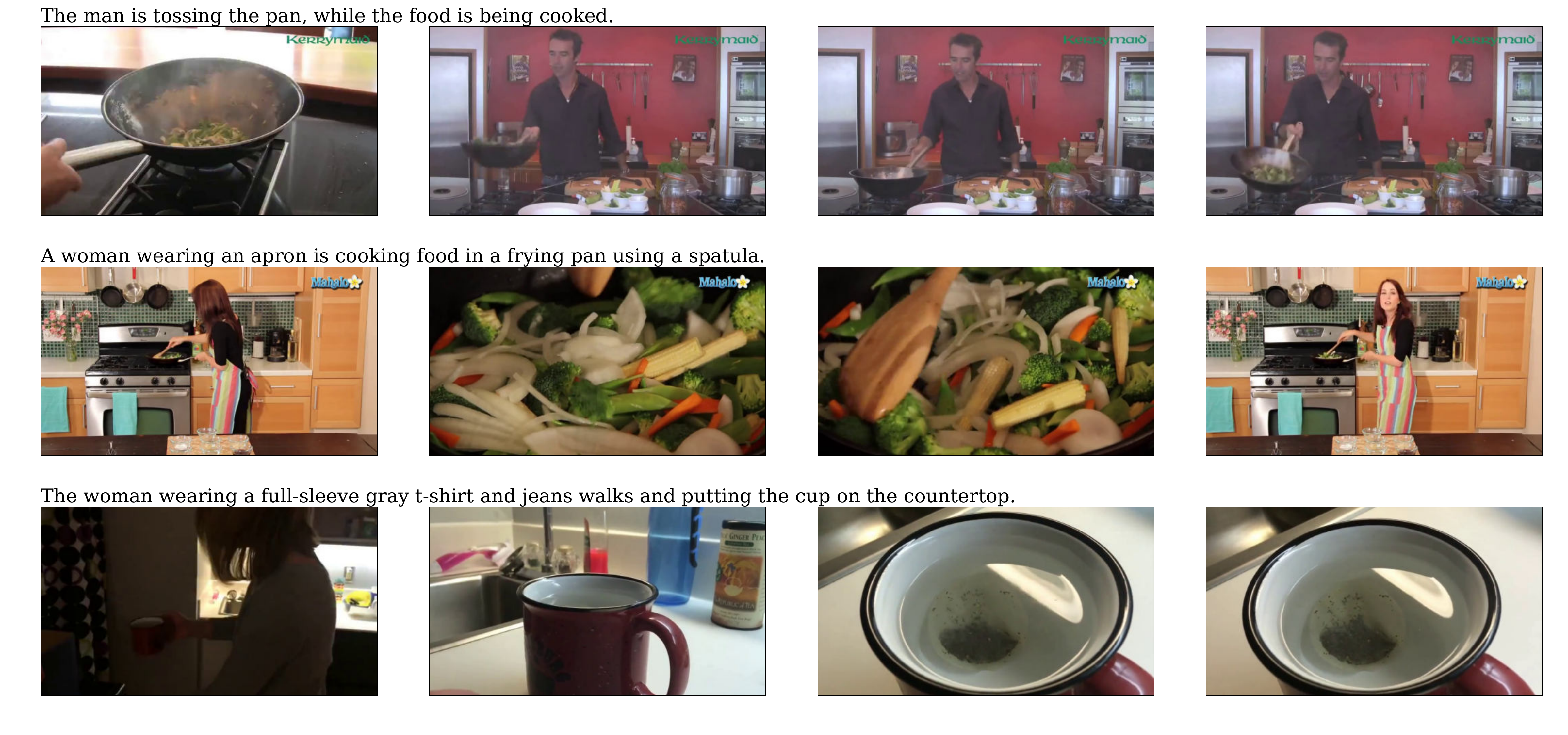}
    \caption{Discarded samples due to scene cuts.}
    \label{fig:drop_pipeline_2}
\end{figure*}

\begin{figure*}[t!]
    \centering
    \includegraphics[width=\linewidth, trim={0 2cm 0 0}, clip]{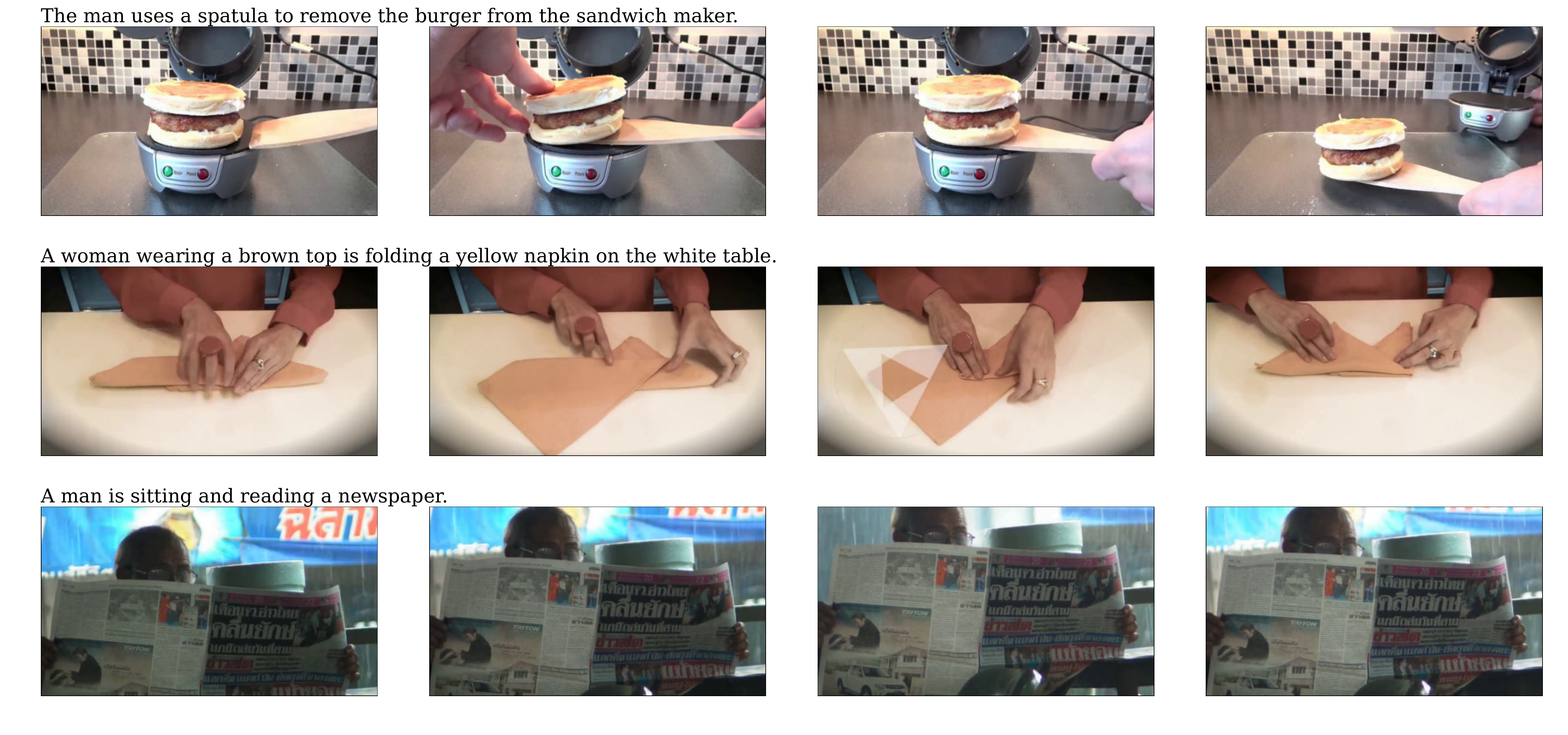}
    \caption{Discarded samples due to low human actor's visibility.}
    \label{fig:drop_pipeline_3}
\end{figure*}

\begin{figure*}[t!]
    \centering
    \includegraphics[width=\linewidth, trim={0 2cm 0 0}, clip]{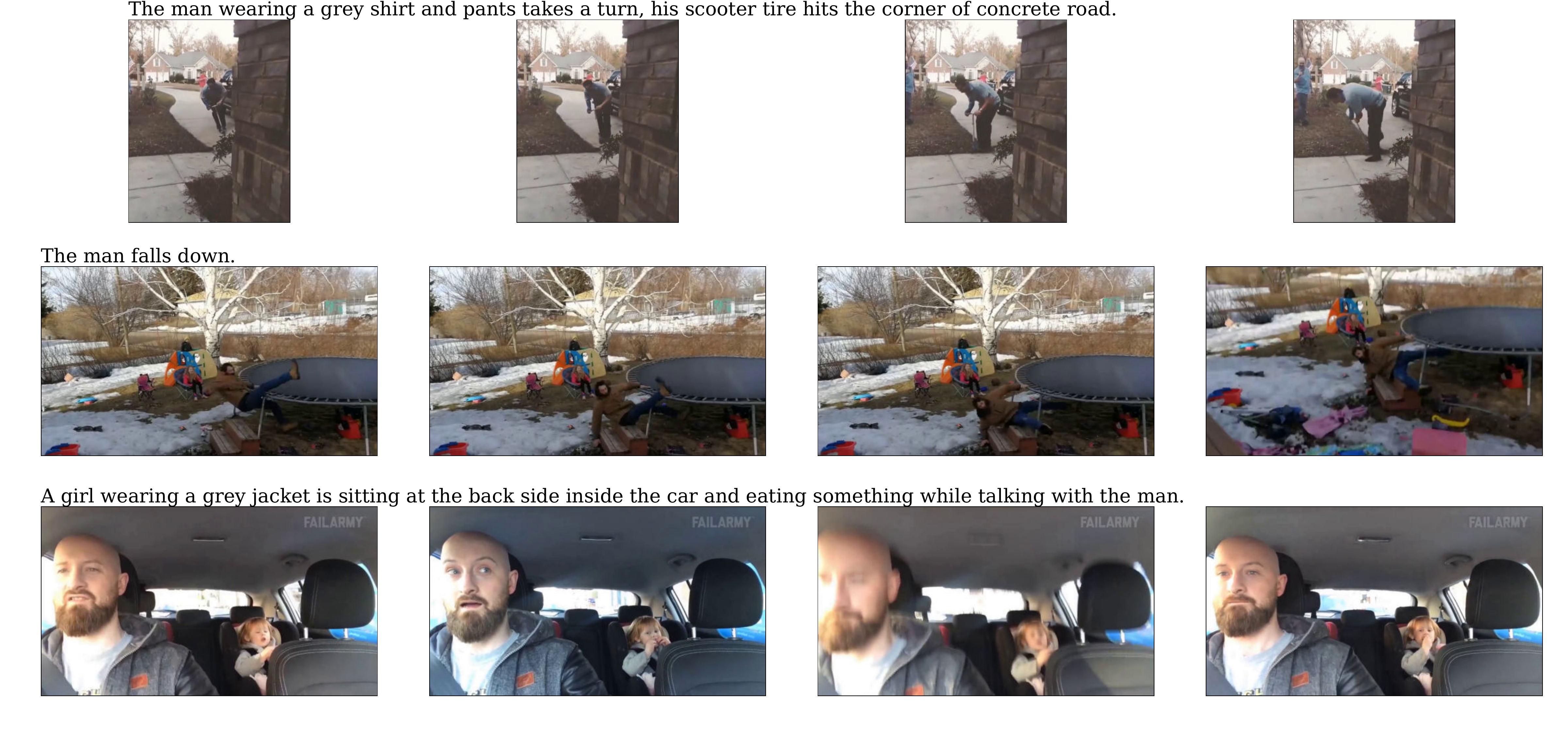}
    \caption{Discarded samples due to video duration.}
    \label{fig:drop_pipeline_4}
\end{figure*}

\begin{figure*}[t!]
    \centering
    \includegraphics[width=\linewidth, trim={0 2cm 0 0}, clip]{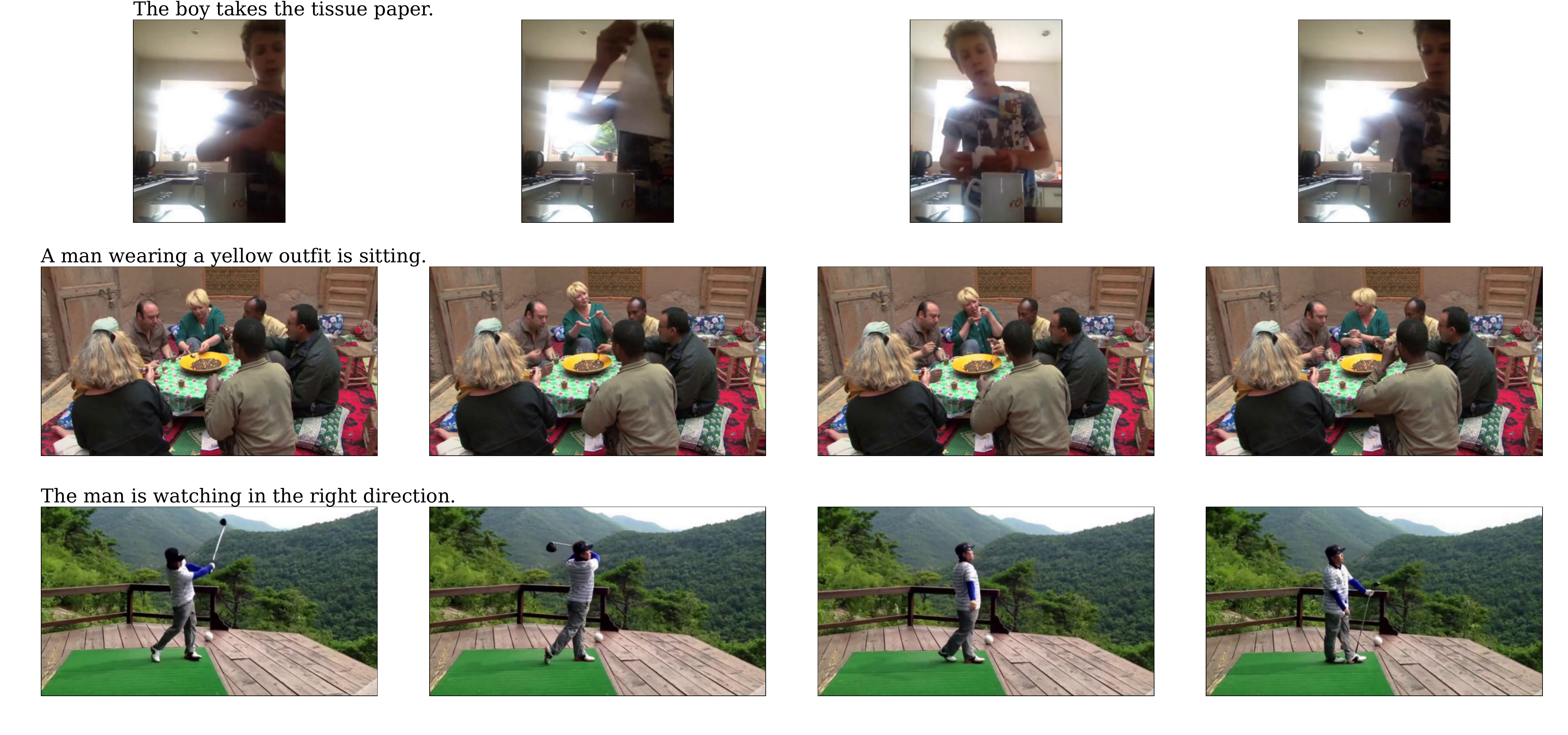}
    \caption{Discarded samples due to low video--text alignment.}
    \label{fig:drop_pipeline_5}
\end{figure*}

\begin{figure*}[t!]
    \centering
    \includegraphics[width=\linewidth, trim={0 2cm 0 0}, clip]{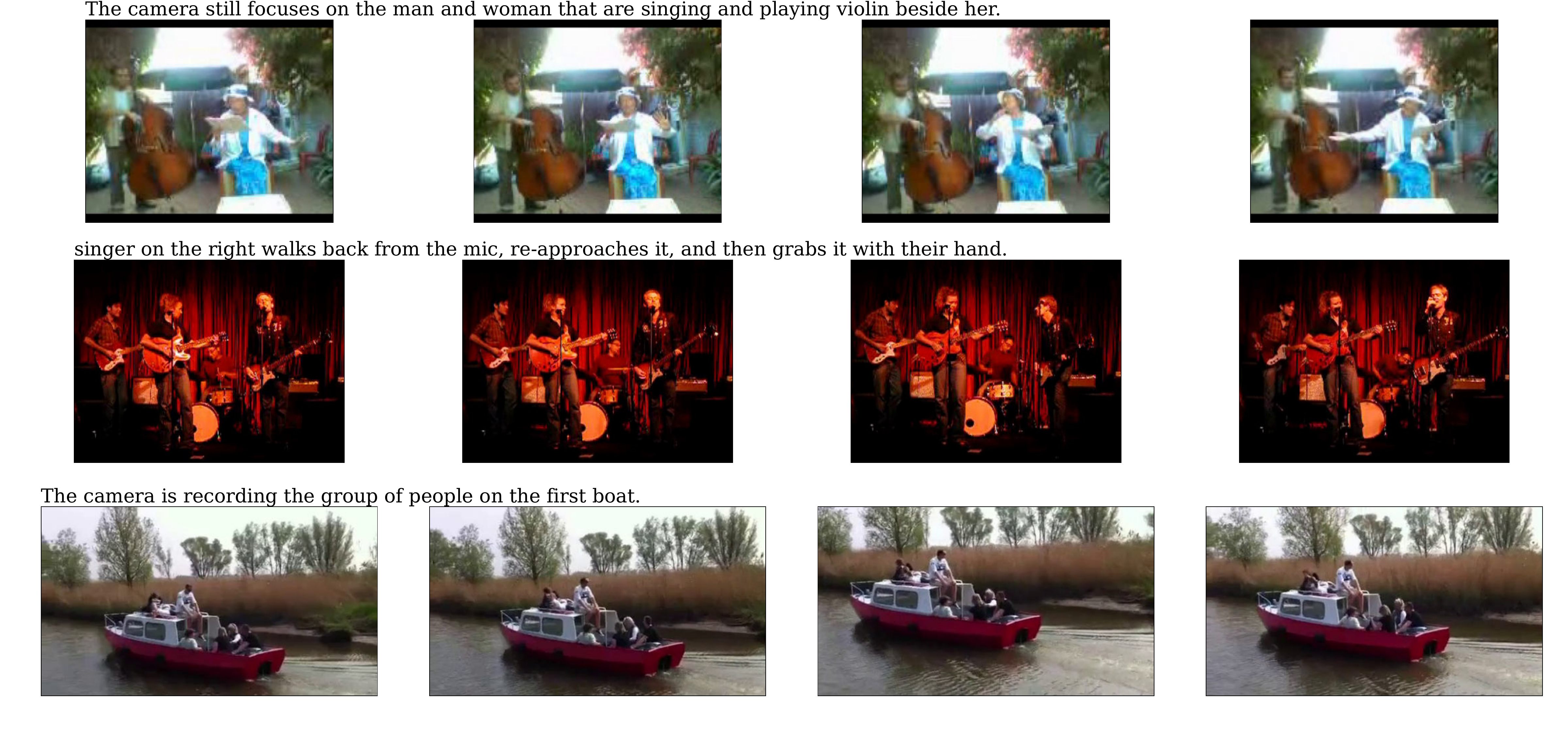}
    \caption{Discarded samples due to low video resolution.}
    \label{fig:drop_pipeline_6}
\end{figure*}

\begin{figure*}[t!]
    \centering
    \includegraphics[width=\linewidth, trim={0 2cm 0 0}, clip]{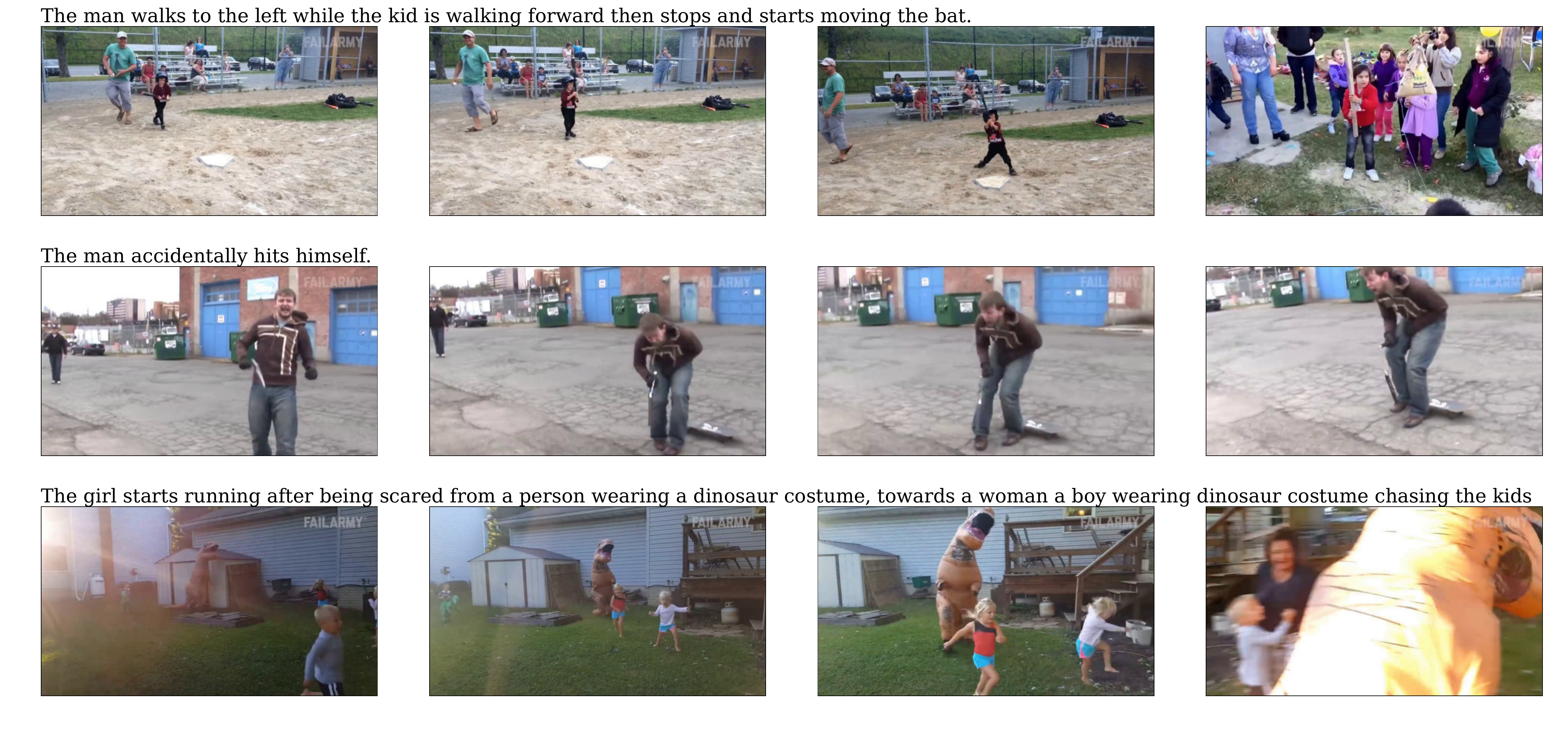}
    \caption{Discarded samples due to manual verification.}
    \label{fig:drop_pipeline_7}
\end{figure*}

\begin{figure*}[t]
  \centering
  \includegraphics[width=\linewidth, trim={0 0 0 0}, clip]{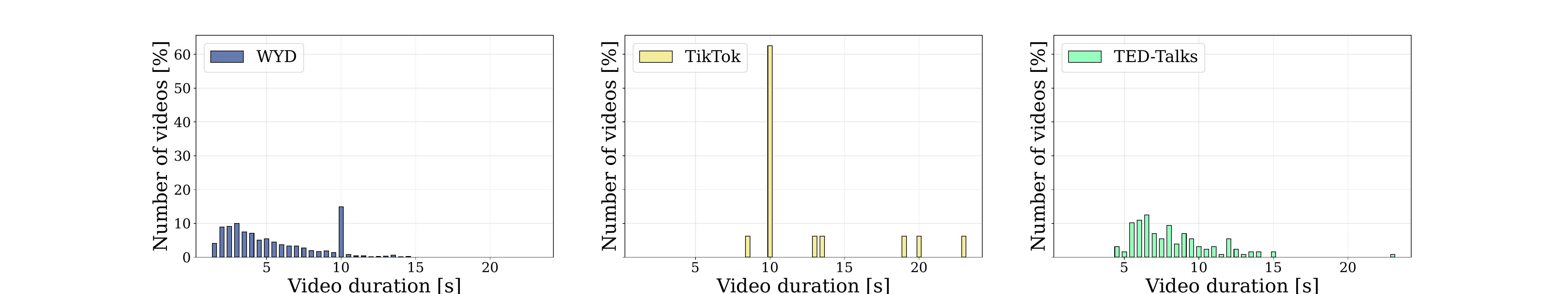}
  \caption{
  \textbf{Distribution of video duration in \wyd, \tiktok and \tedtalks.} \wyd{} covers actions lasting a few seconds and up to 15s.
  }
  \label{fig:duration_distr}%
  \vspace{0.5\baselineskip}
  \includegraphics[width=\linewidth, trim={0 0 0 0}, clip]{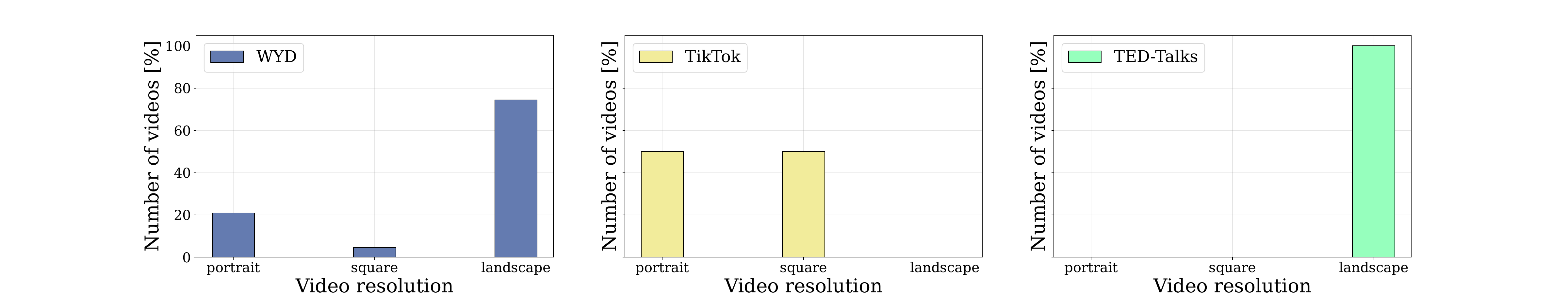}
  \caption{
  \textbf{Distribution of video resolution in \wyd, \tiktok and \tedtalks.} \wyd{} contains videos that have more diverse aspect ratios.
  }
  \label{fig:resolution_distr}%
\end{figure*}

In this section, we provide additional details, statistics and samples of our data preparation and categorization processes.

\paragraph{Data licenses.}
We rely on three publicly available datasets to source videos for our benchmark.
Kinetics~\cite{kinetics400} is released under a CC BY 4.0 license, DiDeMo~\cite{didemo} is released under a CC BY-NC-SA 2.0 license, and Oops~\cite{oops} is released under CC BY-NC-SA 4.0 license.
Moreover, we use captions and metadata collected in StoryBench~\cite{storybench}, which are also released under a CC BY 4.0 license.

\subsection{Data filtering}
High-quality text descriptions are necessary to accurately evaluate text-guided video models.
For the datasets above, StoryBench~\citep{storybench} includes human-written captions for video generation, as well as useful metadata, including event boundaries and actor identification (\ie, the entities with a key role) in a video~\citep{vidln}.
\wyd{} leverages StoryBench annotations, but we treat each video segment separately (rather than as part of a sequence) in our 7-step pipeline (see \cref{fig:wyd_pipeline}). \cref{tab:wyd_filtering_size} details how each step of our pipeline affected the number of entries of StoryBench, and \cref{fig:drop_pipeline_1,fig:drop_pipeline_2,fig:drop_pipeline_3,fig:drop_pipeline_4,fig:drop_pipeline_5,fig:drop_pipeline_6,fig:drop_pipeline_7} show examples that were dropped at each step.

\textbf{1. Videos with human actors.}
We start by filtering out videos where the main actors are not humans.
For Kinetics and Oops videos, we use human-labeled metadata in StoryBench, which associates each caption to its main actor (\eg, ``man with white t-shirt'').
For DiDeMo, we extract the main actors of each caption with an instruction-tuned LLM~\citep{gemini} (note that it is not trivial to use object detectors to extract the main actors due to non-salient humans in a video) with the prompt: ``Which living being is performing the main action in the following caption? Reply with one word.''
From 600+ actors, we manually identify 224 referring to humans (\cref{tab:human_actors}).

\textbf{2. Removing scene cuts.}
Most of the original videos are single shot, but we found a few of them with multiple scenes.
We use a shot detector~\citep{pyscenedetect} and whenever it detects 2--4 cuts, we remove the video if none of the parts lasts for at least 80\% of the original duration. Otherwise we replace the original video with that part.
Doing so allows us to remove scene cuts while preserving the actions described in the captions.

\textbf{3. Ensuring visible humans.}
People performing the main action in the video need to be visible, especially in the first frame, to evaluate how image-to-video systems can generate them.
We annotate our videos with a state-of-the-art human pose estimator~\citep{dwpose}, and keep only those in which people are `mostly visible' (defined as 11/18 body keypoints detected) in the first, and at least 70\% of the frames.

\textbf{4. Removing short and long videos.}
Given that our primary goal is to evaluate human video generation, we take into account the capabilities and the computational resources of most existing models to date, and opt to limit videos in our benchmark to a duration between 1.5s and 15s.

\textbf{5. High text alignment.}
The captions for Kinetics and Oops videos were originally made from the perspective of a specific actor~\citep{vidln}.
As a result, some captions fail to naturally describe the most salient actors in a video.
To minimize such cases, we only keep the actor whose video--caption pairs have the highest similarity according to a fine-grained contrastive VLM~\citep{pyramidclip}.
With the same approach, we then remove the bottom 25\% of the videos to maximize text controllability.

\textbf{6. Minimum resolution.}
We further filter down our data to only include videos with smaller edge of at least 360 pixels, to ensure references of higher quality while keeping enough samples for statistical significance.

\textbf{7. Manual verification.}
Finally, we meticulously scrutinize the quality of the resulting videos and remove those with (i) significant blur, (ii) poor lighting, (iii) unstable camera, (iv) low motion, (v) unclear captions, or (vi) where the first frame does not capture the main actors.
This process, in conjunction with the video categorization below, was done in multiple rounds and took over 500 hours of annotation time.
The authors validated the annotations and sought to ensure that the videos had high diversity and would be challenging for current video generation models.

At the end of this pipeline, \wyd{} contains 1{,}544 high-quality videos (from the original 18{,}351) which enable the fine-grained analyses described next at a tractable runtime.
\cref{fig:duration_distr,fig:resolution_distr} show the distribution of video duration and resolution for \wyd, \tiktok and \tedtalks.

Manual video quality verification and labeling were carried through an extensive period (over two months).
Most of the resulting videos (99.5\%) have a size of at least 512p.
Moreover, we note that a fraction of the frames display motion blur, which is unavoidable in videos with high motion.

\subsection{Video categorization}

A key goal of our benchmark is to enable fine-grained understanding of the capabilities of video generation models to synthesize humans across different facets; rather than reporting a single aggregated score, as done with other datasets.
To achieve this, we annotate our data with nine categories that capture important aspects for synthesizing videos of humans.
Each category in turn contains sub-categories (see \cref{fig:stats_distr} for an overview), each with at least $\approx$100 samples so as to provide sufficient statistical power for our analyses.

\textbf{Number of human actors.}
For each video in \wyd{}, we manually label the exact number of humans performing the main actions (\ie, salient for generation), and then group them in three groups (1, 2, 3--8).
Notably, each sub-category presents specific challenges, from consistently generating multiple people to more dynamic videos with a single person.

\textbf{Human actor size.}
The size of human actors can affect how well a video generation model performs.
We manually estimate the area covered by the human actors in each video, and categorize them into seven splits of actor size~(\cref{fig:stats_distr}).

\textbf{Human occlusion.}
Object consistency is crucial in generated videos, and humans need to keep their appearance despite partial or full occlusions.
We measure the average number of body keypoints detected by the pose estimator~\citep{dwpose}, and categorize our videos into five ranges of human actor occlusion (\ie, percentage of keypoints that are not visible).

\textbf{Human actions.}
\label{sec:method_action_classes}
The ability to perform a wide range of actions is a distinctive characteristic of humans, and different actions may require disparate generation capabilities (\eg, swimming vs. eating).
We manually assign action labels to each video in multiple rounds, adjusting the levels of specificity after each round.
This process yields sixteen sub-categories of visually similar actions (as shown in \cref{fig:stats_distr}).

\textbf{Human locomotion.}
We manually classify human body movements into three categories: full-body, partial-body, and hand-focused.
While full-body motion indicates actors changing location in the video, partial-body motion involves only moving part of the body (\eg, the arms).
We label videos where hands' motion is crucial separately, as existing models often struggle to generate hands~\citep{Zhang_2024_CVPR,lu2024handrefiner,lei2024comprehensive}.

\textbf{Camera motion.}
We manually label each video as dynamic if the camera follows the actor, or static otherwise.

\textbf{Video motion.}
The primary aspect that differentiates videos from images is the motion within them.
We use an optical flow model~\citep{teed2020raft} to estimate the amount of motion in each video, which we then use to study how this key aspect affects human video generation across seven motion ranges.

\textbf{Actor interactions.}
Humans often interact with their environment, either through inanimate objects, animals or other humans.
Object interactions are often hard to generate, as they require a deeper understanding of shapes, texture and the physical laws of the world.
While previous work only evaluates solo actions~\citep{tiktok,tedtalks,Chan_2019_ICCV}, our annotations show that most of the videos in our dataset involve interactions.

\textbf{Scene.}
Different actions are associated with different environments (\eg, swimming) and video generation models should be able to synthesize a variety of environments.
We manually annotate the videos in \wyd{} with nine different scenes where actions take place (both indoors and outdoors).

\begin{figure*}[t!]
    \centering
    \includegraphics[width=\textwidth, trim={0 0 0 0}, clip]{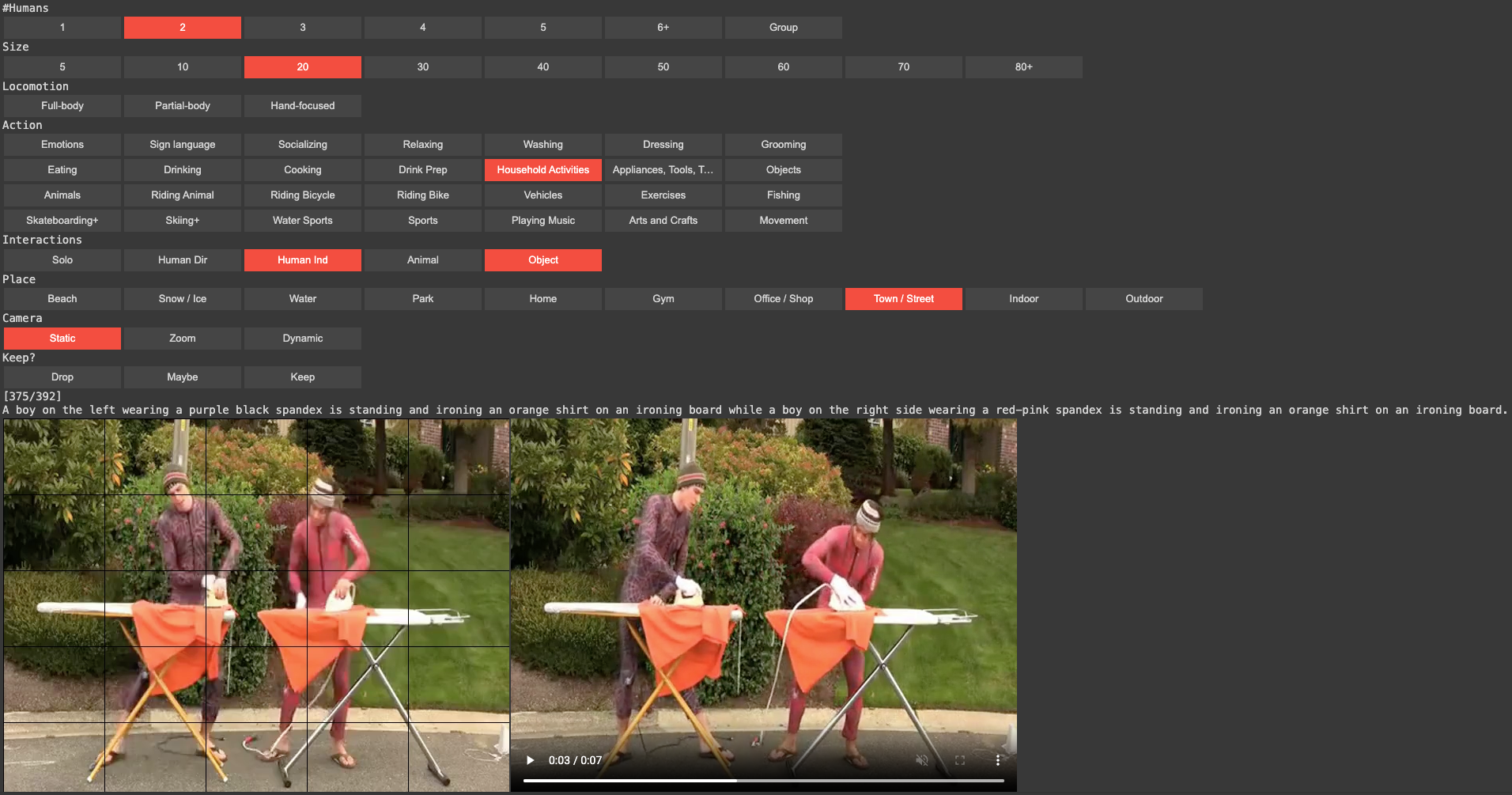}
    \caption{UI used to manually filter and label videos in \wyd{} according to different categories.}
    \label{fig:category_ui}
\end{figure*}

\begin{figure*}[t!]
    \centering
    \includegraphics[width=\textwidth, trim={0 0 0 0}, clip]{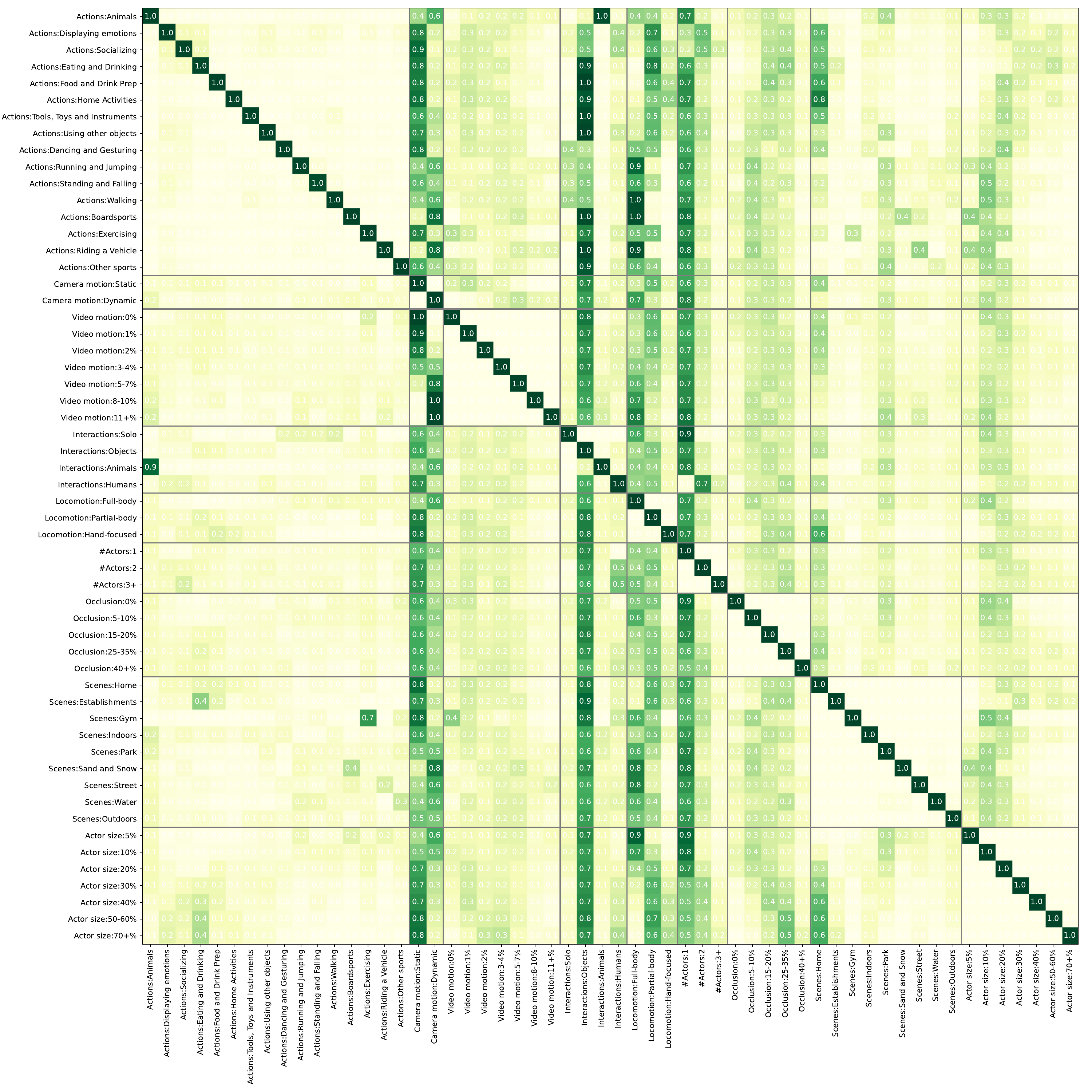}
    \caption{\textbf{Percentage of videos overlapping between two categories.} For a given row, computes how many of its videos (in percentage) are also available in another category (column). Best viewed on a screen due to its large number of entries.}
    \label{fig:category_overlaps}
\end{figure*}

\paragraph{Discussion.}
\cref{fig:category_ui} shows our UI for video categorization.
We find that only a few categories overlap with each other significantly (see \cref{fig:category_overlaps} in \cref{app:data_prep}).
Namely, actions and interactions with animals; and video and camera motion, where high-motion videos come from dynamic camera, and videos with low motion correspond to static camera.
Interestingly, high-motion videos do not necessarily involve fewer people, but small people are often associated with high motion and full-body movement of a single actor.

\subsection{Dense annotations}

\begin{figure*}[t!]
    \centering
    \includegraphics[width=\textwidth, trim={0 0 0 2mm}, clip]{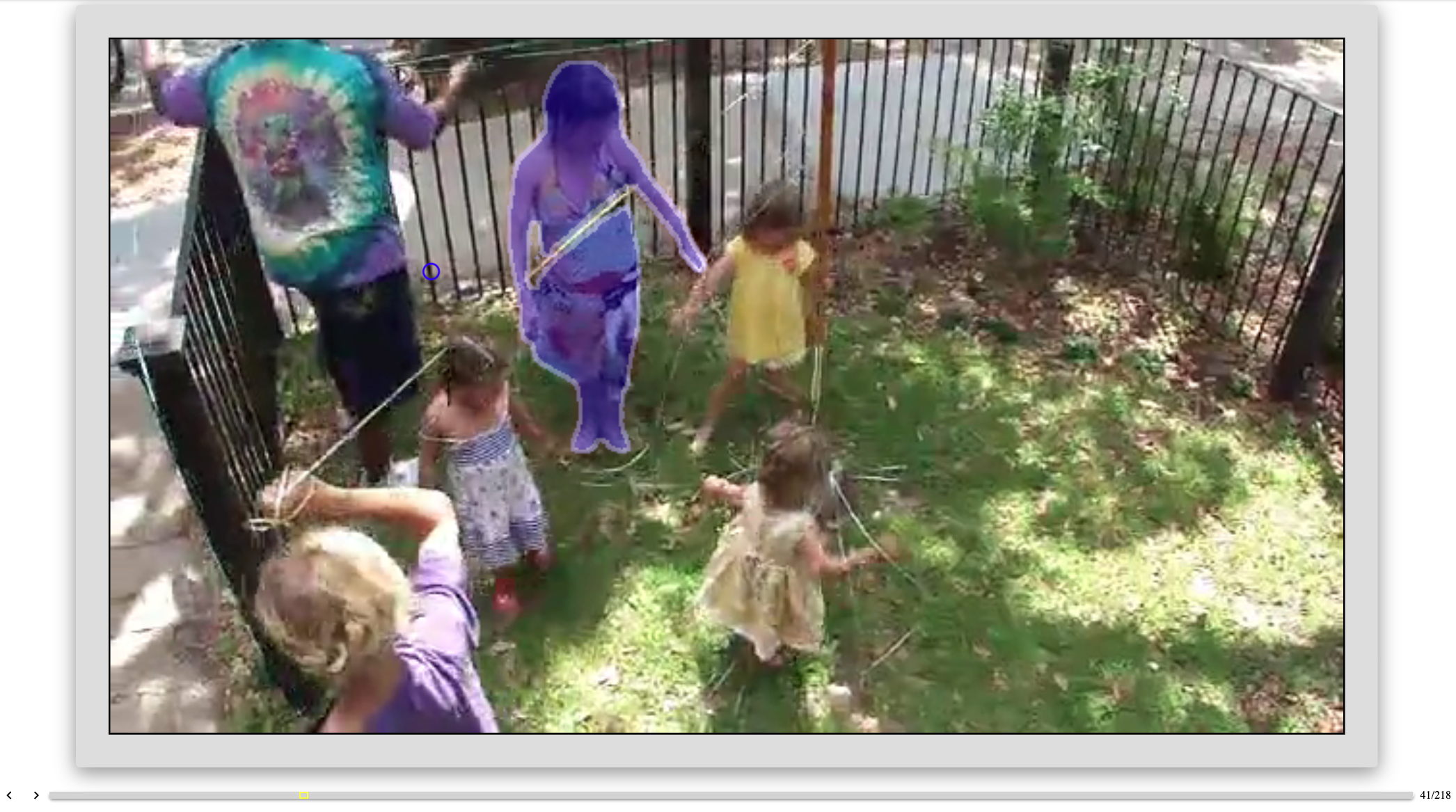}
    \caption{UI used to manually fix video segmentation masks in \wyd{} through a brush to select pixels corresponding to an actor's mask.}
    \label{fig:segmentation_ui}
\end{figure*}

\paragraph{Video segmentation masks.}
In addition to labeling our videos with 9 categories, we also annotate each human actor in a video with tracked segmentation masks.
To do this, we first identify people in the first frame of each video via bounding boxes using OWLv2~\citep{owlv2}.
After selecting and refining the bounding boxes corresponding to the actors only, we feed them as input to SAM 2~\citep{sam2}, which returns video segmentation tracks for each of the actors.
These tracks are further verified and manually corrected at the frame level by the authors, an effort that took over 1000 hours.
We use them to define new automatic metrics for \wyd{} by analyzing model performance at the human level, as discussed next.
\cref{fig:segmentation_ui} shows our UI for verifying and fixing video segmentation masks for each actor.
Annotations were made at every frame using a brush to extend or delete pixels for the automatically generated segmentation masks for each human actor.

\begin{figure*}[t!]
    \centering
    \includegraphics[width=\textwidth, trim={0 0 0 0}, clip]{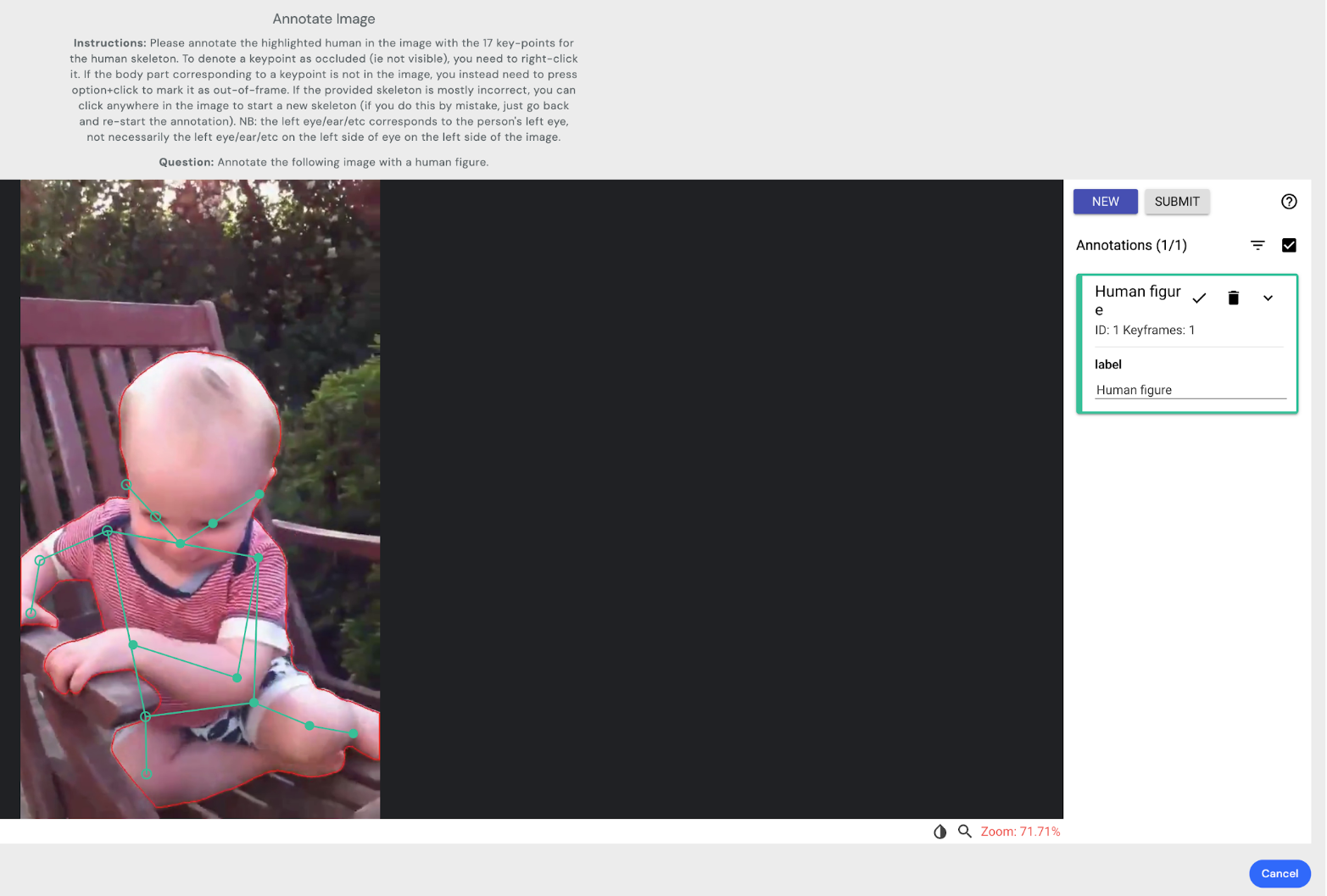}
    \caption{UI used to manually annotate 2D body key-points in \wyd{}.}
    \label{fig:pose_ui}
\end{figure*}

\paragraph{Actor 2D pose key-points.}
We further collect human-annotated 2D pose key-points for the (184) actors across 100 randomly selected videos (UI shown in \cref{fig:pose_ui}).
Human raters are given the option to modify existing keypoints extracted using DWPose~\citep{dwpose} or to discard them and annotate human skeletons from scratch.
This data collection campaign also took over 1000 hours of human rating time.
We use these skeletons to verify the trustworthiness of our evaluation framework.

\clearpage
\section{Additional results} \label{app:more_results}

In this section, we report complementary results from our experiments.\\

\begin{figure*}[t!]
  \centering
  \includegraphics[width=\linewidth]{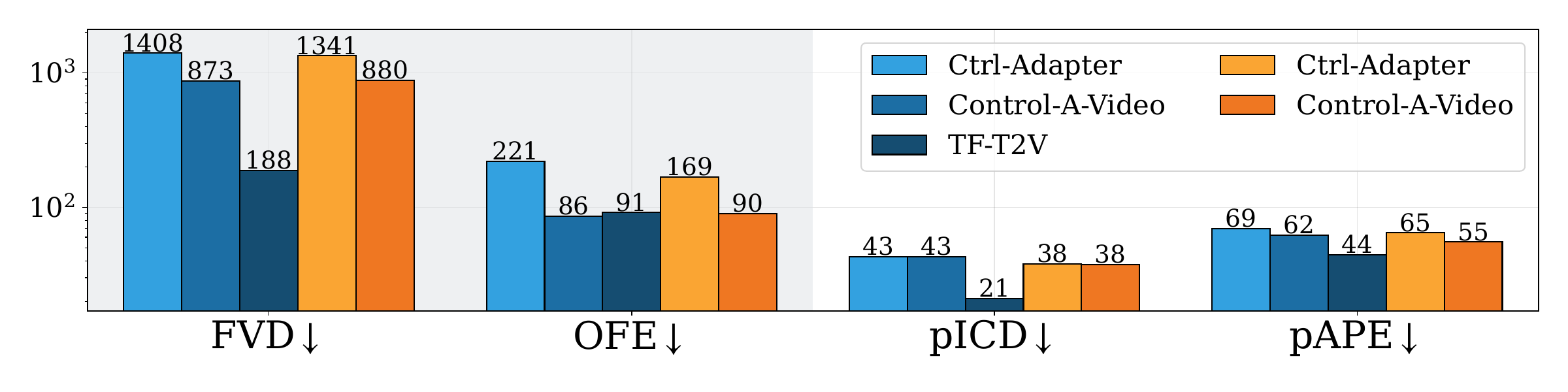}
  \vspace{-2.3\baselineskip}
  \caption{\small
  \textbf{Overall performance (left: video-level, right: human-level) of SOTA controllable depth- and edge-conditioned image-to-video models on \wyd$_{16}$.}
  Depth models are shown in blue, while edge models in shades of orange.
  \tftv obtains the overall best performance across our models.}
  \label{fig:dense_overall}
\end{figure*}

\begin{figure*}[t!]
  \centering
  \includegraphics[width=\linewidth]{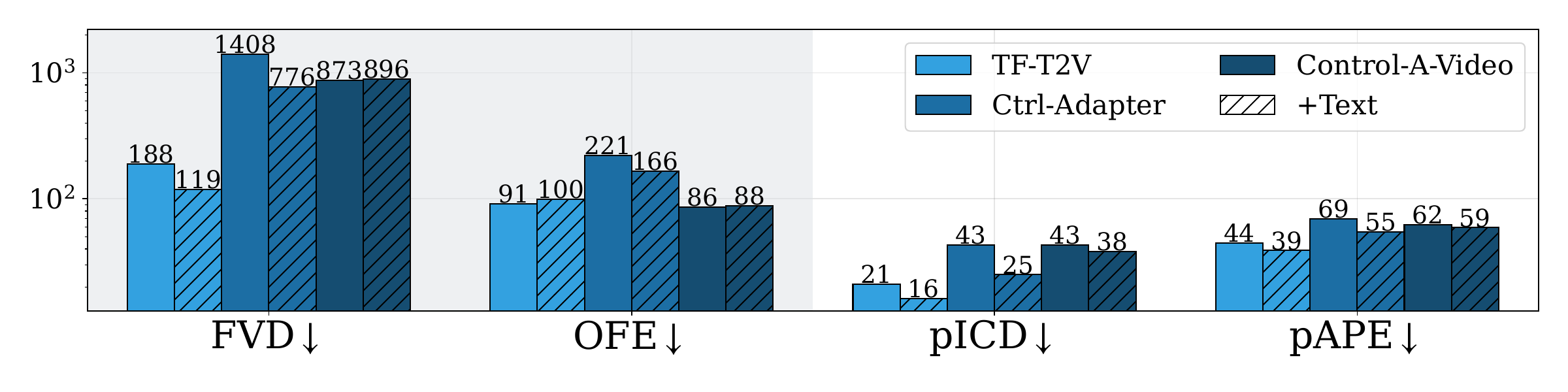}
  \vspace*{-.5\baselineskip}
  \includegraphics[width=\linewidth]{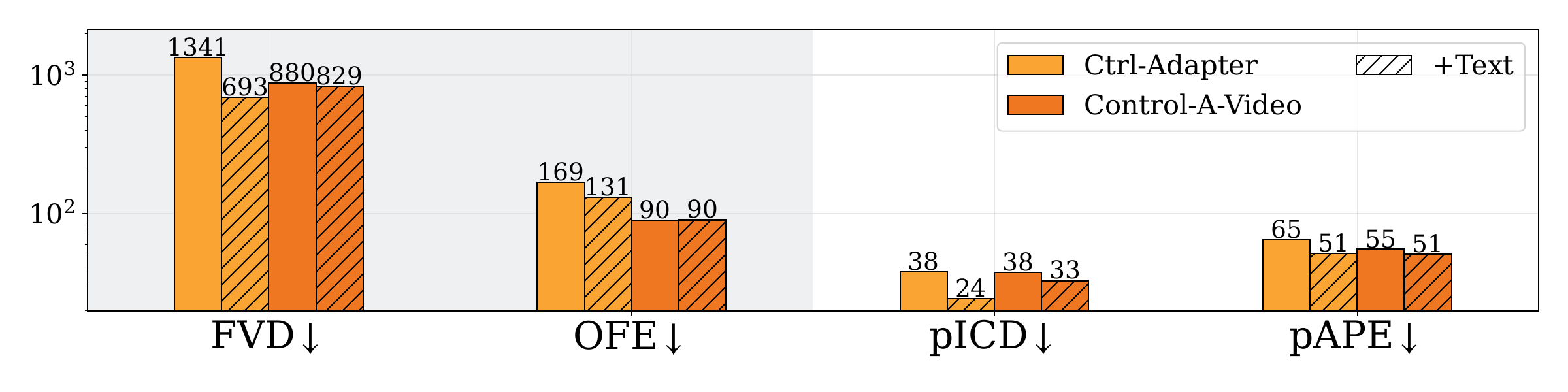}
  \vspace{-2\baselineskip}
  \caption{\small
  \textbf{Difference in errors in \wyd{} after adding captions to depth- and edge-conditioned models.}
  Adding text guidance usually improves performance of depth-guided (top) and edge-guided (bottom) models.}
  \label{fig:text_deltas}
\end{figure*}

\cref{fig:dense_overall} reports performance for our depth- and edge-conditioned models, while \cref{fig:text_deltas} shows the difference in errors when adding captions as an additional source of guidance to these models.\\

\begin{figure*}[t!]
  \centering
  \includegraphics[width=\linewidth]{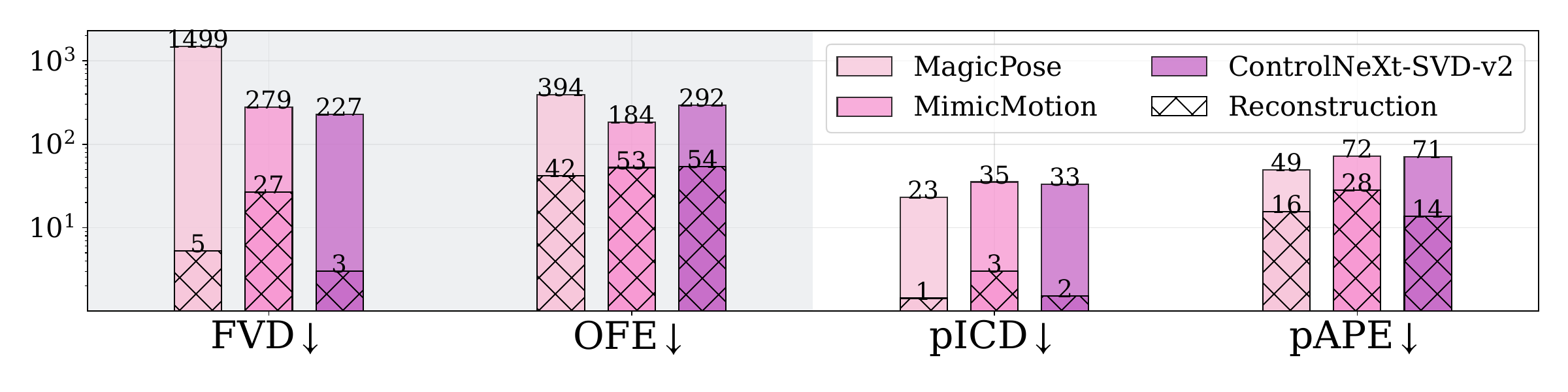}
  \vspace{-2.3\baselineskip}
  \caption{\small
  \textbf{Performance comparison between generation and auto-encoder reconstruction capabilities of pose-conditioned models on \wyd$_{16}$.}
  We see a clear gap between generation and reconstruction across all metrics, showing that models are capable of generating the reference videos but struggle during the generation process.
  }
  \label{fig:pose_reconstruct}
\end{figure*}

\begin{figure*}[t!]
  \centering
  \includegraphics[width=\linewidth]{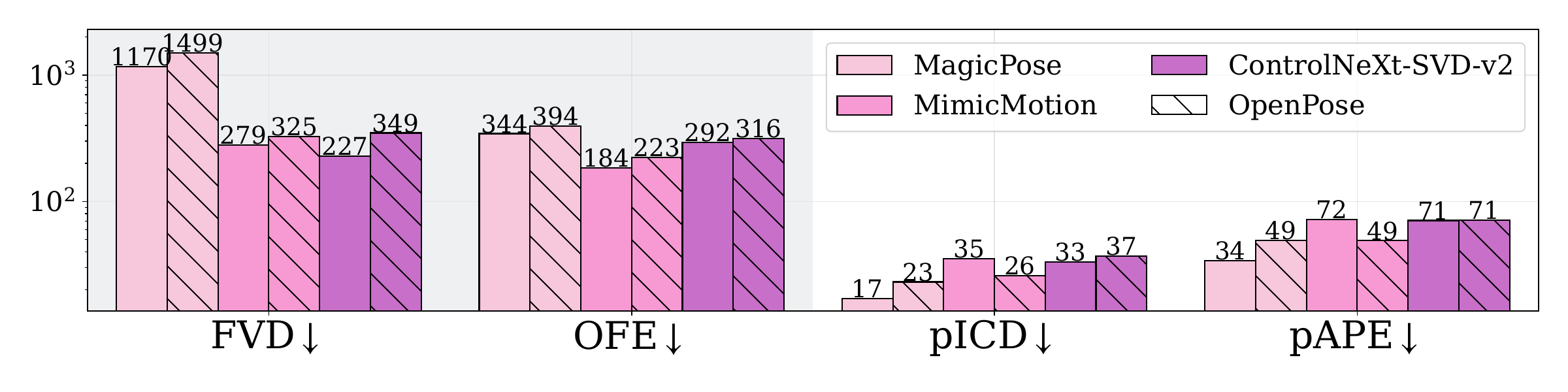}
  \vspace{-2.3\baselineskip}
  \caption{\small
  \textbf{Performance of pose-conditioned models when using the OpenPose detector rather than DWPose (default) on \wyd$_{16}$.}
  We see that using DWPose gives typically better performance, even when applied to \magicpose, which was trained using OpenPose.
  }
  \label{fig:pose_detector}
\end{figure*}

\begin{figure*}[t!]
  \centering
  \includegraphics[width=\linewidth]{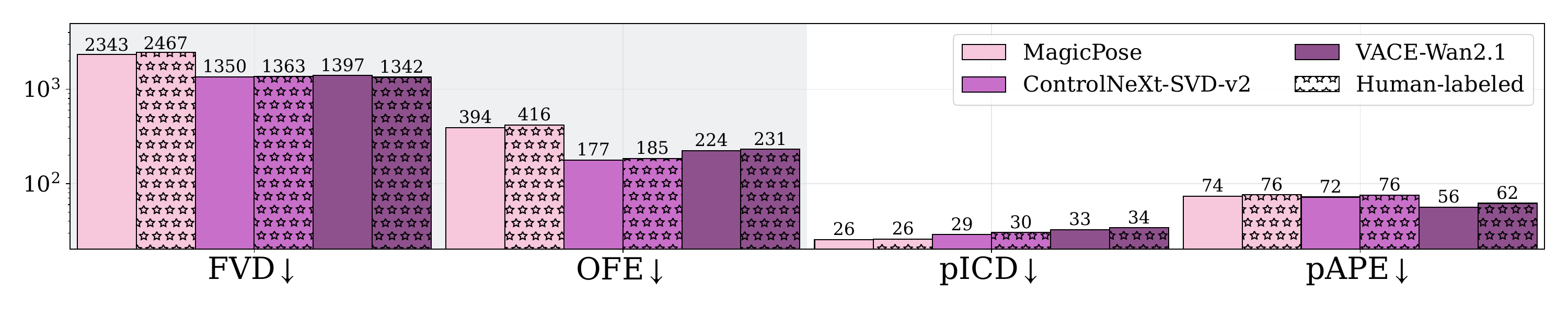}
  \vspace{-2.3\baselineskip}
  \caption{\small
  \textbf{Performance of pose-conditioned models when using the human-labeled 2D body key-points rather than DWPose (default) on a subset of 100 videos from \wyd$_{16}$.}
  We see that models achieve similar performance in these two scenarios, verifying the correctness of our results based on poses detected with DWPose.
  }
  \label{fig:pose_labels}
\end{figure*}

\cref{fig:pose_reconstruct,fig:pose_detector,fig:pose_labels} report our investigations of pose-conditioned models w.r.t. (i) auto-encoder reconstruction capabilities, (ii) role of different pose detectors, and (iii) the need for human-labeled pose key-points.\\

\begin{figure*}[t!]
  \centering
  \includegraphics[width=\linewidth]{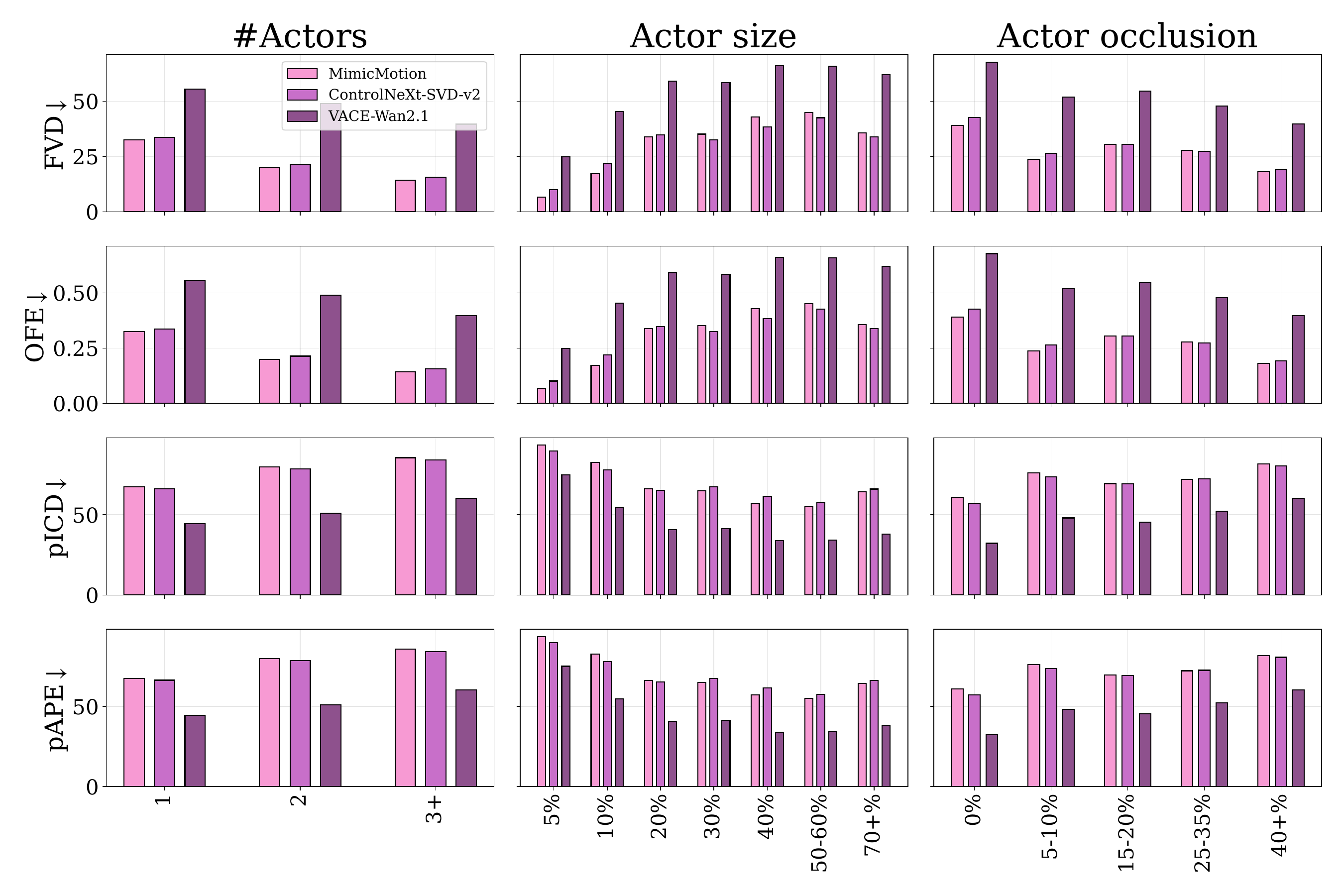}
  \caption{
  \textbf{Performance of best models w.r.t. `\#Actors,' `Actor size' and `Actor occlusion.'}
  Animating multiple actors is harder than a single one. Small humans are also harder to generate precisely compared to when they cover a large portion of the frame. Performance also tends to degrade as the amount of occlusion increases.}
  \label{fig:category_col1}
\end{figure*}

\begin{figure*}[t!]
  \centering
  \includegraphics[width=\linewidth]{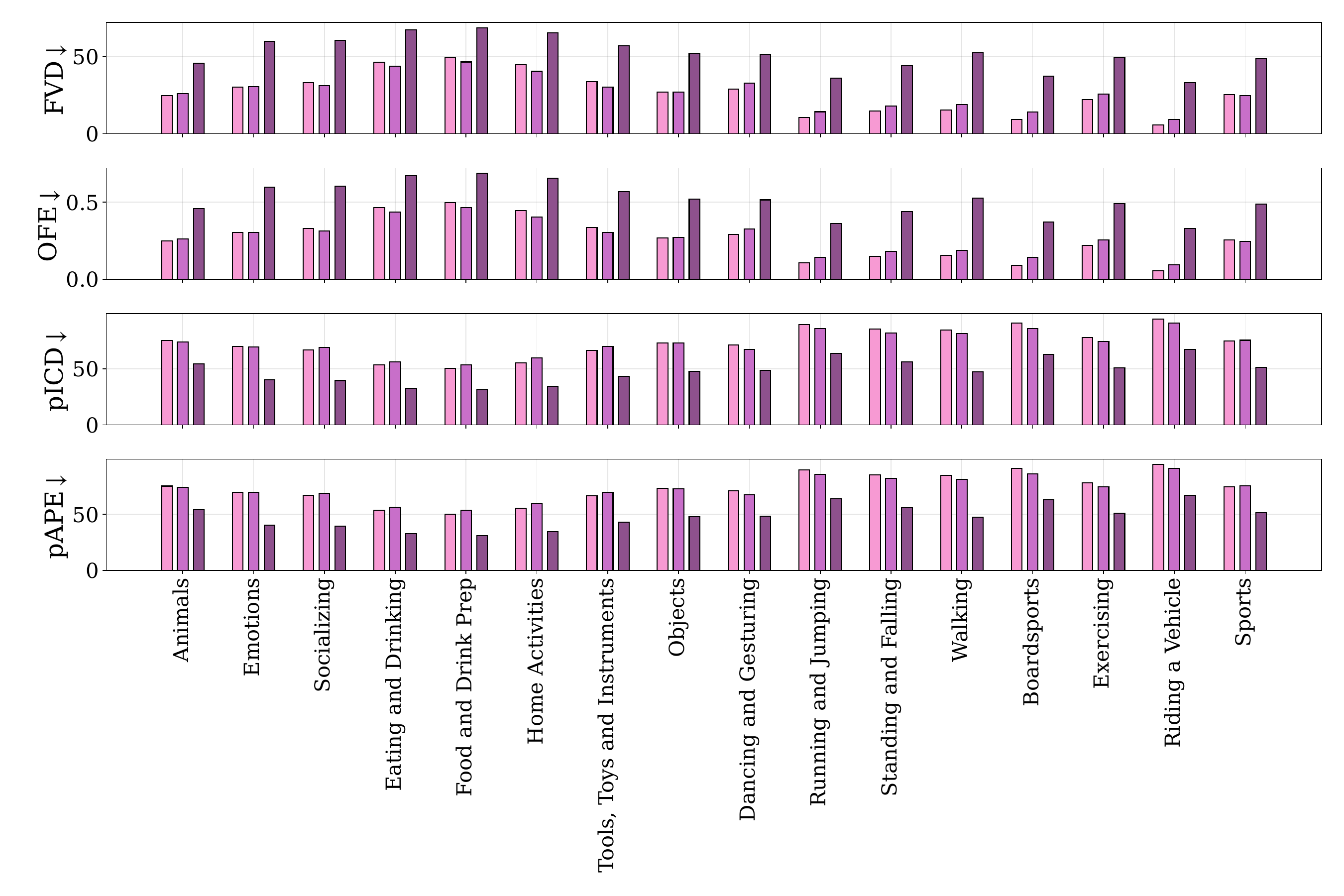}
  \caption{
  \textbf{Performance of best models w.r.t. `Actions.'}
  Actions involving animals, riding a vehicle, running and jumping, and boardsports are challenging for SOTA models.
  Atypical movements, \eg, standing up and falling down, are also hard.
  }
  \label{fig:category_col2}
\end{figure*}

\begin{figure*}[t!]
  \centering
  \includegraphics[width=\linewidth]{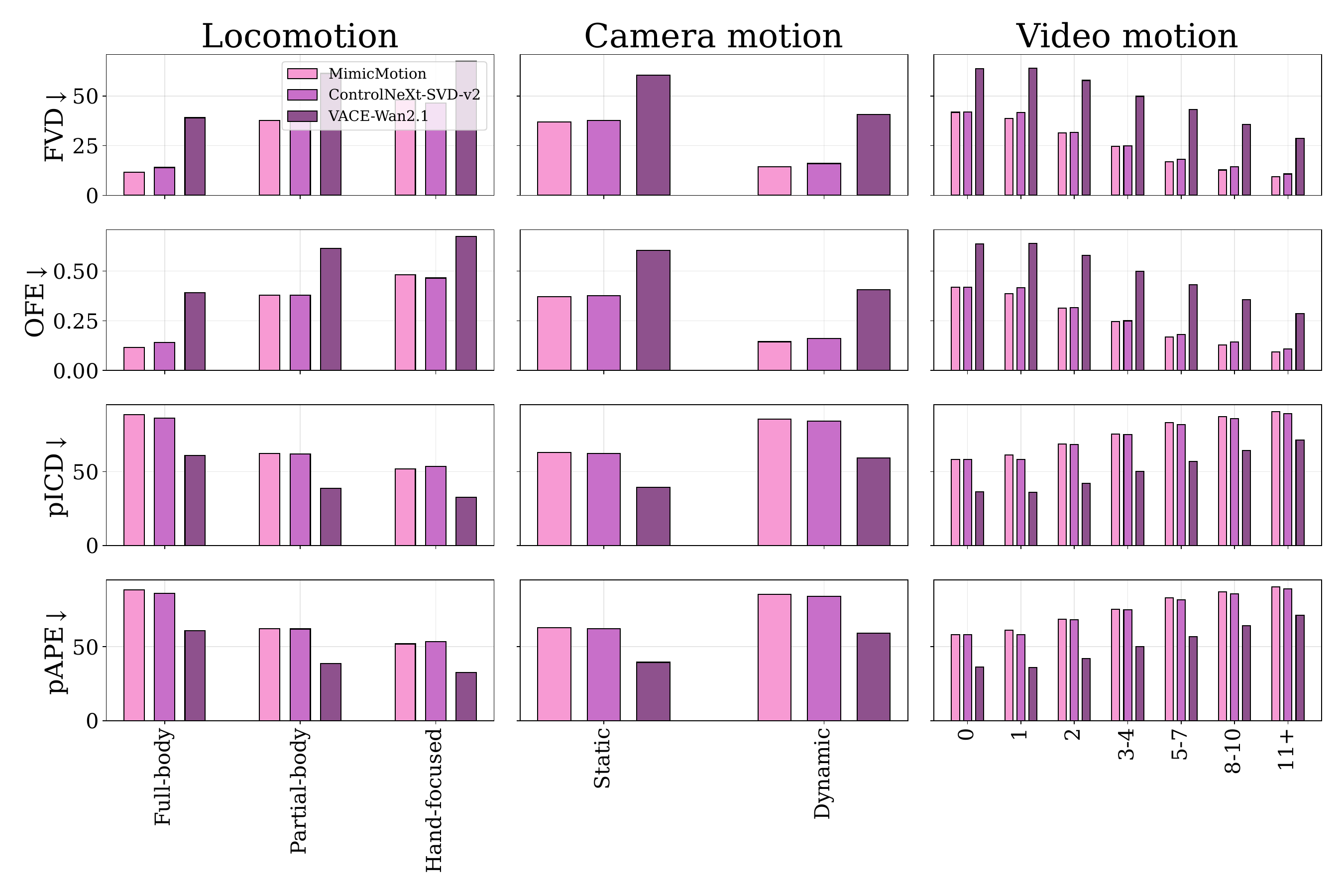}
  \caption{
  \textbf{Performance of best models w.r.t. `Locomotion,' `Camera motion' and `Video motion.'}
  Videos with full-body locomotion are more challenging to generate due to the larger changes required. Dynamic videos with high levels of motion are more challenging.
  }
  \label{fig:category_col3}
\end{figure*}

\begin{figure*}[t!]
  \centering
  \includegraphics[width=\linewidth]{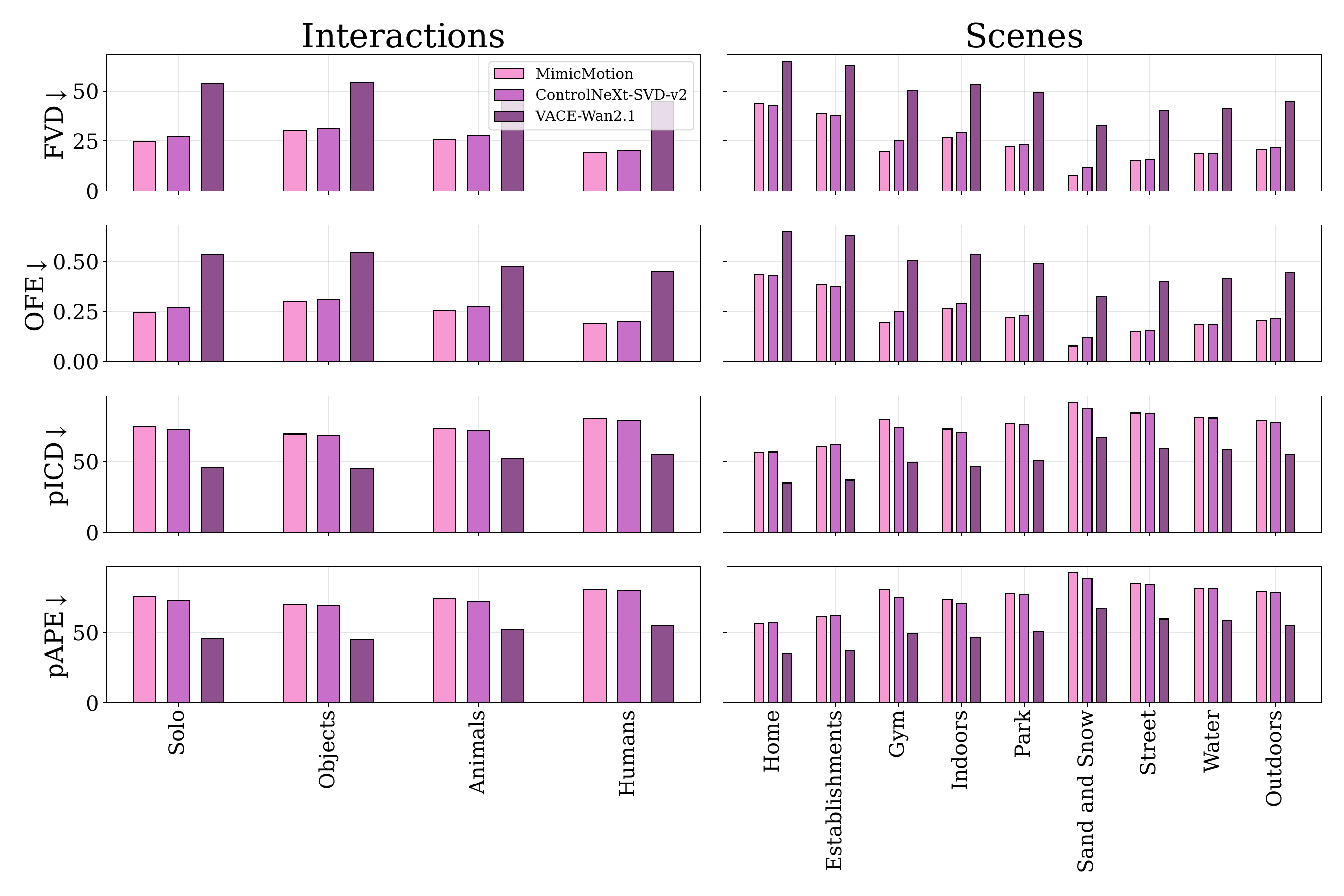}
  \caption{
  \textbf{Performance of best models w.r.t. `Interactions' and `Scenes.'}
  Generating videos of humans interacting with animals or other humans is more difficult than solo videos.
  Outdoors scenes (\eg, on sand and snow, street, by the water) are also harder for SOTA models.
  }
  \label{fig:category_col4}
\end{figure*}

\cref{fig:category_col1,fig:category_col2,fig:category_col3,fig:category_col4} report category-level performance of our best models (\mimicmotion, \controlnext and \tftv) according to sample-level metrics (ICD, OFE, pICD, pAPE).


\begin{figure*}[t]
  \centering
  \includegraphics[width=\linewidth, trim={0 0 0 0}, clip]{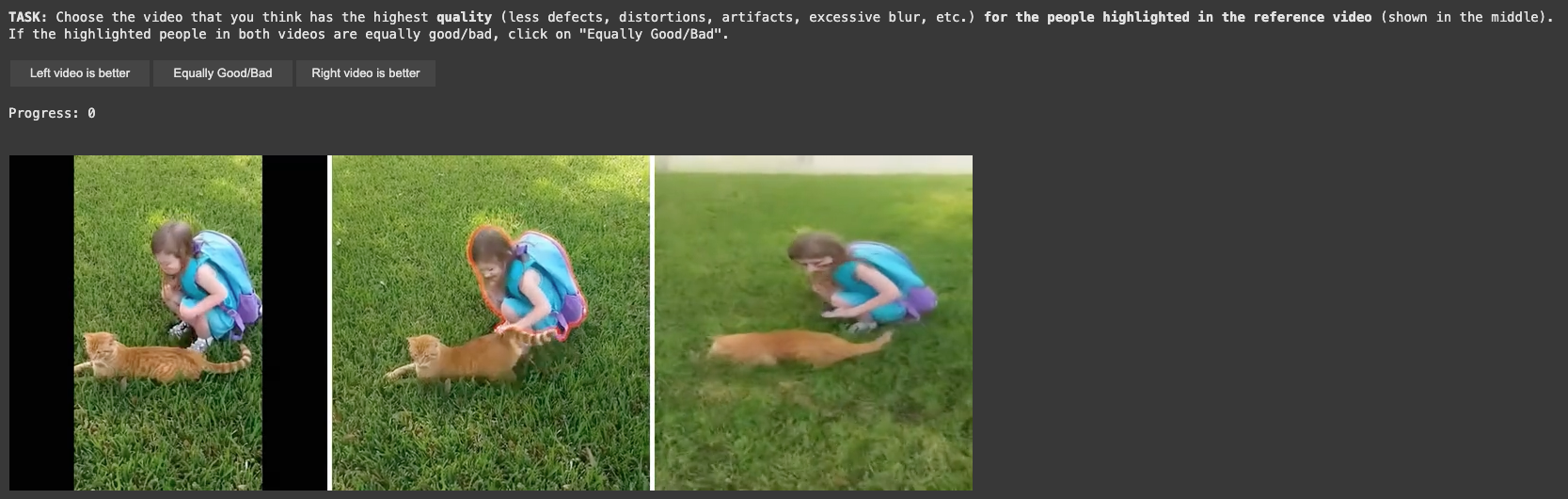}
  \caption{
  \textbf{UI used to collect human preferences in side-by-side evaluations for \wyd.}
  We remove the actors' segmentation masks when comparing video motion similarity.
  No reference video is shown for video quality comparisons.
  }
  \label{fig:sidexside_ui}%
\end{figure*}
\clearpage
\section{Evaluations}\label{app:human_evals}

In this section, we report the performance of our evaluated models according to several automatic metrics and detail our human evaluation protocol.

\begin{figure*}[t]
  \centering
  \includegraphics[width=\linewidth, trim={0 0 0 0}, clip]{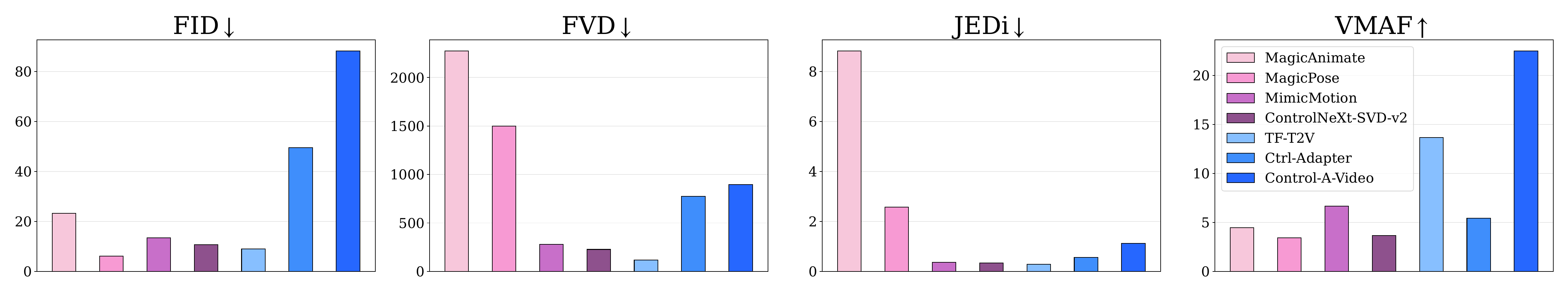}
  \caption{
  \textbf{Comparison of video quality metrics on \wyd$_{16}$.}
  FID favors \magicpose's flickering videos, and VMAF ranks \cavideo's videos with distortions and artifacts first.
  FVD and JEDi rank video generations with high agreement to human judgments.
  }
  \label{fig:video_quality}%
\end{figure*}

\begin{figure}[t!]
  \centering
  \includegraphics[width=\linewidth, trim={0cm 0cm 0cm 0}, clip]{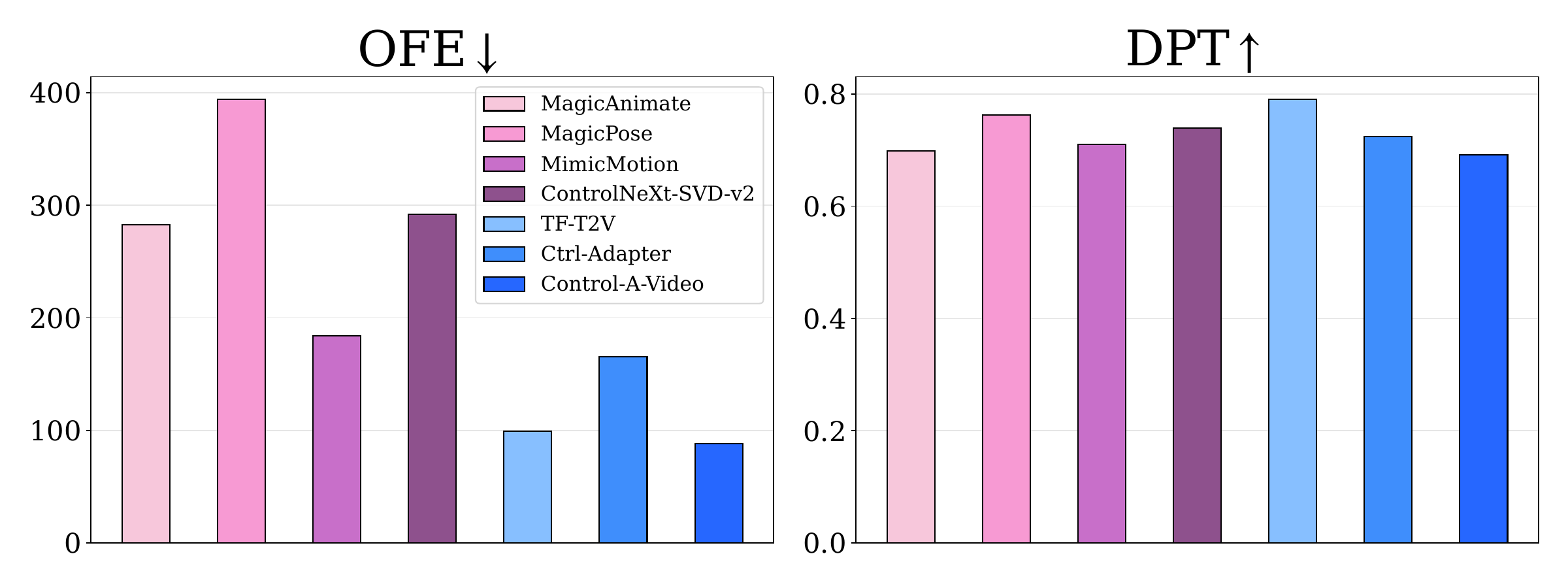}
  \caption{
  \textbf{Comparison of video motion similarity metrics on \wyd$_{16}$.}
  We see that a depth-based metric (DPT) ranks the flickering generations from \magicpose as the second best ones, and those of pose-guided \controlnext better than those of depth-conditioned \ctrladapter and \cavideo, which better generate videos with dynamic camera.
  OFE better agrees with the human rankings.
  }
  \label{fig:video_motion}%
  \vspace{\baselineskip}

  \centering
  \includegraphics[width=\linewidth, trim={0cm 0cm 0cm 0}, clip]{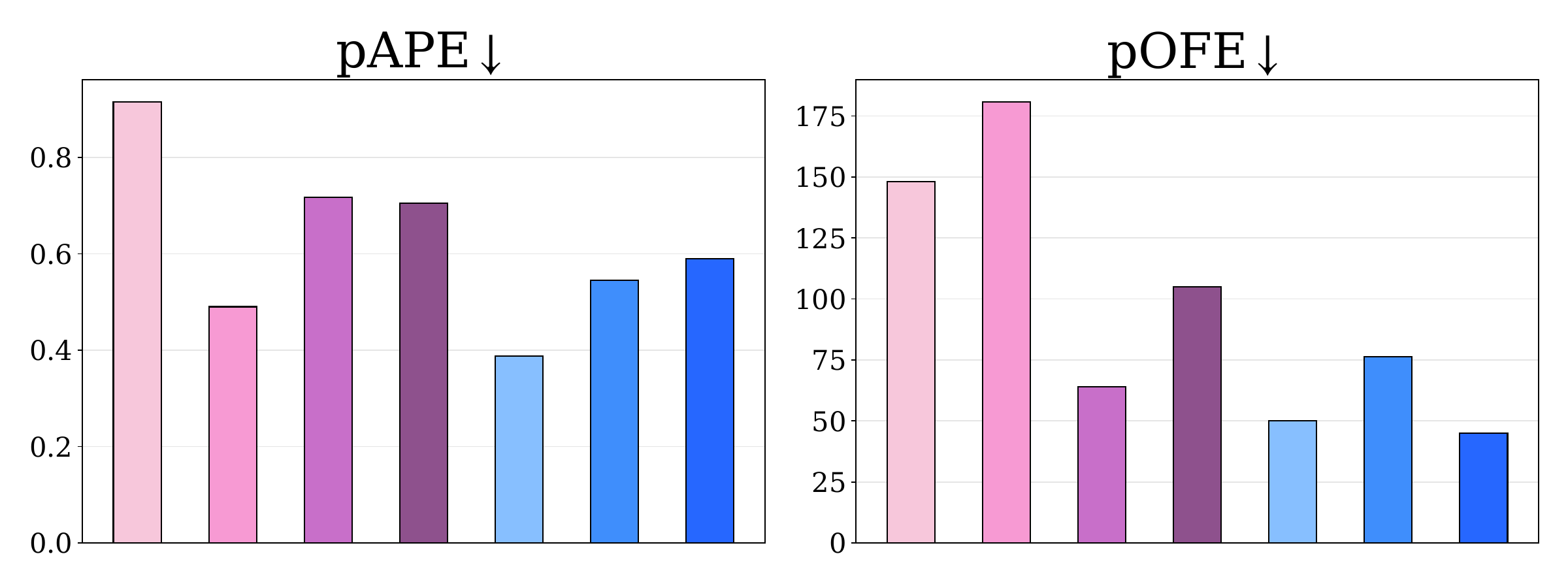}
  \caption{\textbf{Comparison of people motion metrics on \wyd$_{16}$.} \mimicmotion and \controlnext do not achieve good pAPE despite good visual quality (FVD). Analyzing further, we find that they always re-scale and re-center the generated humans (see \cref{fig:pose_rescaling})
  }
  \label{fig:people_motion}%
\end{figure}

\vspace{-\baselineskip}
\paragraph{Automatic metrics.}
\cref{fig:video_quality,fig:video_motion,fig:people_motion} show how additional metrics score our seven models w.r.t. overall and human video quality, frame-by-frame similarity and video motion.

We note that, when computing automatic metrics, the pre-processing operations of a given video generation model are also applied to the reference videos.
This is because current models were not initially designed to support all aspect ratios, and comparing their generations with the original reference videos would result in unfair evaluations.
For instance, \controlnext generates portrait videos only.
If we had not applied the same transformations to the reference videos, we would not have been able to compute pixel-level metrics.
Moreover, this ensures that the results obtained with pAPE capture scale differences (despite model-specific resolutions) and correctly identifies the re-scale and re-center issue in \mimicmotion and \controlnext described in \cref{sec:metrics}.
We encourage future work to develop models that can process different aspect ratios so as to ensure fully comparable results and benchmarking on \wyd.

\vspace{-\baselineskip}
\paragraph{Side-by-side human evaluations setup.}
For side-by-side evaluation, we sample 100 random videos from \wyd{} and ask four researchers familiar with the task.
Each researcher annotates 25 comparisons for each model pair and across four evaluation setups: (i) video quality, (ii) video motion similarity, (iii) human quality w.r.t. reference, and (iv) human motion w.r.t. reference.

For video quality, we compare all 21 model pairs; while we only compare five model pairs (\mimicmotion vs. \controlnext, \magicpose vs. \controlnext, \magicpose vs. \ctrladapter, \mimicmotion vs. \tftv, \tftv vs. \cavideo) on the other tasks due to the increased amount of time required for careful evaluations.

An example of our UI for side-by-side evaluations is shown in \cref{fig:sidexside_ui}, where we remove the actors' segmentation masks for setup (ii), and only show the generated videos for setup (i).
We used the following the templates for evaluation.
\begin{enumerate}
    \item \textbf{Video quality:}
    Choose the video that you think is of highest quality (less defects, distortions, artifacts, excessive blur, etc.). If both have the same quality, choose the one that is more appealing to you (more interesting, better composition, etc.). If both videos are equally appealing, click on ``Equally Good/Bad.''
    \item \textbf{Video motion similarity:}
    Choose the video that you think best matches the motion of the entire reference video (shown in the middle). Please try to ignore potential defects or bad quality of the videos. If both videos equally follow the motion of the reference video, click on ``Equally Good/Bad.''
    \item \textbf{Human quality w.r.t. reference:}
    Choose the video that you think has the highest quality (less defects, distortions, artifacts, excessive blur, etc.) for the people highlighted in the reference video (shown in the middle). If the highlighted people in both videos are equally good/bad, click on ``Equally Good/Bad.''
    \item \textbf{Human motion w.r.t. reference:}
    Choose the video that you think best matches the movements of the people in the reference video (shown in the middle). Please try to ignore potential defects or bad quality of the videos. If both videos equally follow the motion of the reference video, click on ``Equally Good/Bad.''
\end{enumerate}

\section{Ethics statement}\label{app:ethics}

The aim of \emph{\wydlong} (\wyd) is to enable better evaluation of current and future controllable video generation models with respect to human characters and motion, which arguably are of particular importance to people.
While this kind of models have great potential to assist and augment human creativity, there are broader societal issues that need to be considered when developing these models.

Video generative models may be misused to generate fake, hateful, explicit or harmful content.
For example, they could be used to spread misinformation and portray false situations by synthesizing fake content (\ie, deepfakes).
To mitigate these harms, digital watermarks can be applied to generated videos~\citep{Luo_2020_CVPR} to identify whether a given video was produced by a particular model.

Generative models rely on massive amounts of data harvested from the Web, which reflect social stereotypes, oppressive viewpoints, and harmful associations to marginalized identity groups~\citep{birhane2021large,birhane2021multimodal,meister2022gender}.
It is essential that generated content avoids perpetuating harmful stereotypes and respects cultural sensitivities.
In fact, models primarily trained on samples with English data may reflect Western cultures~\citep{marvl,nofilter}.
We acknowledge that our benchmark does not explicitly aim to encompass several cultures and populations, and it may perpetuate biases present in the datasets on which it is based.
We encourage future work to develop training and evaluation setups that aim to widen the social and cultural representations of these technologies.

Moreover, we note that training video generative models is computationally expensive, both financially (\eg, hardware and electricity) and environmentally, due to the carbon footprint of modern tensor processing hardware.
We encourage future research that explores more efficient architectures.

Due to the impacts and limitations described above, we remark that \wyd{} aims to measure progress in video generation research.
By no means should our data be extended for use in sensitive domains.
We believe that generative technologies, like the type of controllable image-to-video models that can be evaluated in \wyd, can become useful tools to enhance human productivity and creativity.

% WARNING: do not forget to delete the supplementary pages from your submission 
% \input{sec/X_suppl}

\end{document}